\documentclass{article}

\PassOptionsToPackage{numbers,sort&compress}{natbib}

\usepackage[main, final]{neurips_2026}

\usepackage[utf8]{inputenc} 
\usepackage[T1]{fontenc}    
\usepackage{hyperref}       
\usepackage{url}            
\usepackage{booktabs}       
\usepackage{amsfonts}       
\usepackage{amsmath}        
\usepackage{amssymb}        
\usepackage{mathtools}      
\usepackage{nicefrac}       
\usepackage{microtype}      
\usepackage{xcolor}         
\usepackage{graphicx}       
\usepackage{float}          
\usepackage{threeparttable} 
\usepackage{tcolorbox}      
\usepackage{subfiles}       
\usepackage{xspace}         

\usepackage{multicol}
\usepackage{multirow}
\usepackage{makecell}
\usepackage{pifont}
\usepackage{enumitem}

\AtBeginDocument{
  \setlength{\abovedisplayskip}{0.6ex}
  \setlength{\belowdisplayskip}{0.6ex}
  \setlength{\abovedisplayshortskip}{0.4ex}
  \setlength{\belowdisplayshortskip}{0.4ex}
}
\allowdisplaybreaks[4]

\usepackage{bbm}

\usepackage{cleveref}
\usepackage{amsthm}
\theoremstyle{plain}
\newtheorem{theorem}{Theorem}[section]
\newtheorem{proposition}[theorem]{Proposition}
\newtheorem{lemma}[theorem]{Lemma}
\newtheorem{corollary}[theorem]{Corollary}
\theoremstyle{definition}
\newtheorem{definition}[theorem]{Definition}
\newtheorem{assumption}[theorem]{Assumption}
\theoremstyle{remark}
\newtheorem{remark}[theorem]{Remark}

\newcommand{\ie}{\textit{i}.\textit{e}., }
\newcommand{\eg}{\textit{e}.\textit{g}., }
\newcommand{\wrt}{\textit{w}.\textit{r}.\textit{t}. }

\DeclareMathOperator{\sgn}{sgn}
\DeclareMathOperator{\diag}{diag}
\DeclareMathOperator{\var}{Var}
\DeclareMathOperator{\cov}{Cov}
\newcommand{\diff}{\mathrm{d}}

\newcommand{\herm}{\mathrm{H}}
\newcommand{\ssnr}{\textit{SSNR}}
\newcommand{\expt}{\mathbb{E}}
\newcommand{\bias}{\mathcal{B}}
\newcommand{\mse}{\text{MSE}}
\newcommand{\mcL}{\mathcal{L}}

\newcommand{\bst}[1]{{\textbf{\textcolor{red}{#1}}}}
\newcommand{\subbst}[1]{\textcolor{blue}{\underline{#1}}}
\newcommand{\scalea}[1]{\scalebox{0.78}{#1}}
\newcommand{\scaleb}[1]{\scalebox{0.8}{#1}}

\tcbset{
  tcbset/.style={
    colback=teal!10,
    colframe=teal!70,
    colbacktitle=teal!90,
    top=1pt,
    bottom=1pt,
    left=1pt,
    right=1pt
  }
}

\title{The Procrustean Bed of Time Series: Optimization Bias in Point-wise Loss Functions}

\author{
\begin{tabular}{c}
    Rongyao Cai$^{1}$ \quad
    Yuxi Wan$^{1}$ \quad
    Kexin Zhang$^{1\ast}$ \quad
    Ming Jin$^{2\ast}$ \\
    Zhiqiang Ge$^{3}$ \quad
    Daoyi Dong$^{4}$ \quad
    Hang Yu$^{5}$ \quad
    Yong Liu$^{1}$ \quad
    Qingsong Wen$^{6}$ \\
    \\[-0.6em]
    $^{1}$Zhejiang University \quad
    $^{2}$Griffith University \quad
    $^{3}$Southeast University \\
    $^{4}$University of Technology Sydney \quad
    $^{5}$Ant Group \quad
    $^{6}$Squirrel Ai Learning \\
    $^\ast$Corresponding authors
\end{tabular}
}

\begin{document}

\maketitle

\begin{abstract}
    Intuitively, a more deterministic time series should be easier to forecast.
    However, point-wise loss functions (\eg MSE and MAE), serving as differentiable surrogates for the ideal optimization target, score each timestamp independently and therefore disregard temporal dependence.
    This mismatch induces a systematic optimization bias that cannot be eliminated merely by improving model expressiveness or optimizer.
    To formalize this issue, we define the \textit{Expectation of Optimization Bias} (EOB) as the Kullback--Leibler divergence between the true joint distribution and the factorized i.i.d. surrogate induced by the point-wise paradigm.
    Under covariance-stationary Gaussian assumptions, we derive closed-form expressions for the stochastic component of EOB, establishing it as an irreducible lower bound on the total bias in linear systems, and further extend it to nonlinear regimes through a Gaussian mixture model lower bound.
    Crucially, we prove this bias is governed intrinsically by two data properties, \ie \textit{sequence length} and \textit{Structural Signal-to-Noise Ratio} (SSNR), regardless of specific model architecture, optimizer, or point-wise loss forms.
    This theory motivates a principled debiasing program based on sequence length reduction and structural orthogonalization, which we instantiate through DFT/DWT combined with a novel harmonized $\ell_p$ norm.
    Extensive experiments validate the predicted SSNR--horizon dynamics, resolve the classic trigonometric fitting failure as an objective-induced pathology, and demonstrate substantial plug-and-play gains.
    Notably, on iTransformer, our proposed objective reduces average MSE/MAE by 5.2\%/5.0\% in forecasting across 11 datasets and by 27.4\%/19.4\% in imputation across 9 datasets.
    \href{https://anonymous.4open.science/r/OptBias-99E5/README.md}{[Code Link]}
\end{abstract}

\section{Introduction}

\begin{figure}[h]
    \centering
    \includegraphics[width=0.98\linewidth]{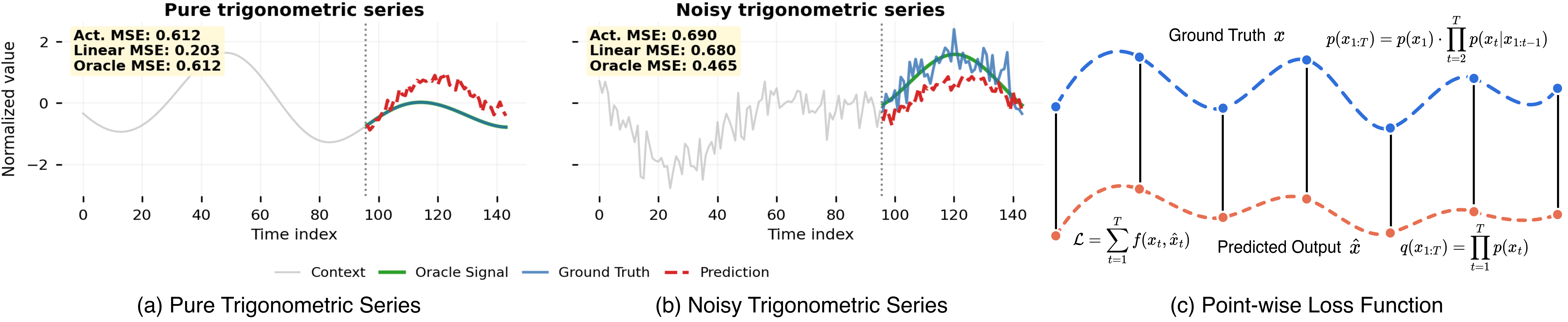}
    \vspace{-10pt}
    \caption{
        Motivation of our work.
        (a)--(b) Transformer predictions on standardized pure and noisy trigonometric series.
        (c) Point-wise losses match aligned values while ignoring temporal dependence.
    }
    \label{fig:motivation}
    \vspace{-10pt}
\end{figure}

Intuitively, a more deterministic time series should be easier to forecast. 
Yet, Figure~\ref{fig:motivation} presents a counterexample: training the same Transformer with standard Mean Squared Error (MSE) on a pure trigonometric series and on its noise-corrupted counterpart yields surprising results. 
Since actual MSE is not directly comparable across the pure and noisy processes due to an additional irreducible error floor in the noisy target, we use a process-specific linear-optimal forecaster as the reference baseline.
This baseline only exploits second-order statistics~\citep{cramer1946mathematical}; a nonlinear deep model should, in principle, extract richer structures.
However, relative to this baseline, Transformer performs worse on the pure trigonometric series than on the noisy one.
More strikingly, under the noisy setting, the Transformer prediction is closer to the oracle signal than under the pure setting.
\textit{Why does adding noise make the underlying deterministic signal easier to recover under the same point-wise objective?}

Since both cases use the same Transformer model~\citep{NIPS2017_3f5ee243}, the discrepancy cannot be explained by model expressiveness alone. 
A deep time series model is shaped by two coupled components: the \textit{architecture}, which determines what temporal patterns can be represented, and the \textit{loss function}, which serves as a differentiable surrogate for the ideal target and dictates what the model is actually optimized to learn.
The latter sets an optimization ceiling: regardless of architecture richness or optimizer employed, any temporal structure not rewarded by the loss receives no explicit gradient signal, and thus cannot be systematically recovered.
While existing research~\citep{11314178, 10496248, liu2022pyraformer, liu2024itransformer} treats progress as the search for more expressive sequence models, point-wise loss functions (\eg MSE) are adopted as defaults.
Yet, Figure~\ref{fig:motivation} shows that these defaults are not neutral. 
The central question is not just what temporal structure a model can represent, but what structure the training objective actually rewards.

Recent studies have begun to recognize the optimization bias induced by temporal MSE, but explanations remain largely heuristic~\citep{pmlr-v267-kudrat25a, qiu2025DBLoss} or tied to specific algebraic forms~\citep{wang2025fredf, wang2025timeo1}. 
Fundamentally, this flaw is shared by the entire point-wise paradigm: the point-wise objective induces an independent and identically distributed (i.i.d.) factorization at the optimization abstraction. 
Point-wise losses penalize the discrepancy between $\hat{x}_t$ and $x_t$ at each timestamp for mutli-step forecasting, but ignores whether the predicted sequence $\hat{x}_{1:T}$ preserves the dependence structure among future values. 
This omission is critical: intrinsic temporal dependence is precisely what distinguishes a time series from an unordered set of observations. 
Probabilistically, point-wise optimization replaces the true joint law $p(x_{1:T})=p(x_1)\prod_{t=2}^{T}p(x_t|x_{1:t-1})$ with a factorized surrogate $q(x_{1:T})=\prod_{t=1}^{T}p(x_t)$, collapsing temporal causality into independent marginals. \textit{The stronger the temporal structure, the more information this factorization discards.}

To quantify this mismatch, we define the \textit{Expectation of Optimization Bias} (EOB) as the Kullback--Leibler (KL) divergence~\citep{cover1999elements} between the true joint distribution and the factorized i.i.d. surrogate induced by point-wise training paradigm. 
From this distributional perspective, EOB characterizes the gap between the ideal target and the differentiable surrogate.
Importantly, model expressiveness~\citep{NEURIPS2021_5989add1} and optimizers merely bound the training's approximation of the surrogate objective.
Thus, the discrepancy between the surrogate and true target constitutes an irreducible lower bound.
Moreover, as a distributional divergence, EOB is independent of specific point-wise loss forms, yielding a theoretical lower bound for them under explicit distribution presets.

We first show that EOB decomposes into a deterministic surplus (arising from perfectly predictable components) and a stochastic term with a closed-form formula. 
Focusing on this stochastic component as an irreducible lower bound, we derive its expressive for linear systems under covariance-stationary Gaussian assumptions, extending it to nonlinear regimes via a Gaussian Mixture Model (GMM) lower bound. Critically, EOB is governed by two intrinsic data properties: \textit{sequence length} and \textit{Structural Signal-to-Noise Ratio} (SSNR), independent of any particular model architecture, optimizer, or point-wise loss form.
This diagnosis informs a principled debiasing program: sequence length reduction or structural orthogonalization. 
We instantiate the latter via DFT/DWT with a harmonized $\ell_p$ norm for stable optimization.
To summarize, our contributions are primarily theoretical in nature:
\begin{itemize}[leftmargin=*] 
    \item 
    We formalize the objective-induced bias of point-wise supervision as EOB, a KL divergence between the true joint law and its factorized surrogate, as an irreducible optimization bias.
    
    \item 
    By decomposing EOB, closed-form stochastic lower bounds are derived for linear Gaussian systems, with further extensions to nonlinear regimes via a GMM lower bound. 
    Crucially, this bound is proven to be purely data-driven (governed by sequence length and SSNR), independent of model architecture, optimizer, or loss specifics.

    \item 
    To systematically reduce EOB, a principled debiasing program is proposed, instantiating structural orthogonalization via DFT/DWT with a novel harmonized $\ell_p$ norm. The theoretical claims are rigorously validated through synthetic, trigonometric, plug-and-play, as well as practical forecasting and imputation experiments.
\end{itemize}

\section{Related Works}

\textbf{Development of Time-Series Forecasting Architectures.}\;
Model expressiveness dominates time series forecasting research, driving a rapidly evolving landscape of architectures to capture complex dependencies. Representative directions include recurrent models~\citep{unlocking_lstm_forecasting}, CNNs~\citep{donghao2024moderntcn}, GNNs~\citep{chen2024biased}, Transformers~\citep{liu2024itransformer, chen2024pathformer, jin2024timellm}, and advanced MLPs~\citep{TSMxier2023}.
While part of these works~\citep{10.1145/3711896.3736571, yang2024rethinking, zhong2025timevlm, 10.1145/3711896.3736952, xu2024fits} primarily use DFT/spectral transforms for representation or efficient modeling, our usage of DFT/DWT is fundamentally motivated by mitigating objective-induced bias.

Despite these architectural strides, point-wise losses like MSE and MAE remain unquestioned training defaults. As recent reflections note, increasingly complex designs may obscure fundamental forecasting principles~\citep{lu2026renf}. 
Existing progress largely expands what temporal structures models \textit{can} represent, neglecting whether the objective itself actually \textit{rewards} their preservation.

\textbf{Loss Functions and Heuristic Mitigation of Objective-Induced Bias.}\;
Conversely to architectural advancements, loss design has received limited systematic attention.
The shift from autoregressive to direct multi-step forecasting~\citep{informer2021} solved error accumulation but fully exposed the inherent bias of point-wise objectives. 
By outputting the entire future window simultaneously, direct forecasting removes the implicit chain-rule structure of recursive rollout, leaving the predicted steps to be evaluated as conditionally independent points by the point-wise loss.

To mitigate this, recent studies mitigate point-wise limitations via auxiliary statistical constraints: PSLoss~\citep{pmlr-v267-kudrat25a} enforces patch-level statistics, DBLoss~\citep{qiu2025DBLoss} penalizes trend/seasonal components, and DistDF~\citep{wang2026distdf} applies a joint-distribution Wasserstein constraint. 
While imposing stronger statistical supervision, they lack a unified theoretical explanation for why the point-wise paradigm fails.

A closer line of work studies the training objective of direct forecasting. 
FreDF~\citep{wang2025fredf}, Time-o1~\citep{wang2025timeo1}, and QDF~\citep{wang2026quadratic} attribute temporal MSE's weakness to operational issues (label correlation causes likelihood mismatch, and multi-task difficulty) and attempt corrections via decoupled sequences.

In contrast, we critique the point-wise paradigm itself: probabilistically, it replaces the true sequential joint law with a factorized i.i.d. surrogate.
From this perspective, label correlation and multi-task optimization difficulties are merely symptoms of this fundamental factorization, rather than isolated flaws.
This paradigm-level formalization enables us to define EOB as a KL divergence and derive closed-form bounds governed intrinsically by sequence length and SSNR.

\section{Expectation of Optimization Bias: Formalization and Quantification}
\label{sec:opt_bias}
We formalize EOB as the distributional mismatch induced by point-wise supervision, decompose it into deterministic and stochastic components (Sec.~\ref{sec:def_EOB}), quantify its stochastic lower bound for linear systems (Sec.~\ref{sec:linear_modeling}), and extend this to nonlinear regimes via GMMs (Sec.~\ref{sec:non_linear_modeling}).

\subsection{Distributional Formalization and Decomposition of EOB}
\label{sec:def_EOB}
Point-wise loss functions score a sequence by summing timestamp-wise discrepancies. 
Think of grading a song note by note: you catch if each is in tune, but miss the melody's flow.
Probabilistically, this optimizes a factorized surrogate $q(x_{1:T})$ that matches individual marginals but ignores temporal dependencies.
By contrast, the true joint law $p(x_{1:T})$ of temporal dependence follows the chain rule:
\begin{equation}
    q(x_{1:T}) = \prod_{t=1}^T p(x_t) , \quad
    p(x_{1:T}) = p(x_1) \cdot \prod_{t=2}^{T} p(x_t | x_{1:t-1}) .
\end{equation}

Decomposing the true log-likelihood isolates the temporal information discarded by the surrogate:
\begin{equation}
    \log p(x_{1:T}) 
    = \underbrace{\log q(x_{1:T})}_{\text{Implicit Objective}}
    +
    \underbrace{
    \sum_{t=2}^{T} \log \frac{p(x_t \vert x_{1:t-1})}{p(x_t)}
    }_{\text{Correction Term}} .
    \label{eq:decomp_log_p}
\end{equation}

This correction term is the sample-level "tax" for ignoring temporal structure (\ie dependence). Its expectation under the true law defines the \textbf{Expectation of Optimization Bias (EOB)}:
\begin{equation}
    \expt[\bias]
    = D_{\text{KL}}(p(x_{1:T}) \Vert q(x_{1:T})) .
\end{equation}
Thus, EOB measures the distributional gap between the true sequential law and the factorized surrogate induced by point-wise supervision.
To analyze how different structures contribute to this KL divergence, we decompose the process $x$ into deterministic $v$ and stochastic $z$ components.

\begin{theorem}
\label{thm:decomp_of_EOB}
    \textbf{(Decomposition and Lower Bound of EOB)}
    Let $\{ x_t \}$ be a discrete-time process of \textbf{continuous random variables} admitting a Cram\'er decomposition~\citep{cramer1946mathematical} $x_t = v_t + z_t$, where $\{ v_t \}$ is the deterministic component and $\{ z_t \}$ is the stochastic component. 
    Assume the following conditions hold: \ref{assump:cont_det} Continuous Determinism, \ref{assump:indep} Independence, and \ref{assump:str_ident} Structural Identifiability.
    Then, the EOB for the sequence $x_t$ over duration $T$, denoted as $\expt[\bias]$, admits the exact decomposition:
    \begin{equation}
        \expt [\bias]
        =
        \underbrace{
            \expt \left[ \sum_{t=2}^{T} \log \frac{p(z_t \vert z_{1:t-1})}{p(z_t)} \right]
        }_{ \text{Stochastic EOB: } \expt[\bias_z] := I(z_{1:T}) }
        +
        \underbrace{
            \sum_{t=2}^{T} \left[ H(x_t) - H(z_t) \right]
        }_{ \text{Deterministic Surplus: } \Delta_v \ge 0 }
    \end{equation}
	    where $H(\cdot)$ denotes differential entropy and $I(z_{1:T})$ is the total temporal mutual information of the stochastic component, \ie stochastic EOB.
\end{theorem}

\noindent\textit{Proof.}\; See Appendix~\ref{sec:proof_decomp_lower_bound_EOB}. \qed

Since $H(z_t)=H(x_t|v_t)$ via Assumption~\ref{assump:indep}, the deterministic surplus $\Delta_v$ measures the entropy reduction explained by $v_t$, which increases and can diverge in limiting deterministic regimes as this structure explains more of the observed sequence.
This leads to a counterintuitive paradox.
\begin{tcolorbox}[tcbset]
\begin{theorem}
\label{thm:paradigm_paradox}
    \textbf{(Paradigm Paradox of Point-wise Loss)} 
    The more predictable a sequence is (\ie the larger its deterministic part), the greater the EOB will be under point-wise training methods.
\end{theorem}
\end{tcolorbox}
As demonstrated experimentally in the introduction, pure trigonometric signals contain complete deterministic structure; entirely ignoring this dependence incurs massive optimization bias.
Since the deterministic surplus is data-specific to formulation and potentially divergent, we hereafter focus on the finite, estimable lower bound: the \textbf{stochastic EOB} $\expt[\bias_z] := I(z_{1:T})$.

\subsection{Linear Quantification of Stochastic EOB}
\label{sec:linear_modeling}
To quantify the stochastic EOB lower bound, we approach the problem from two complementary perspectives: a parametric dynamical Autoregressive Model (AR) and a non-parametric distributional Multivariate Gaussian Model (MGM), which elegantly converge to a unified formulation.

\textbf{View 1 - Parametric Dynamical Autoregressive Model.}\;
We parameterize temporal dependence via a Gaussian $AR(p)$ process through finite-order innovation:
\begin{equation}
    z_{t} = c + \sum_{i=1}^{p} \phi_i z_{t-i} + \epsilon_{t}
    \label{eq:def_ar_p}
\end{equation}
where $\epsilon_t \sim \mathcal{N}(0, \sigma_\epsilon^2)$ is a independent white innovation process independent of the past. 
For a covariance-stationary process, $\sigma_z^2=\var(z_t)$ is the unconditional variance, and $\sigma_\epsilon^2$ represents the irreducible one-step prediction uncertainty.


\begin{tcolorbox}[tcbset]
\begin{proposition}
\label{prop:quan_EOB_AR}
    \textbf{(Quantification of Stochastic EOB via $AR(p)$)} 
    For any covariance-stationary Gaussian $AR(p)$ with $T>p$, the stochastic EOB incurred by point-wise factorized surrogate is:
    \begin{equation}
        \expt[\bias_{z}] 
        = \frac{T-p}{2} \log \frac{\sigma_z^2}{\sigma_\epsilon^2} + C_p
        \label{eq:EOB_AR}
    \end{equation}
    where $C_p = -\frac{1}{2}\log \vert R_p \vert$ is a $T$-independent initial-state correction with correlation matrix $R_p$.
\end{proposition}
\end{tcolorbox}
\noindent\textit{Proof.}\; See Appendix~\ref{app:quan_EOB_AR}. \qed


Eq.~\eqref{eq:EOB_AR} isolates stochastic EOB into a corrected sequence length ($T-p$) and a variance ratio ($\sigma_z^2/\sigma_\epsilon^2$).
The former quantifies the steady-state horizon over which the point-wise surrogate severs temporal dependence, while the latter contrasts the unconditional sequence uncertainty against its residual innovation noise.
This ratio motivates an intrinsic measure of predictability:
\begin{tcolorbox}[tcbset]
\begin{definition}
\label{def:ssnr}
    \textbf{(Structural Signal-to-Noise Ratio (SSNR))} 
    For a covariance-stationary process, SSNR is the ratio of total unconditional variance ($\sigma_{z}^2$) to its steady-state one-step-ahead optimal prediction error variance ($\sigma_{\epsilon}^2$):
    \begin{equation}
        \ssnr 
        = \frac{\sigma_{z}^2}{\sigma_{\epsilon}^2} 
        = \frac{\var(z_t)}{\var(z_t - \expt[z_t | z_{<t}])} .
    \end{equation}
\end{definition}
\end{tcolorbox}

Intuitively, SSNR quantifies the proportion of variability predictable from history.
Higher SSNR indicates stronger internal temporal structure, whereas $\ssnr=1$ implies pure white noise.
Its relationship to classical SNR is detailed in Appendix~\ref{app:relationship_snr}.

\textbf{View 2 - Non-Parametric Distributional Multivariate Gaussian Model.}\;
Rather than specifying step-wise evolution, this non-parametric distributional MGM view captures the joint correlation structure across all $T$ observations in a finite window simultaneously: $\boldsymbol{\mathrm{z}} \sim \mathcal{N}(\boldsymbol{\mu}, \Sigma)$.

\begin{tcolorbox}[tcbset]
\begin{proposition}
\label{prop:quan_EOB_MGM}
    \textbf{(Quantification of Stochastic EOB via MGM)} 
    For any finite window governed by MGM, the stochastic EOB is determined by the determinant of its correlation matrix, $\vert R \vert$:
    \begin{equation}
        \expt[\bias_z] 
        = - \frac{1}{2} \log \vert R \vert.
    \label{eq:EOB_MGM}
    \end{equation}
\end{proposition}
\end{tcolorbox}
\noindent\textit{Proof.}\; See Appendix~\ref{app:quan_EOB_MGM}. \qed

The determinant $\vert R \vert$ measures window-wise correlation: as dependence grows, $\vert R \vert \to 0$ and the bias diverges. This is a compact, elegant way to capture the same intuition as the AR formula.

\textbf{Unified Expression of Stochastic EOB.}\;
Although the AR and MGM views look different, they ultimately tell the same story.
Taking the window length to infinity connects this non-parametric view to the infinite-history representation of stationary Gaussian dynamics.
By Lemma~\ref{lem:convergence_R}, as $T \to \infty$,
\begin{equation}
    \lim_{T \to \infty} \vert R \vert^{1/T} 
    = \frac{\sigma_{\epsilon}^2}{\sigma_z^2}
    = \frac{1}{\ssnr}.
\end{equation}

Consequently, the average per-step stochastic EOB converges to:
\begin{equation}
    \lim_{T \to \infty} - \frac{1}{2T} \log \vert R \vert
    = \lim_{T \to \infty} \frac{1}{2} \log \frac{1}{\vert R \vert^{1/T}} 
    = \frac{1}{2} \log (\ssnr).
    \label{eq:EOB_asym_rate}
\end{equation}
In other words, every additional horizon contributes $\frac{1}{2} \log (\ssnr)$ of bias on average.
This asymptotic rate is the common limit of MGM and AR views, regardless of the specific parametrization.
For a practical finite window, Lemma~\ref{lem:decomp_R_p} provides the exact decomposition:
\begin{equation}
    \vert R \vert 
    = \vert R_p \vert \cdot ( \ssnr )^{-(T-p)}
\end{equation}
where $\vert R_p \vert$ is the determinant of the correlation matrix of the initial $p$ observations. 

Substituting this into Eq.~\eqref{eq:EOB_MGM} recovers the AR expression in Eq.~\eqref{eq:EOB_AR}, yielding our central theorem:

\begin{tcolorbox}[tcbset]
\begin{theorem}
\label{thm:quan_EOB_general}
    \textbf{(Quantification of Stochastic EOB)} 
    For any covariance-stationary Gaussian $AR(p)$ process with $T>p$, the stochastic EOB, $\expt[\bias_z]$, induced by the point-wise factorized surrogate is determined exclusively by the sequence length $T$ and its SSNR:
    \begin{equation}
        \expt[\bias_z] = \frac{T}{2} \log (\ssnr) + C(p, \boldsymbol{\phi})
    \label{eq:EOB_general}
    \end{equation}
    where $C(p, \boldsymbol{\phi}) = - \frac{1}{2} \log (\vert R_p \vert \cdot (\ssnr)^p)$ is a constant about data properties and order $p$.
\end{theorem}
\end{tcolorbox}

Crucially, this irreducible bias imposed by the point-wise paradigm is governed purely by $T$ and SSNR, independent of the chosen model architecture or optimizer. This provides a foundational theoretical grounding for related empirical studies~\citep{wang2025fredf,wang2025timeo1,wang2026distdf,wang2026quadratic}: label correlation is a quantitative manifestation of SSNR, while multi-task optimization difficulty is the operational consequence of linear accumulation with $T$.

\subsection{Nonlinear Extension of Stochastic EOB}
\label{sec:non_linear_modeling}

While the preceding analysis assumes a single linear Gaussian process, real-world time series frequently exhibit nonlinear dynamics and transition across multiple behavioral regimes. To accommodate such complexities, we extend our quantification via a Gaussian Mixture Model (GMM), representing the stochastic component as a probabilistic ensemble of $K$ stationary Gaussian regimes:
\begin{equation}
    p(z_{1:T}) = \sum_{k=1}^K \pi_k p_k(z_{1:T}), 
    \quad
    p_k(z_{1:T}) = \mathcal{N}(\boldsymbol{\mu}_k, \Sigma_k).
\end{equation}
For the mixture log-likelihood involves a log-sum structure, $\expt[\bias_z]$ lacks a direct closed-form solution. 
Nevertheless, leveraging \textit{Jensen's inequality} on differential entropy yields a tractable lower bound:
\begin{equation}
    \expt[\bias_z]
    \ge \sum_{k=1}^K \pi_k \expt[\bias_k] - H(\pi)
\label{eq:EOB_GMM}
\end{equation}
where $\expt[\bias_k]$ is the stochastic EOB within $k$-th Gaussian component, and $H(\pi)$ is the Shannon entropy of discrete mixing distribution.
Crucially, each component-specific bias $\expt[\bias_k]$ is governed by its own $\ssnr_k$ and $T_k$.
Thus, this nonlinear extension preserves the same intrinsic drivers, aggregating them across mixture regimes while subtracting $H(\pi)$ to account for the uncertainty of regime assignment.

\noindent\textit{Proof.}\; See Appendix~\ref{app:GMM_modeling}. \qed

\section{Discussion and Limitation of EOB Theory}
\label{sec:discussion}

An immediate consequence of EOB analysis is the \textbf{Predictability Penalty} of the point-wise paradigm.
\begin{corollary}
\label{cor:predictability_penalty}
    \textbf{(Predictability Penalty of Point-wise Loss)}
    The total EOB, $\expt[\bias]$, increases with the deterministic surplus (predictability) of the observed sequence, while its stochastic lower bound, $\expt[\bias_z]$, increases with the SSNR of the stochastic component.
\end{corollary}

\textbf{Distinguishing EOB from Empirical Error Metrics.}
Error metrics (\eg MSE) are \textit{a posteriori} observations of a trained model's realized performance, heavily influenced by model expressiveness, optimization dynamics, and engineering factors.
By contrast, EOB is an \textit{a priori} theoretical estimate of the irreducible mismatch fundamentally imposed by the optimization objective itself, independent of training. 
While EOB does not numerically coincide with empirical metrics, it constrains the best attainable performance under point-wise supervision and translates into lower bounds on observable errors under specific distributional assumptions.

\textbf{Reframing Empirical Observations.}
The paradigm paradox also reframes a classic empirical observation: neural models often struggle to fit highly periodic signals~\citep{NEURIPS2020_11604531}.
Typically attributed to limited expressiveness or spectral bias, our theory identifies a fundamental objective-level cause.
Highly periodic signals possess extreme temporal predictability, resulting in a massive EOB under point-wise supervision.
Thus, failure on pure periodic curves often reflects a systematically biased training objective rather than an inability to represent the signal.

\textbf{Further Remarks.}
A systematic interpretation of existing loss studies through the lens of EOB, along with a detailed discussion of the theory's scope and limitations, is provided in Appendix~\ref{app:discussion}.

\section{Principled Debiasing Program}
\label{sec:debiasing_prog}

Having identified sequence length ($T$) and SSNR as the intrinsic drivers of EOB, we now turn from diagnosis to intervention. A principled debiasing program must not merely explain the bias, but directly act on the structural quantities that generate it.

\textbf{Existing Strategy: Heuristic Statistical Alignment.}
Before introducing our principles, we contextualize existing mitigations.
A common approach to alleviating point-wise limitations is appending auxiliary terms to align local or global statistics $\mathcal{S}$, (\eg patch statistics, or forecast-label distributions; see Appendix~\ref{app:F1}). 
Abstracted as $\hat{\mcL} = \mcL_{\text{PW}} + \mcL(\mathcal{S})$, these methods impose structural constraints inherently ignored by point-wise losses.
However, they largely treat the symptoms rather than the root cause. A fundamental intervention must instead target $T$ and SSNR.

\textbf{Principle 1: Sequence Length Reduction.}
Since EOB accumulates linearly with sequence length, one direct intervention is to reduce the effective prediction horizon while maintaining or reducing SSNR, thereby decreasing the number of conditionally independent assumptions penalized by the surrogate.
This is achieved via a reversible \textit{compression operator}, $\mathcal{T}_{\text{comp}}(\cdot)$, mapping the target series to a compact representation of length $T' < T$. 
The model is optimized in this compressed space:
\begin{equation}
    x_c = \mathcal{T}_{\text{comp}}(x) \in \mathbb{R}^{T'}, 
    \quad \hat{x}_c = f_\theta(\cdot) \in \mathbb{R}^{T'},
    \quad \hat{x} = \mathcal{T}_{\text{comp}}^{-1}(\hat{x}_c) \in \mathbb{R}^{T},
    \quad \textit{s.t. } T' < T .
\end{equation}

\textbf{Principle 2: Structural Orthogonalization.}
Since SSNR is the \textbf{structural source} of temporal dependence, the most fundamental intervention is to transform the target into a coordinate system where components are statistically uncorrelated.
If the sequence is perfectly orthogonalized such that $\cov\!\left(\mathcal{T}_{\text{ortho}}(x)\right) \approx \diag(\sigma_1^2, \dots, \sigma_T^2)$, the factorized point-wise surrogate aligns exactly with the true joint distribution in the transformed domain, theoretically neutralizing the EOB. 
This operates via a reversible \textit{orthogonality operator} $\mathcal{T}_{\text{ortho}}(\cdot)$:
\begin{equation}
    x_o = \mathcal{T}_{\text{ortho}}(x), 
    \quad \hat{x}_o = f_\theta(\cdot),
    \quad \hat{x} = \mathcal{T}_{\text{ortho}}^{-1}(\hat{x}_o),
    \quad \textit{s.t. } \cov\!\left(\mathcal{T}_{\text{ortho}}(X)\right) \approx \diag(\sigma_1^2, \dots, \sigma_T^2).
\end{equation}
Crucially, applying a unitary transform alone is insufficient. Due to \textit{Parseval's theorem}, standard $\ell_2$ distance is invariant under unitary transformations (See Theorem~\ref{thm:mse_invariance}).

\section{Concrete Instantiations of the Debiasing Program}
We instantiate Principle~2 by a reversible structural transform $\mathcal{T}$ to the supervised target.
Rather than constrains the sequence directly, the model's prediction $\hat{x}$ is transformed to $\hat{f}=\mathcal{T}(\hat{x})$, supervised by the exact transformed target $f=\mathcal{T}(x)$.
We employ the \textit{Discrete Fourier Transform} (DFT) to isolate global periodic structure and the \textit{Discrete Wavelet Transform} (DWT) to capture multi-scale dynamics.
By representing the sequence through approximately orthogonal frequency or wavelet components, these transforms effectively suppress off-diagonal covariance.
Provided the chosen basis aligns with the intrinsic data characteristics, this orthogonalization inherently minimizes the SSNR.
Detailed analyses of their structural and gradient properties are provided in Appendix~\ref{app:adv_analysis_DFT}, \ref{app:adv_analysis_DWT}, and~\ref{app:grad_analysis}.

\subsection{Harmonized $\ell_p$ Norm Optimization}
Structural orthogonalization weakens dependence, but transformed coordinates can still have highly non-uniform magnitudes with large dynamics.
To avoid scale-dominated optimization without changing the component-wise optimum, we use a harmonized weighted norm:
\begin{equation}
    \mcL_{\text{Harm}, \ell_p}
    =
    \sum_{k=1}^{K} w_k \Vert f_k-\hat{f}_k \Vert_p^p,
    \quad w_k>0 .
\end{equation}
For the main experiments with $\ell_1$, we set weight $w_k$ Exponential Moving Average (EMA) updated with ground truth $f_k$ in each epoch:
\begin{equation}
    w_k = 1+\gamma\bar{f}_k,
    \quad
    \bar{f}^{(e+1)}_k 
    = \beta \bar{f}^{(e)}_k
      + (1 - \beta) \vert f^{(e)}_k \vert .
\end{equation}
Thus,
\begin{equation}
    \mcL_{\text{Harm}, \ell_1}
    =
    \sum_{k=1}^{K} (1+\gamma\bar{f}_k)\Vert f_k-\hat{f}_k \Vert_1 .
\end{equation}
The EMA update costs only $\mathcal{O}(K)$ additional storage.
Detailed harmonized MSE (hMSE) / hMAE gradient behavior and the optimum-preservation analysis are provided in Appendices~\ref{app:grad_l_p_analysis} and~\ref{app:stat_harm}.

\section{Experiments}
We organize the experiments as a logical evidence chain: we first validate the theoretical predictions of EOB under controlled processes, then verify the mechanism of structural orthogonalization, and finally evaluate whether the resulting objective improves practical forecasting and imputation tasks.

\subsection{Experimental Settings}
\label{sec:settings}
\textbf{Datasets.}
We utilize two categories of datasets to support this progression:
\textbf{(1)} A synthetic hybrid process combining deterministic trigonometric components with a stochastic autoregressive component. This allows controlled variation of SSNR and forecast horizon to rigorously test the Paradigm Paradox (Appendix~\ref{app:synthetic_dataset}).
\textbf{(2)} Eleven real-world benchmarks covering diverse domains and temporal resolutions, used to evaluate the practical efficacy of the proposed debiasing program (Appendix~\ref{app:actual_dataset}).

\textbf{Baselines.}
We compare against four groups of state-of-the-art baselines. The first three groups evaluate architectural advancements:
\textbf{(1)} Transformer-based: iTransformer, PatchTST, Pyraformer, FEDformer, Autoformer;
\textbf{(2)} MLP-based: TimeMixer, TSMixer, DLinear, FreTS;
\textbf{(3)} CNN-based: TimesNet, MICN;
\textbf{(4)} To specifically evaluate objective-level improvements, we compare against FreDF, a recent transform-domain training objective.
Details are provided in Appendix~\ref{app:backbone}.

\textbf{Implementation.} 
We reproduce all baselines using the official scripts from \citep{liu2024itransformer} and \citep{wang2025fredf}. 
To strictly isolate the impact of the objective function, both $\mcL_{\text{Harm}, \ell_p}$ and FreDF are deployed on the iTransformer backbone.
Unless otherwise stated, our model is optimized solely using $\mcL_{\text{Harm}, \ell_p}$ without any auxiliary temporal MSE term ($\gamma=0.5$, $\beta=0.3$). 
Further implementation details are provided in Appendix~\ref{app:implementation}.

\subsection{Empirical Verification of EOB Theory}
\label{sec:EOB_verif}

\begin{figure}[t]
    \centering
    \vspace{-6pt}
    \begin{minipage}[t]{0.48\linewidth}
        \centering
        \includegraphics[width=\linewidth]{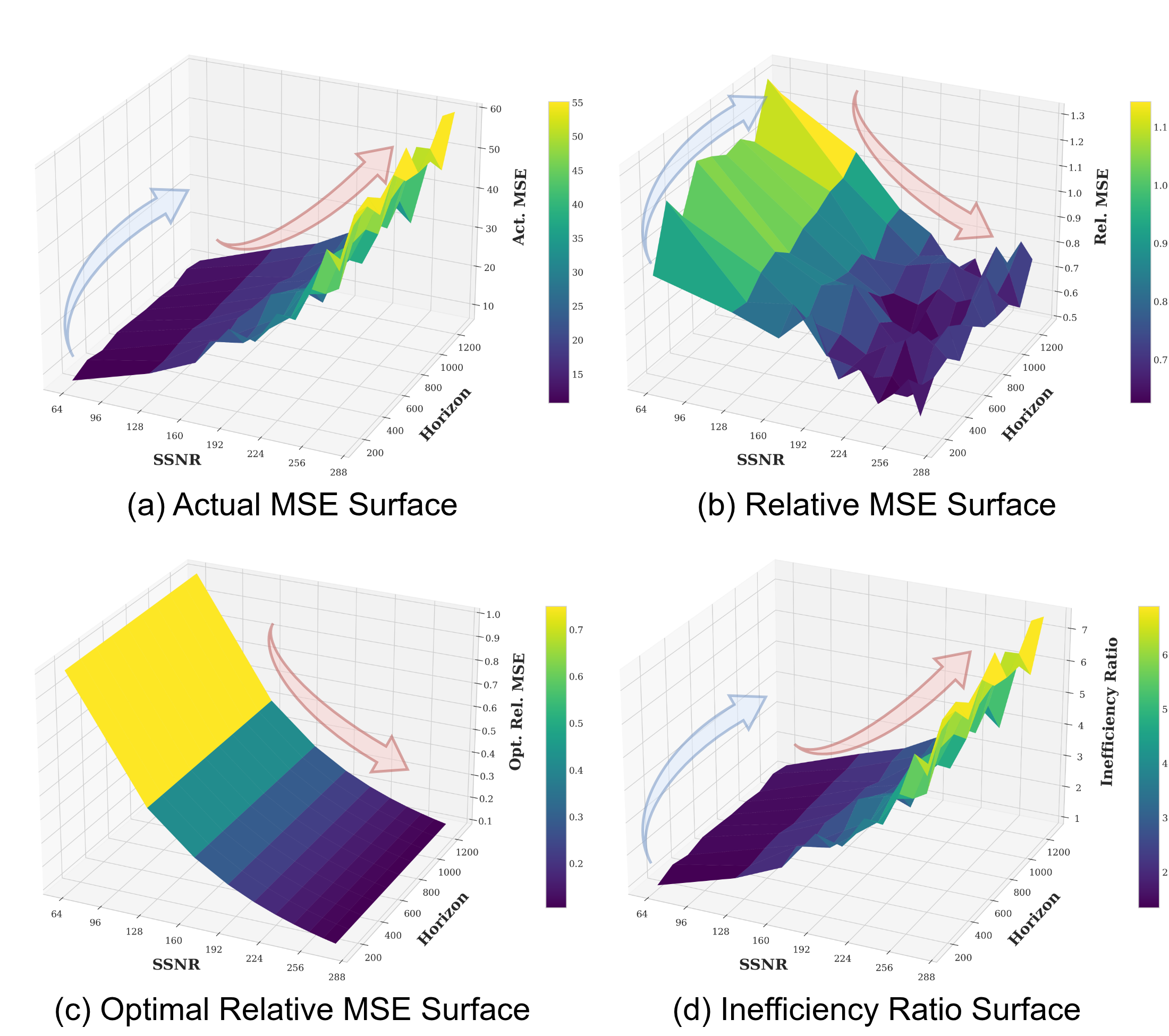}
        \vspace{-14pt}
        \caption{Transformer error surfaces under Gaussian innovations. Arrows indicate variation along horizon ($h$) and total SSNR ($\ssnr_x$).}
        \label{fig:simulation}
    \end{minipage}
    \hfill
    \begin{minipage}[t]{0.48\linewidth}
        \centering
        \includegraphics[width=\linewidth]{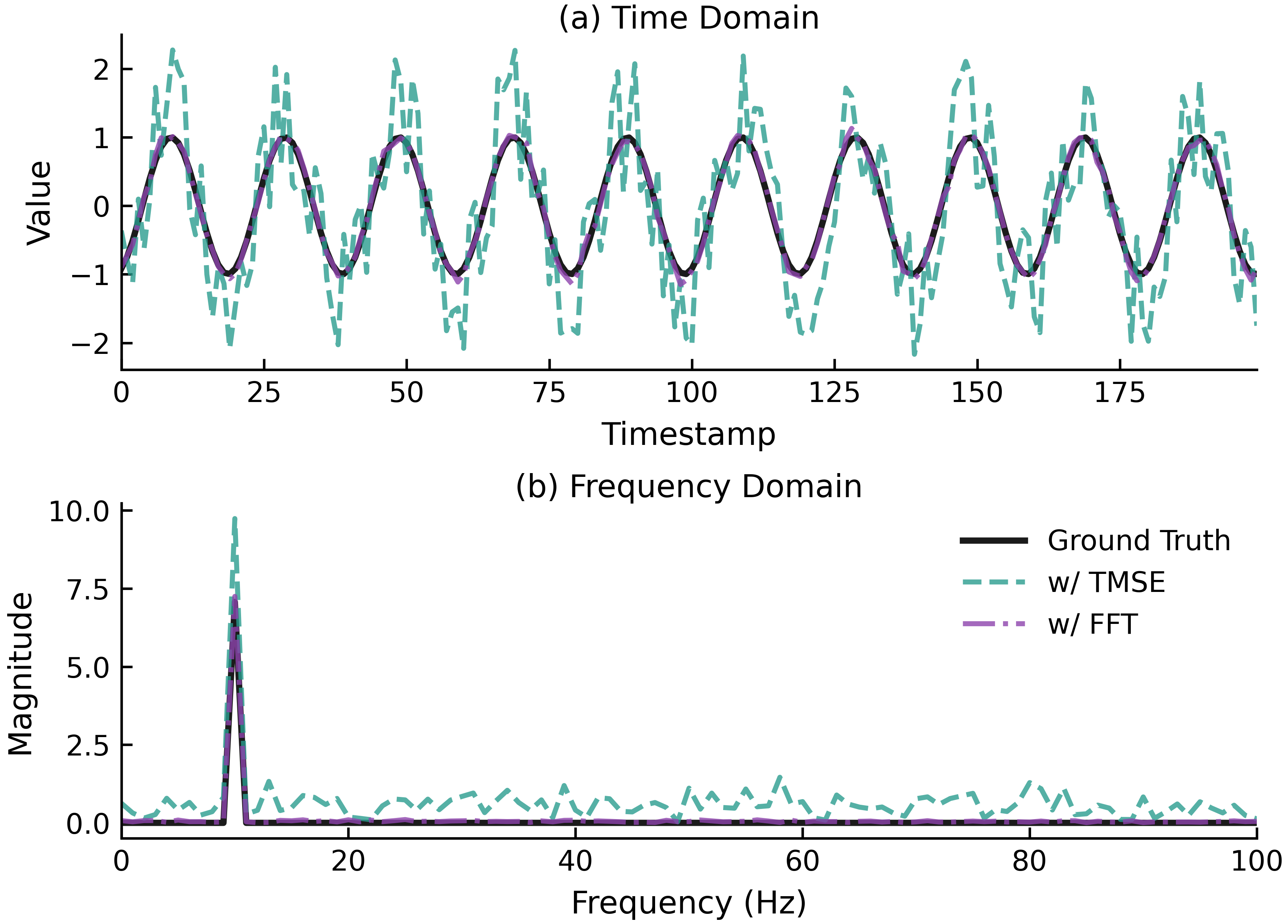}
        \vspace{-12pt}
        \caption{Mechanistic insight. Temporal MSE captures the coarse trend but with high-frequency artifacts, while pure FFT amplitude--phase supervision recovers the periodic structure cleanly.}
        \label{fig:trig_insight}
    \end{minipage}
    \vspace{-14pt}
\end{figure}

We test EOB theory on a controlled hybrid process by amplifying the deterministic component while fixing the stochastic one, thereby increasing the total SSNR ($\ssnr_x$).
Grid experiments span \textit{five architectures and six innovation distributions}; Figure~\ref{fig:simulation} illustrates the Transformer results under Gaussian innovations (full settings in Appendix~\ref{app:impl_EOB_theory}).

As shown in Figure~\ref{fig:simulation}(a), the actual MSE ($\mse_{\text{act.}}$) alone is an insufficient diagnostic, because increasing $\ssnr_x$ inherently inflates the total variance $\sigma_x^2$.
Indeed, Figure~\ref{fig:simulation}(b) demonstrates that relative MSE ($\mse_{\text{rel.}}$) decreases with $\ssnr_x$, confirming the intuition that more deterministic signals are easier to approximate in terms of relative error.
The Paradigm Paradox emerges only when comparing this outcome against the theoretical linear optimum.
Denoting the theoretical $h$-step irreducible error as $\mse^{\text{opt.}}_{\text{act.}}$, we relate actual and relative errors as follows:
\begin{equation}
    \mse_{\text{act.}} 
    = \sigma_x^2 \cdot \mse_{\text{rel.}}, \quad
    \mse_{\text{rel.}}^{\text{opt.}}
    =  \mse^{\text{opt.}}_{\text{act.}} / \sigma_x^2 .
\end{equation}

\textbf{Inefficiency ratio.}
We then evaluate the gap between actual performance and theoretical optimality:
\begin{tcolorbox}[tcbset]
\begin{definition}
\label{def:inefficiency_ratio}
    \textbf{(Inefficiency Ratio)}
    For a trained predictor under a given $(h,\ssnr_x)$ setting, the \textit{inefficiency ratio} evaluates the gap of its actual performance from the theoretical linear optimum:
    \begin{equation}
        \eta(h, \ssnr_x)
        = \frac{\mse_{\text{act.}}}{\mse^{\text{opt.}}_{\text{act.}}}
        = \frac{\mse_{\text{rel.}}}{\mse^{\text{opt.}}_{\text{rel.}}}
    \end{equation}
\end{definition}
\end{tcolorbox}

\textbf{Empirical Validation.}
$\eta(h,\ssnr_x)$ serves as an empirical proxy for the optimization gap predicted by the per-step EOB.
Theoretically, Eq.~\eqref{eq:EOB_asym_rate} establishes that the average per-step stochastic EOB scales with SSNR and approaches the limitation $\frac{1}{2}\log(\ssnr)$ over extended horizons.
Consistently, Figure~\ref{fig:simulation}(d) shows that $\eta(h,\ssnr_x)$ grows sharply with $\ssnr_x$ and saturates along $h$.
This firmly verifies the Paradigm Paradox and our asymptotic analysis: highly predictable sequences may appear easier to fit in relative terms, yet they become profoundly inefficient against their theoretical optimum.

\subsection{Mechanistic Insight on Trigonometric Series}
\label{sec:trig_insight}

This experiment revisits the \textit{classic empirical observation} discussed in Section~\ref{sec:discussion}: simple neural models often struggle to fit highly periodic signals.
We evaluate pure Fourier series with harmonics $k=1,\dots,5$, presenting the single-frequency case in Figure~\ref{fig:trig_insight} and multi-harmonic cases in Appendix~\ref{app:insight_exp}.
Across all cases, while temporal MSE captures the coarse trend, its local waveform is severely distorted, and its frequency distribution suffers from massive spectral leakage across almost all bins.
In contrast, pure FFT amplitude--phase supervision perfectly reconstructs the ground truth, aligns with the true frequencies, and exhibits no visible high-frequency noise.
Mechanistically, FFT supervision constrains approximately orthogonal amplitude and phase coordinates, making the supervised targets nearly statistically independent.
The frequency-domain SSNR is much lower than the temporal SSNR, reducing the EOB induced by point-wise temporal supervision.
This supports the EOB view that the bottleneck lies in the optimization objective rather than model expressiveness.

\subsection{Plug-and-Play Versatility and Practical Tasks}
\label{sec:plug_and_play_versatility}

This experiment investigates whether the debiasing effect stems directly from the supervision objective.
If EOB is induced by fundamental consequence of point-wise temporal supervision, modifying only the loss function should confer performance across diverse architectures without requiring any changes to their underlying modeling modules.
We therefore hold each backbone fixed and replace temporal MSE with $\mcL_{\text{Harm}, \ell_1}$, evaluating the impact across six datasets and five representative backbones.

\begin{table*}[t]
\centering
\vspace{-6pt}
\caption{\textbf{Plug-and-Play average results across horizon $h \in \{96, 192, 336, 720\}$.} See Appendix~\ref{app:plug-and-play}.}
\label{tab:plug_and_play_avg}
\renewcommand{\arraystretch}{0.82}
\setlength{\tabcolsep}{1.8pt}
\scriptsize
\resizebox{0.98\linewidth}{!}{
\begin{tabular}{c|cc|cc|cc|cc|cc|cc|cc|cc|cc|cc}
    \toprule
    \scaleb{Models} & 
    \multicolumn{4}{c}{\scaleb{TimeMixer}} & \multicolumn{4}{c}{\scaleb{iTransformer}} &
    \multicolumn{4}{c}{\scaleb{PatchTST}} & \multicolumn{4}{c}{\scaleb{TimesNet}} &
    \multicolumn{4}{c}{\scalea{DLinear}} \\
    \cmidrule(lr){2-5} \cmidrule(lr){6-9} \cmidrule(lr){10-13} \cmidrule(lr){14-17} \cmidrule(lr){18-21}
    \scaleb{Loss} & 
    \multicolumn{2}{c}{$\mcL_{\text{Harm}, \ell_1}$} & \multicolumn{2}{c}{$\mcL_{\text{TMSE}}$} & 
    \multicolumn{2}{c}{$\mcL_{\text{Harm}, \ell_1}$} & \multicolumn{2}{c}{$\mcL_{\text{TMSE}}$} & 
    \multicolumn{2}{c}{$\mcL_{\text{Harm}, \ell_1}$} & \multicolumn{2}{c}{$\mcL_{\text{TMSE}}$} & 
    \multicolumn{2}{c}{$\mcL_{\text{Harm}, \ell_1}$} & \multicolumn{2}{c}{$\mcL_{\text{TMSE}}$} & 
    \multicolumn{2}{c}{$\mcL_{\text{Harm}, \ell_1}$} & \multicolumn{2}{c}{$\mcL_{\text{TMSE}}$} \\
    \cmidrule(lr){2-3} \cmidrule(lr){4-5} \cmidrule(lr){6-7} \cmidrule(lr){8-9}
    \cmidrule(lr){10-11} \cmidrule(lr){12-13} \cmidrule(lr){14-15} \cmidrule(lr){16-17}
    \cmidrule(lr){18-19} \cmidrule(lr){20-21}
    \scaleb{Metrics} & 
    \scalea{MSE} & \scalea{MAE} & \scalea{MSE} & \scalea{MAE} &
    \scalea{MSE} & \scalea{MAE} & \scalea{MSE} & \scalea{MAE} &
    \scalea{MSE} & \scalea{MAE} & \scalea{MSE} & \scalea{MAE} &
    \scalea{MSE} & \scalea{MAE} & \scalea{MSE} & \scalea{MAE} &
    \scalea{MSE} & \scalea{MAE} & \scalea{MSE} & \scalea{MAE} \\
    \toprule
    \scalea{ETTh1} & 
    \bst{\scalea{0.448}} & \bst{\scalea{0.430}} & \scalea{0.455} & \scalea{0.443} &
    \bst{\scalea{0.436}} & \bst{\scalea{0.434}} & \scalea{0.461} & \scalea{0.453} &
    \bst{\scalea{0.433}} & \bst{\scalea{0.426}} & \scalea{0.454} & \scalea{0.450} &
    \bst{\scalea{0.470}} & \bst{\scalea{0.455}} & \scalea{0.475} & \scalea{0.464} &
    \bst{\scalea{0.452}} & \bst{\scalea{0.448}} & \scalea{0.461} & \scalea{0.457} \\
    \scalea{ETTm1} & 
    \bst{\scalea{0.383}} & \bst{\scalea{0.382}} & \scalea{0.385} & \scalea{0.399} &
    \bst{\scalea{0.393}} & \bst{\scalea{0.390}} & \scalea{0.414} & \scalea{0.414} &
    \scalea{0.393} & \bst{\scalea{0.390}} & \bst{\scalea{0.390}} & \scalea{0.403} &
    \bst{\scalea{0.401}} & \bst{\scalea{0.402}} & \scalea{0.407} & \scalea{0.416} &
    \bst{\scalea{0.401}} & \bst{\scalea{0.395}} & \scalea{0.410} & \scalea{0.412} \\
    \scalea{ECL} & 
    \bst{\scalea{0.186}} & \bst{\scalea{0.264}} & \scalea{0.243} & \scalea{0.347} &
    \bst{\scalea{0.170}} & \bst{\scalea{0.260}} & \scalea{0.176} & \scalea{0.267} &
    \bst{\scalea{0.203}} & \bst{\scalea{0.289}} & \scalea{0.208} & \scalea{0.296} &
    \bst{\scalea{0.208}} & \bst{\scalea{0.297}} & \scalea{0.214} & \scalea{0.310} &
    \bst{\scalea{0.261}} & \bst{\scalea{0.352}} & \scalea{0.266} & \scalea{0.362} \\
    \scalea{BE} & 
    \bst{\scalea{0.287}} & \bst{\scalea{0.298}} & \scalea{0.300} & \scalea{0.313} &
    \bst{\scalea{0.295}} & \bst{\scalea{0.293}} & \scalea{0.332} & \scalea{0.324} &
    \bst{\scalea{0.304}} & \bst{\scalea{0.315}} & \scalea{0.313} & \scalea{0.324} &
    \bst{\scalea{0.279}} & \bst{\scalea{0.286}} & \scalea{0.316} & \scalea{0.326} &
    \scalea{0.366} & \bst{\scalea{0.360}} & \bst{\scalea{0.363}} & \scalea{0.368} \\
    \scalea{Weather} & 
    \bst{\scalea{0.243}} & \bst{\scalea{0.270}} & \scalea{0.245} & \scalea{0.275} &
    \bst{\scalea{0.256}} & \bst{\scalea{0.272}} & \scalea{0.260} & \scalea{0.281} &
    \bst{\scalea{0.258}} & \bst{\scalea{0.274}} & \scalea{0.260} & \scalea{0.281} &
    \bst{\scalea{0.262}} & \bst{\scalea{0.285}} & \bst{\scalea{0.262}} & \scalea{0.288} &
    \bst{\scalea{0.266}} & \bst{\scalea{0.312}} & \bst{\scalea{0.266}} & \scalea{0.317} \\
    \scalea{PJM} & 
    \bst{\scalea{0.289}} & \bst{\scalea{0.354}} & \scalea{0.304} & \scalea{0.373} &
    \bst{\scalea{0.285}} & \bst{\scalea{0.353}} & \scalea{0.294} & \scalea{0.366} &
    \bst{\scalea{0.280}} & \bst{\scalea{0.353}} & \scalea{0.307} & \scalea{0.374} &
    \bst{\scalea{0.275}} & \bst{\scalea{0.348}} & \scalea{0.299} & \scalea{0.366} &
    \scalea{0.327} & \bst{\scalea{0.389}} & \bst{\scalea{0.324}} & \scalea{0.398} \\
    \bottomrule
\end{tabular}
}
\vspace{-10pt}
\end{table*}

Table~\ref{tab:plug_and_play_avg} presents the average results across four forecasting horizons.
Notably, $\mcL_{\text{Harm}, \ell_1}$ improves performance in 55 out of 60 comparisons (with 2 ties), demonstrating that these gains are not specific to a single dataset, architecture, or metric.
This robust outcome substantiates our objective-level hypothesis: structural orthogonalization yields an effective, plug-and-play loss that mitigates EOB.

Broader forecasting and imputation benchmarks provide a comprehensive test of whether this objective enhances its host backbone while maintaining competitive against full model baselines.
As $\mcL_{\text{Harm}, \ell_p}$ is integrated with iTransformer, direct comparison with vanilla iTransformer strictly isolates the contribution of the objective.
In long-term forecasting tasks (Appendix~\ref{app:long_term_forecast}), the proposed loss reduces the average MSE/MAE by 5.2\%/5.0\% over iTransformer across 11 datasets. It achieves 17/10 first/second places in MSE and 31/10 first/second places in MAE across 44 evaluations.
In imputation tasks  (Appendix~\ref{app:imputation}), it delivers a substantial 27.4\%/19.4\% reduction in average MSE/MAE to iTransformer across 9 datasets, and secures 13/20 first/second places in MSE and 13/8 first/second places in MAE among 36 scenarios.
These findings confirm that the supervision-level intervention not only elevates the base iTransformer, but also competes favorably against other SOTA models.

\section{Conclusion}
This work shifts the focus from network expressiveness to the fundamental training paradigm.
By abstracting the point-wise i.i.d. assumption through a rigorous information-theoretic lens, we formalize the \textbf{Optimization Bias}, unveil the \textbf{Paradigm Paradox}, derive a \textbf{unified closed-form quantification} of the linear, stationary, and Gaussian stochastic EOB as the lower bound of total EOB, and extend to the nonlinear systems with Jensen's lower bound.
These insights motivate a \textbf{Principled Debiasing Program}, which we instantiate as a harmonized structural loss.
Two synthetic data experiments verify the rationality of EOB theory, and practical experiments on forecasting and imputation tasks validate the effectiveness of debiasing program.
Ultimately, we hope to dispel \textit{the haze of AI alchemy} and nurture its sustainable growth.

\bibliographystyle{unsrt}
\bibliography{refs}

\newpage
\appendix

\section*{Appendix}
\addcontentsline{toc}{section}{Appendix}

\vspace{0.2cm}

\section*{Detailed Appendix Directory}
\addcontentsline{toc}{subsection}{Table of Appendix Contents}
\vspace{0.2em} 
\hrule height 0.5pt 
\vspace{0.5em}
\begin{itemize}

\item[] \textbf{\hyperref[app:A]{\textbf{A} \: Proof of Theorem~\ref{thm:decomp_of_EOB} (Decomposition and Lower Bound of EOB)}} \dotfill \textbf{14}
    \begin{itemize}
        \item[] \hyperref[app:A1]{A.1 \: Part 1: Proof of Lower Bound $I(z_{1:T}) \le I(x_{1:T})$} \dotfill 14
        \item[] \hyperref[app:A2]{A.2 \: Part 2: Exact Decomposition and Upper Bound of $\expt[\bias]$} \dotfill 15
    \end{itemize}

\item[] \textbf{\hyperref[app:B]{\textbf{B} \: Proof of Proposition~\ref{prop:quan_EOB_AR} (Quantification of Stochastic EOB via $AR(p)$)}} \dotfill \textbf{17}

\item[] \textbf{\hyperref[app:C]{\textbf{C} \: Relationship Between SSNR and Classical SNR}} \dotfill \textbf{19}

\item[] \textbf{\hyperref[app:D]{\textbf{D} \: Quantification of Stochastic EOB via Multivariate Gaussian Model}} \dotfill \textbf{20}
    \begin{itemize}
        \item[] \hyperref[app:D1]{D.1 \: Proof of Proposition~\ref{prop:quan_EOB_MGM} (Quantification of Stochastic EOB via MGM)} \dotfill 20
        \item[] \hyperref[app:D2]{D.2 \: Unified Expression of Stochastic EOB} \dotfill 20
        \item[] \hyperref[app:D3]{D.3 \: Technical Lemmas} \dotfill 21
    \end{itemize}

\item[] \textbf{\hyperref[app:E]{\textbf{E} \: Nonlinear Extension of Stochastic EOB}} \dotfill \textbf{23}

\item[] \textbf{\hyperref[app:F]{\textbf{F} \: Discussion}} \dotfill \textbf{25}
    \begin{itemize}
        \item[] \hyperref[app:F1]{F.1 \: Systematic Explanation of Existing Loss Studies} \dotfill 25
        \item[] \hyperref[app:F2]{F.2 \: Scope of EOB Theory} \dotfill 25
    \end{itemize}

\item[] \textbf{\hyperref[app:G]{\textbf{G} \: Characteristic Analysis of DFT, DWT, and Harmonized $\ell_p$ Norm}} \dotfill \textbf{26}
    \begin{itemize}
        \item[] \hyperref[app:G1]{G.1 \: Characteristic of Discrete Fourier Transform} \dotfill 26
        \item[] \hyperref[app:G2]{G.2 \: Characteristic of Discrete Wavelet Transform} \dotfill 26
        \item[] \hyperref[app:G3]{G.3 \: Duality of Gradient Flaws in $\ell_p$ Norms} \dotfill 27
        \item[] \hyperref[app:G4]{G.4 \: Statistical Analysis of Harmonized $\ell_p$ Norm} \dotfill 27
    \end{itemize}

\item[] \textbf{\hyperref[app:H]{\textbf{H} \: Gradient Analysis of Discrete Fourier Transform}} \dotfill \textbf{30}
    \begin{itemize}
        \item[] \hyperref[app:H1]{H.1 \: Point-wise loss function based on $\ell_2$ Norm} \dotfill 30
        \item[] \hyperref[app:H2]{H.2 \: Point-wise loss function based on $\ell_1$ Norm} \dotfill 31
    \end{itemize}
    
\item[] \textbf{\hyperref[app:I]{\textbf{I} \: Experimental Settings}} \dotfill \textbf{34}
    \begin{itemize}
        \item[] \hyperref[app:I1]{I.1 \: Synthetic Dataset Generation Mathematical Mechanism} \dotfill 34
        \item[] \hyperref[app:I2]{I.2 \: Real-world Benchmark Description} \dotfill 35
        \item[] \hyperref[app:I3]{I.3 \: Backbone Description} \dotfill 36
        \item[] \hyperref[app:I4]{I.4 \: Implementation Details} \dotfill 37
    \end{itemize}

\item[] \textbf{\hyperref[app:J]{\textbf{J} \: Extensive Experimental Analysis}} \dotfill \textbf{40}
    \begin{itemize}
        \item[] \hyperref[app:J1]{J.1 \: Empirical Verification of EOB Theory} \dotfill 40
        \item[] \hyperref[app:J2]{J.2 \: Mechanistic Insight on Trigonometric Series} \dotfill 46
        \item[] \hyperref[app:J3]{J.3 \: Plug-and-Play Versatility} \dotfill 47
        \item[] \hyperref[app:J4]{J.4 \: Long-Term Forecasting Results and Analysis} \dotfill 48
        \item[] \hyperref[app:J5]{J.5 \: Missing Data Imputation Results} \dotfill 52
        \item[] \hyperref[app:J6]{J.6 \: Ablation Experiments} \dotfill 54
    \end{itemize}

\end{itemize}

\vspace{0.3em} 
\hrule height 0.5pt 

\newpage
\section{Proof of Theorem~\ref{thm:decomp_of_EOB} (Decomposition and Lower Bound of EOB)}
\label{app:A}
\label{sec:proof_decomp_lower_bound_EOB}

The proof establishes the decomposition and lower bound for the EOB. To accommodate the potentially singular nature of deterministic components (\eg Dirac delta functions), we explicitly frame this analysis within the domain of \textbf{continuous random variables}. Accordingly, all entropy terms $H(\cdot)$ refer to \textbf{differential entropy}.

\begin{proof}

We use the following notation for differential entropy, temporal mutual information, and total correlation:
\begin{equation}
\begin{aligned}
    H(a_t)
    &= \expt[-\log p(a_t)], \\
    H(a_t \vert a_{1:t-1})
    &= \expt[-\log p(a_t \vert a_{1:t-1})], \\
    I(a_t; a_{1:t-1})
    &= \expt \left[ \log \frac{p(a_t \vert a_{1:t-1})}{p(a_t)} \right]
     = H(a_t) - H(a_t \vert a_{1:t-1}), \\
    I(a_{1:T})
    &= \sum_{t=2}^T I(a_t; a_{1:t-1}) .
\end{aligned}
\end{equation}

The proof relies on the following formalized assumptions regarding the decomposition $x_t = v_t + z_t$, together with finite differential entropy of the stochastic marginals:
\begin{tcolorbox}[tcbset]
\begin{assumption}
\label{assump:cont_det}
    \textbf{(Continuous Determinism)}
    The component $v_t$ is governed by a deterministic function of its past: $v_t = f(v_{1:t-1})$. Its conditional distribution is a Dirac delta, $p(v_t \vert v_{1: t-1}) = \delta(v_t - f(v_{1:t-1}))$, and its conditional differential entropy satisfies $H(v_t \vert v_{1:t-1}) = - \infty$.
\end{assumption}

\begin{assumption}
\label{assump:indep}
    \textbf{(Independence)}
    The component processes are independent: $\sigma(v_{1:T})$ and $\sigma(z_{1:T})$ are independent. 
    For the Gaussian processes considered in this work, this can be verified by zero cross-covariance at all lags, \ie $\cov(v_s,z_t)=0$ for all $s,t$.
\end{assumption}

\begin{assumption}
\label{assump:str_ident}
    \textbf{(Structural Identifiability)}
    The observed history $x_{1:t-1}$ is sufficient to recover the underlying components $\{v_{1:t-1}, z_{1:t-1}\}$, \ie the sigma-algebra generated by $x_{1:t-1}$ is equivalent to that of the joint pair: $\sigma(x_{1:t-1}) = \sigma(v_{1:t-1}, z_{1:t-1})$. This justifies $H(\cdot \vert x_{1:t-1}) = H(\cdot \vert v_{1:t-1}, z_{1:t-1})$.
\end{assumption}
\end{tcolorbox}

\begin{remark}
\label{remark:A3_conditions}
    \textbf{(On Assumption~\ref{assump:str_ident})}
    This is a separability condition: it holds when the deterministic and stochastic components are recoverable from the observed history, \eg through non-overlapping spectral supports or a known deterministic form. It may fail when the two components have overlapping structure, making the decomposition non-identifiable.
\end{remark}

\subsection{Part 1: Proof of Lower Bound $I(z_{1:T}) \le I(x_{1:T})$}
\label{app:A1}

To prove the lower bound, it suffices to show that $I(z_t; z_{1:t-1}) \le I(x_t; x_{1:t-1})$ for any $t \in [2, T]$.

\textbf{Step 1: Equivalence of Conditional Entropy.}
We first establish that the uncertainty of the current observation $x_t$ given its history is entirely due to the stochastic component $z_t$.
\begin{equation}
\begin{aligned}
    H(x_t \vert x_{1:t-1})
    &= H(v_t + z_t \vert x_{1:t-1})             && \text{(By Decomposition)}
    \\
    &= H(v_t + z_t \vert v_{1:t-1}, z_{1:t-1})  && \text{(By Assumption~\ref{assump:str_ident}: Identifiability)}
    \\
    &= H(z_t \vert v_{1:t-1}, z_{1:t-1})        && \text{(By Assumption~\ref{assump:cont_det}: } v_t = f(v_{1:t-1}) \text{ is a constant given } v_{1:t-1}\text{)}
    \\
    &= H(z_t \vert z_{1:t-1})                   && \text{(By Assumption~\ref{assump:indep}: Process-level Independence)}
\end{aligned}
\label{eq:equivalence_conditional_entropy_x_z}
\end{equation}
Step 3 utilizes the translation invariance of differential entropy: $H(z_t + c \vert \cdot) = H(z_t \vert \cdot)$ for any constant $c$. Step 4 removes the conditioning on $v_{1:t-1}$ because the joint independence of the two processes ensures $z_t \perp\!\!\!\perp v_{1:t-1}$.

\textbf{Step 2: Comparison of Marginal Entropy.}
We establish that adding an independent component cannot decrease differential entropy.

Applying Lemma~\ref{lem:entropy_of_sum} with $X = v_t$ and $Y = z_t$ (which are independent by Assumption~\ref{assump:indep}, and $z_t$ has finite entropy by the finite-entropy assumption on the stochastic component):
\begin{equation}
\begin{aligned}
    H(x_t)
    &= H(v_t + z_t) \\
    &\ge H(z_t)      && \text{(By Lemma~\ref{lem:entropy_of_sum} and Assumption~\ref{assump:indep})}
\end{aligned}
\end{equation}

\textbf{Step 3: Combining Results.}
Comparing the mutual information terms:
\begin{equation}
\begin{aligned}
    I(x_t; x_{1:t-1}) - I(z_t; z_{1:t-1})
    &= \left[ H(x_t) - H(x_t \vert x_{1:t-1}) \right] - \left[ H(z_t) - H(z_t \vert z_{1:t-1}) \right]
    \\
    &= H(x_t) - H(z_t)
    \\
    &\ge 0 .
\end{aligned}
\end{equation}
The second line follows because the conditional entropy terms cancel by Step 1: $H(x_t \vert x_{1:t-1}) = H(z_t \vert z_{1:t-1})$. Summing over $t$ yields $I(x_{1:T}) \ge I(z_{1:T})$.

\subsection{Part 2: Exact Decomposition and Upper Bound of $\expt[\bias]$}
\label{app:A2}

We now derive an \textbf{exact decomposition} of $\expt[\bias]$ that provides a more informative characterization.

\textbf{Step 4: Exact Decomposition.}
From the proof of Part 1, we obtained two key identities:
\begin{itemize}
    \item Step 1: $H(x_t \vert x_{1:t-1}) = H(z_t \vert z_{1:t-1})$
    \item Step 2: $H(x_t) \ge H(z_t)$
\end{itemize}

Using these, we can decompose $\expt[\bias] = I(x_{1:T})$ exactly:
\begin{equation}
\begin{aligned}
    \expt[\bias]
    &= \sum_{t=2}^{T} I(x_t; x_{1:t-1})
    = \sum_{t=2}^{T} \left[ H(x_t) - H(x_t \vert x_{1:t-1}) \right]
    \\
    &= \sum_{t=2}^{T} \left[ H(x_t) - H(z_t \vert z_{1:t-1}) \right]
    && \text{(By Step 1)}
    \\
    &= \underbrace{ \sum_{t=2}^{T} \left[ H(z_t) - H(z_t \vert z_{1:t-1}) \right] }_{= \, I(z_{1:T}) \, = \, \expt[\bias_z]}
    + \underbrace{ \sum_{t=2}^{T} \left[ H(x_t) - H(z_t) \right] }_{=: \, \Delta_v \, \ge \, 0} .
\end{aligned}
\label{eq:exact_decomp_EOB}
\end{equation}

This yields the \textbf{exact decomposition}:
\begin{equation}
    \expt[\bias] = \expt[\bias_z] + \Delta_v
\label{eq:EOB_decomposition}
\end{equation}
where $\expt[\bias_z] = I(z_{1:T})$ is the stochastic EOB (lower bound), and $\Delta_v = \sum_{t=2}^{T} [H(x_t) - H(z_t)] \ge 0$ is the \textbf{deterministic marginal entropy surplus} quantifying the additional bias introduced by the deterministic component.

\textbf{Step 5: Explicit Bounds on $\expt[\bias]$.}
Since $\Delta_v \ge 0$ (by Lemma~\ref{lem:entropy_of_sum}, each summand $H(x_t) - H(z_t) \ge 0$), the decomposition immediately yields the \textbf{lower bound}:
\begin{equation}
    \expt[\bias] \;\ge\; I(z_{1:T}) = \expt[\bias_z] .
\label{eq:lower_bound_EOB}
\end{equation}

When $z_t$ has a Gaussian marginal distribution with common variance $\sigma_z^2$ and the deterministic component $v_t$ has finite variance $\sigma_{v,t}^2$, we further obtain a \textbf{finite upper bound}. Since the Gaussian distribution maximizes differential entropy for a given variance, $H(v_t + z_t) \le \frac{1}{2}\log\bigl(2\pi e\,(\sigma_{v,t}^2 + \sigma_z^2)\bigr)$, while $H(z_t) = \frac{1}{2}\log(2\pi e\,\sigma_z^2)$. Subtracting yields $H(x_t) - H(z_t) \le \frac{1}{2}\log\!\bigl(1 + \sigma_{v,t}^2/\sigma_z^2\bigr)$, and hence the \textbf{sandwich inequality}:
\begin{equation}
    I(z_{1:T})
    \;\le\; \expt[\bias] \;\le\;
    I(z_{1:T}) + \frac{1}{2}\sum_{t=2}^{T} \log\!\left(1 + \frac{\sigma_{v,t}^2}{\sigma_z^2}\right) .
\label{eq:sandwich_EOB}
\end{equation}

The bounds exhibit the following behavior:
\begin{itemize}[leftmargin=*]
    \item \textbf{Lower bound is tight} when $v_t$ is a constant ($\sigma_v^2 = 0$): $\Delta_v = 0$ and $\expt[\bias] = I(z_{1:T})$.
    \item \textbf{Both bounds coincide} for purely stochastic processes ($v_t \equiv 0$): $\expt[\bias] = I(z_{1:T}) = \expt[\bias_z]$, reducing to the stochastic EOB that is the focus of subsequent quantification.
    \item \textbf{$\expt[\bias]$ diverges} for non-trivial deterministic processes with growing marginal variance: when $\sigma_{v,t}^2 \to \infty$ (\eg a deterministic trend $v_t = at$ with random slope), each summand $H(x_t) - H(z_t)$ grows without bound, so $\Delta_v \to \infty$ and $\expt[\bias] \to \infty$. This establishes the \textbf{Paradigm Paradox}: the more deterministic the process, the greater the optimization bias.
\end{itemize}

\end{proof}

\begin{lemma}
\label{lem:entropy_of_sum}
    \textbf{(Entropy of Independent Sum)}
    Let $X$ and $Y$ be independent continuous random variables. If $Y$ has a well-defined density $p_Y$ and finite differential entropy $H(Y) > -\infty$, then $H(X + Y) \ge H(Y)$.
\end{lemma}

\begin{proof}
By the non-negativity of mutual information~\citep{cover1999elements} and independence:
\begin{equation}
    H(X + Y) \;\ge\; H(X + Y \,\vert\, X) \;=\; H(Y \,\vert\, X) \;=\; H(Y)
\end{equation}
where the inequality is $I(X+Y;\, X) \ge 0$, the first equality uses translation invariance of differential entropy ($H(Y + c \,\vert\, \cdot) = H(Y \,\vert\, \cdot)$ for any value $c$ measurable \wrt the conditioning), and the second equality uses independence ($Y \perp\!\!\!\perp X$). This argument requires no regularity on $X$: it holds whether $X$ has a density, is discrete, or is degenerate. 

\end{proof}

\newpage
\section{Proof of Proposition~\ref{prop:quan_EOB_AR} (Quantification of Stochastic EOB via $AR(p)$)}
\label{app:B}
\label{app:quan_EOB_AR}

To strictly quantify the stochastic EOB for a covariance-stationary Gaussian $AR(p)$ process with $T>p$, we must account for the dependence structure. 
The exact conditional distribution $p(z_t \vert z_{1:t-1})$ varies depending on whether $t \leq p$ (transient phase) or $t > p$ (steady-state phase).

\begin{proof}
We decompose the stochastic EOB into two components: initial transient and steady-state terms:
\begin{equation}
    \bias_z
    = \sum_{t=2}^{T} \log \frac{p(z_t \vert z_{1:t-1})}{p(z_t)}
    = \underbrace{ 
        \sum_{t=2}^{p} \log \frac{ p(z_t \vert z_{1:t-1}) }{ p(z_t) }
    }_{\text{Transient Term } (\bias_{\text{tran}})}
    + \underbrace{ 
        \sum_{t=p+1}^{T} \log \frac{ p(z_t \vert z_{1:t-1}) }{ p(z_t) }
    }_{\text{Steady-State Term } (\bias_{\text{ss}})}
\end{equation}

\paragraph{1. Steady-State Phase ($t > p$):} For $t>p$, the conditional distribution is fully determined by the $p$ preceding observation. Due to the Markov property and stationarity:
\begin{itemize}[leftmargin=*]
    \item \textbf{Conditional Distribution}: 
    \begin{equation}
        z_{t} | z_{1:t-1} \sim \mathcal{N}(\mu_t, \sigma_{\epsilon}^2)
    \end{equation}
    where $\mu_t = c + \sum_{i=1}^{p} \phi_i z_{t-i}$. Note that the conditional variance stabilizes to the innovation variance $\sigma_{\epsilon}^2$.
    
    \item \textbf{Marginal Distribution}: 
    \begin{equation}
        z_{t} \sim \mathcal{N}(\mu, \sigma_z^2), \quad \text{where} \; \mu = \frac{c}{1 - \sum_{i=1}^{p} \phi_i} .
    \end{equation}
    The marginal variance, $\sigma_z^2 = \var(z_t)$, is determined by the model's parameters $(\phi_i, \sigma_{\epsilon}^2)$ via the \textit{Yule-Walker Equations}:
    \begin{equation}
        \sigma_z^2 = \frac{\sigma_{\epsilon}^2}{1 - \sum_{i=1}^{p} \phi_i \rho_i}
    \end{equation}
    where $\rho_i = \gamma_i / \gamma_0$ is the autocorrelation function at lag $i$ and depends on the $\phi_i$ in a complex way.
\end{itemize}

Substituting these into the steady-state sum $\bias_{\text{ss}}$:
\begin{equation} 
\begin{aligned} 
    \bias_{\text{ss}} 
    &= \sum_{t=p+1}^{T} \left[ 
        \frac{1}{2} \log \frac{\sigma_z^2}{\sigma_\epsilon^2} 
        - \frac{(z_{t} - \mu_t)^2}{2\sigma_\epsilon^2} 
        + \frac{(z_{t} - \mu)^2}{2\sigma_z^2}
    \right] . 
\end{aligned} 
\end{equation}

Taking the expectation $\expt[\bias_{\text{ss}}]$, we apply the law of iterated expectations: $\expt[(z_t - \mu_t)^2] = \expt[\expt[(z_t - \mu_t)^2 \mid z_{1:t-1}]] = \expt[\sigma_{\epsilon}^2] = \sigma_{\epsilon}^2$, and $\expt[(z_t - \mu)^2] = \sigma_z^2$. Thus the fluctuation terms cancel, yielding:
\begin{equation}
    \expt[\bias_{\text{ss}}]
    = \frac{T-p}{2} \log \frac{\sigma_z^2}{\sigma_{\epsilon}^2} .
\end{equation}

\paragraph{2. Transient Phase ($2 \le t \le p$):}
The first $p$ observations form a transient segment because the observed history is shorter than the full autoregressive context.
Let
\begin{equation}
    \sigma_t^2 = \var(z_t \vert z_{1:t-1}), \qquad 2 \le t \le p .
\end{equation}
These conditional variances are not generally equal to the innovation variance $\sigma_\epsilon^2$, but they are bounded between the innovation variance and the marginal variance:
\begin{equation}
    \sigma_\epsilon^2 \le \sigma_t^2 \le \sigma_z^2 .
\end{equation}
The upper bound follows because conditioning cannot increase variance.
The lower bound follows because conditioning on fewer than $p$ lagged states cannot reduce uncertainty below the full-information one-step prediction error.
Formally, under a stationary extension with the missing pre-sample states $z_{t-p},\dots,z_0$,
\begin{equation}
    \var(z_t \mid z_{1:t-1})
    \ge
    \var(z_t \mid z_{1:t-1}, z_{t-p},\dots,z_0)
    =
    \sigma_\epsilon^2 .
\end{equation}

Therefore, the transient contribution is a finite correction depending only on the initial $p$-dimensional correlation structure:
\begin{equation}
    \expt[\bias_{\text{tran}}]
    = \sum_{t=2}^p \frac{1}{2} \log \frac{\sigma_z^2}{\sigma_t^2}
    = C_p .
\end{equation}
This correction is independent of $T$.
Each $\sigma_t^2$ can be computed via the Schur complement of the leading $(t{-}1)\times(t{-}1)$ block of the Toeplitz covariance matrix.
Equivalently, the chain rule of conditional variances gives $\prod_{t=1}^{p}\sigma_t^2 = \det(\Sigma_{1:p})$ with $\sigma_1^2 = \sigma_z^2$.
Thus,
\begin{equation}
    C_p
    = \frac{1}{2}\log \frac{\sigma_z^{2p}}{\det(\Sigma_{1:p})}
    = -\frac{1}{2}\log \vert R_p \vert
\end{equation}
where $R_p$ is the $p \times p$ correlation matrix of $(z_1,\dots,z_p)$, and the last equality uses $\Sigma_{1:p}=\sigma_z^2 R_p$, since stationarity gives the same marginal variance $\sigma_z^2$ at each timestamp. This is consistent with the correction term derived from the MGM perspective in Appendix~\ref{app:D}.

\end{proof}

\paragraph{Stochastic EOB:} Combining both parts, the stochastic EOB is: 
\begin{equation} 
    \expt[\bias_z] 
    = \frac{T-p}{2} \log \left( \frac{\sigma_z^2}{\sigma_{\epsilon}^2} \right) + C_p
    = \frac{T-p}{2} \log \left( \frac{1}{1 - \sum_{i=1}^{p} \phi_i \rho_i} \right) + C_p . 
\end{equation}

Since $C_p$ is constant with respect to $T$, for large sequence lengths ($T \gg p$), the stochastic EOB is dominated by the steady-state linear growth. Moreover, using the bounds $\sigma_{\epsilon}^2 \le \sigma_t^2 \le \sigma_z^2$, we obtain the upper bound:
\begin{equation}
    C_p \le \frac{p-1}{2}\log\frac{\sigma_z^2}{\sigma_{\epsilon}^2} = \frac{p-1}{2}\log(\ssnr),
\end{equation}
which confirms that, for $\ssnr>1$, the initial correction is asymptotically negligible relative to the accumulated term as $T \to \infty$. Thus, we focus on the asymptotic rate:
\begin{equation}
    \expt[\bias_z] 
    \approx \frac{T}{2} \log \frac{\sigma_z^2}{\sigma_\epsilon^2} \quad     \text{as } T \to \infty.
\end{equation}

\newpage
\section{Relationship Between SSNR and Classical SNR}
\label{app:C}
\label{app:relationship_snr}
This appendix clarifies the relationship between the Structural Signal-to-Noise Ratio (SSNR), as defined in our main analysis, and the classical Signal-to-Noise Ratio (SNR) widely used in information theory and signal processing.

The classical SNR quantifies the ratio of the power of a signal to the power of confounding noise. 
For explanatory purposes, consider an additive predictive decomposition:
\begin{equation}
    z_t = \hat{z}_t + \epsilon_t
\end{equation}
where $\hat{z}_t=\expt[z_t\mid z_{<t}]$ is the predictable component and $\epsilon_t=z_t-\expt[z_t\mid z_{<t}]$ is the innovation.
If the predictable component and the innovation are uncorrelated, the classical SNR is written as:
\begin{equation}
    \textit{SNR} 
    = \frac{P_{\text{signal}}}{P_{\text{noise}}} 
    = \frac{\hat{\sigma}^2}{\sigma_{\epsilon}^2}
\end{equation}
where $\hat{\sigma}^2=\var(\hat{z}_t)$ and $\sigma_\epsilon^2=\var(\epsilon_t)$.

This orthogonality gives the variance decomposition
\begin{equation}
\begin{aligned}
    \sigma_z^2 
    &= \var(z_t)
    \\
    &= \var(\hat{z}_t) + 2 \cov(\hat{z}_t,\epsilon_t) + \var(\epsilon_t)
    \\
    &= \hat{\sigma}^2 + \sigma_{\epsilon}^2 .
\end{aligned}
\end{equation}

Our SSNR is defined as the ratio of the total variance of the stochastic process to the variance of its innovation,
$\ssnr = \sigma_z^2 / \sigma_{\epsilon}^2$.
Using the variance decomposition above, we obtain the bridge
\begin{equation}
    \ssnr
    = \frac{\hat{\sigma}^2}{\sigma_{\epsilon}^2} + \frac{\sigma_{\epsilon}^2}{\sigma_{\epsilon}^2}
    = \textit{SNR} + 1 .
\end{equation}
Thus, under this predictive additive interpretation, SSNR equals the corresponding classical SNR plus one.
The offset arises because SSNR uses the total process variance in the numerator, whereas classical SNR uses only the predictable signal variance.

\newpage
\section{Quantification of Stochastic EOB via Multivariate Gaussian Model}
\label{app:D}

\subsection{Proof of Proposition~\ref{prop:quan_EOB_MGM} (Quantification of Stochastic EOB via MGM)}
\label{app:D1}
\label{app:quan_EOB_MGM}
The main text first quantifies stochastic EOB through a parametric linear model, the $AR(p)$ process. Here we show that the same finite-window quantity can be characterized without specifying autoregressive transitions.

\begin{proof}

Let the $T$-dimensional stochastic sequence $\mathrm{z}$ be modeled as a Gaussian vector, $\mathrm{z} \sim \mathcal{N}(\boldsymbol{\mu}, \Sigma)$. This multivariate Gaussian view places no autoregressive structure on the covariance matrix $\Sigma$.

The stochastic bias $\bias_z$ is the log-likelihood divergence between the true joint distribution $p(z_{1:T}) = p(z_1) \cdot \prod_{t=2}^T p(z_t \vert z_{1:t-1})$ and the factorized surrogate $q(z_{1:T}) = \prod_{t=1}^T p(z_t)$ induced by point-wise supervision:
\begin{equation}
    \bias_z
    = \sum_{t=2}^{T} \log \frac{p(z_t | z_{1:t-1})}{p(z_t)}
    = \log \frac{p(z_{1:T})}{q(z_{1:T})}
    = \log \frac{p(z_{1:T})}{\prod_{t=1}^T p(z_t)}
    = \log p(z_{1:T}) - \sum_{t=1}^T \log p(z_t) .
\end{equation}

By substituting the log-PDFs for the multivariate and univariate Gaussian distributions, we get:
\begin{gather}
    \log p(z_{1:T}) 
    = - \frac{T}{2} \cdot \log(2 \pi) 
      - \frac{1}{2} \log \vert \Sigma \vert 
      - \frac{1}{2} (\mathrm{z} - \boldsymbol{\mu})^{\top} \Sigma^{-1} (\mathrm{z} - \boldsymbol{\mu})
    \\
    \log p(z_t)
    = - \frac{1}{2} \cdot \log(2 \pi) 
      - \frac{1}{2} \log \sigma_t^2 
      - \frac{(z_t-\mu_t)^2}{2\sigma_t^2}
    \\
    \bias_z
    = -\frac{1}{2} \log \vert \Sigma \vert 
      - \frac{1}{2} (\mathrm{z} - \boldsymbol{\mu})^{\top} \Sigma^{-1} (\mathrm{z} - \boldsymbol{\mu}) 
      + \frac{1}{2} \sum_{t=1}^T \log\sigma_t^2 
      + \frac{1}{2} \sum_{t=1}^T \frac{(z_t-\mu_t)^2}{\sigma_t^2} .
\end{gather}

Taking the expectation and using the standard results that $\expt [(\mathrm{z}-\boldsymbol{\mu})^{\top} \Sigma^{-1} (\mathrm{z}-\boldsymbol{\mu}) ] = T$ and $\expt [(z_t-\mu_t)^2] = \sigma_t^2$, the quadratic terms cancel, leaving
\begin{equation}
    \expt[\bias_z] 
    = - \frac{1}{2} \left( \log \vert \Sigma \vert - \sum_{t=1}^T \log\sigma_t^2 \right)
    = \frac{1}{2} \log \left( \frac{\prod_{t=1}^T \sigma_t^2}{\vert \Sigma \vert} \right) .
\end{equation}

To reveal the core insight, we introduce the correlation matrix $R$ with elements $R_{ij} = \Sigma_{ij}/(\sigma_i \sigma_j)$, defined by the relationship $\Sigma = D^{1/2} R D^{1/2}$, where $D = \diag(\sigma_1^2, \dots, \sigma_T^2)$ is the diagonal matrix of variances. Taking the determinant, we have:
\begin{equation}
    \vert \Sigma \vert 
    = \vert D \vert \vert R \vert 
    = \left( \prod_{t=1}^T \sigma_t^2 \right) \vert R \vert .
\end{equation}

Substituting this into the expression for $\expt[\bias_z]$ yields:
\begin{equation}
    \expt[\bias_z]
    = \frac{1}{2} \log \left( \frac{\prod_{t=1}^T \sigma_t^2}{(\prod_{t=1}^T \sigma_t^2) \vert R \vert} \right)
    = - \frac{1}{2} \log \vert R \vert .
    \label{eq:EOB_MGM_app}
\end{equation}

\end{proof}

\noindent Since $\Sigma$ is positive definite, so is $R$, and $\vert R \vert \in (0,1]$ for a correlation matrix. Eq.~\eqref{eq:EOB_MGM_app} is therefore exactly the Gaussian total correlation discarded by the point-wise factorized surrogate: it is zero under complete independence and increases as finite-window linear dependence strengthens.

\subsection{Unified Expression of Stochastic EOB}
\label{app:D2}
\label{app:link_AR_MGM}
Eq.~\eqref{eq:EOB_MGM_app} packages the effects of both sequence length and structural predictability into the determinant of the finite-window correlation matrix. The sequence length enters through the size of $R$, while the predictability of the process controls how quickly $\vert R \vert$ decays as the window grows.

\begin{proof}
For a covariance-stationary process, where $\sigma_t^2=\sigma_z^2$ for all $t$, the same formula can be written as:
\begin{equation}
    \expt [\bias_z] 
    = \frac{T}{2} \log \frac{\sigma_z^2}{\vert \Sigma \vert^{1/T}} .
\end{equation}
Under the conditions of Lemma~\ref{lem:convergence_R}, this gives the asymptotic rate:
\begin{equation}
    \lim_{T\to\infty} \frac{1}{T}\expt[\bias_z]
    = -\frac{1}{2}\lim_{T\to\infty}\log \vert R \vert^{1/T}
    = \frac{1}{2}\log(\ssnr).
\end{equation}

For a finite sequence of length $T$ from a stationary $AR(p)$ process, Lemma~\ref{lem:decomp_R_p} gives the exact determinant identity
\begin{equation}
    \vert R \vert 
    = \vert R_p \vert \cdot \left( \frac{\sigma_{\epsilon}^2}{\sigma_z^2} \right)^{T-p}
    = \vert R_p \vert \cdot \left( \frac{1}{\ssnr} \right)^{T-p} .
\end{equation}
Substituting this identity into Eq.~\eqref{eq:EOB_MGM_app} recovers the exact $AR(p)$ stochastic EOB:
\begin{equation}
\begin{aligned}
    \expt[\bias_z] 
    &= -\frac{1}{2} \log \vert R \vert
    = -\frac{1}{2} \log \left[ \vert R_p \vert \cdot \left( \frac{1}{\ssnr} \right)^{T-p}\right]
    \\
    &= \frac{T-p}{2} \log ( \ssnr ) - \frac{1}{2} \log \vert R_p \vert .
\end{aligned}
\end{equation}

Thus the $AR(p)$ expression contains a primary accumulated term, $\frac{T-p}{2}\log(\ssnr)$, and an initial-state correction, $-\frac{1}{2}\log\vert R_p\vert$. Equivalently,
\begin{equation}
    \expt[\bias_z]
    = \frac{T}{2} \log ( \ssnr ) + C(p, \boldsymbol{\phi})
\end{equation}
where $C(p, \boldsymbol{\phi})$ is a constant that depends on the model order $p$ and the AR coefficients $\boldsymbol{\phi} = (\phi_1, \dots, \phi_p)$ (through $\vert R_p \vert$ and the SSNR), but is independent of the sequence length $T$. 
This confirms that stochastic EOB is not an artifact of the $AR(p)$ parameterization; the AR formula is the structured special case of the Gaussian finite-window total correlation.

\end{proof}

\subsection{Technical Lemmas}
\label{app:D3}
\begin{tcolorbox}[tcbset]
\begin{lemma}
\label{lem:convergence_R}
    \textbf{(The Asymptotic Convergence of Autocorrelation Matrix Determinant Geometric Mean)}
    For a purely nondeterministic covariance-stationary Gaussian process with log-integrable power spectral density $f(\omega)$, let $R$ be its $T \times T$ correlation matrix. As the sequence length $T$ approaches infinity, the geometric mean of the eigenvalues of $R$, represented by $\vert R \vert^{1/T}$, converges to the reciprocal of the SSNR:
    \begin{equation}
        \lim_{T \to \infty} \vert R \vert^{1/T}
        = \frac{\sigma_{\epsilon}^2}{\sigma_z^2}
        = \frac{1}{\ssnr} .
    \end{equation}
\end{lemma}
\end{tcolorbox}

\begin{proof}

The proof utilizes \textit{Szeg\H{o}'s First Limit Theorem}, which relates the asymptotic determinant of a Toeplitz matrix to the geometric mean of its generating function. We use the spectral convention $\gamma_0=\sigma_z^2=\frac{1}{2\pi}\int_{-\pi}^{\pi} f(\omega)\,\diff \omega$. For the correlation matrix $R$, the generating function is the \textbf{normalized} power spectral density, $g(\omega) = f(\omega)/\sigma_z^2$.

The theorem states:
\begin{equation}
    \lim_{T \to \infty} \frac{1}{T} \log \vert R \vert 
    = \frac{1}{2\pi} \int_{-\pi}^{\pi} \log \left( \frac{f(\omega)}{\sigma_z^2} \right) \diff \omega .
\end{equation}

By \textit{Kolmogorov's Formula}, the geometric mean of the process PSD $f(\omega)$ is equal to the innovation variance $\sigma_{\epsilon}^2$:
\begin{equation}
    \exp\left( \frac{1}{2\pi} \int_{-\pi}^{\pi} \log f(\omega) \diff \omega \right) = \sigma_{\epsilon}^2.
\end{equation}

Substituting this into the limit equation:
\begin{equation}
\begin{aligned}
    \lim_{T \to \infty} \vert R \vert^{1/T} 
    &= \exp\left( \frac{1}{2\pi} \int_{-\pi}^{\pi} \left( \log f(\omega) - \log \sigma_z^2 \right) \diff \omega \right)
    \\
    &= \frac{\exp\left( \frac{1}{2\pi} \int_{-\pi}^{\pi} \log f(\omega) \diff \omega \right)}{\sigma_z^2}
    \\
    &= \frac{\sigma_{\epsilon}^2}{\sigma_z^2}.
\end{aligned}
\end{equation}

This confirms the relationship with the Signal-to-Noise Ratio (SSNR):
\begin{equation}
    \lim_{T \to \infty} \vert R \vert^{1/T} 
    = \frac{1}{\ssnr}.
\end{equation}

\end{proof}

\begin{tcolorbox}[tcbset]
\begin{lemma}
\label{lem:decomp_R_p}
    \textbf{(Determinant Decomposition for $AR(p)$ Correlation Matrix)} 
    Let $\{z_t\}$ be a stationary AR(p) process with unconditional variance $\sigma_z^2$ and innovation variance $\sigma_\epsilon^2$. 
    For a sequence of length $T > p$, let $R$ denote the $T \times T$ correlation matrix of the vector $(z_1, \dots, z_T)$ and $R_p$ be the $p \times p$ correlation matrix of the initial $p$ observations $(z_1, \dots, z_p)$. 
    The determinant of the full correlation matrix is given by:
    \begin{equation}
        \vert R \vert 
        = \vert R_p \vert \cdot \left( \frac{\sigma_{\epsilon}^2}{\sigma_z^2} \right)^{T-p} .
    \end{equation}
\end{lemma}
\end{tcolorbox}

\begin{proof}

Let $\Sigma_t$ denote the covariance matrix of $(z_1,\dots,z_t)$. For $t>p$, partition $\Sigma_t$ as
\begin{equation}
    \Sigma_t
    =
    \begin{bmatrix}
        \Sigma_{t-1} & \boldsymbol{r}_t \\
        \boldsymbol{r}_t^\top & \sigma_z^2
    \end{bmatrix}.
\end{equation}
By the Schur complement determinant identity,
\begin{equation}
    \vert \Sigma_t \vert
    =
    \vert \Sigma_{t-1} \vert
    \left(
        \sigma_z^2
        - \boldsymbol{r}_t^\top \Sigma_{t-1}^{-1}\boldsymbol{r}_t
    \right).
\end{equation}
In the Gaussian AR setting, this Schur complement is exactly the conditional variance,
\begin{equation}
    \sigma_z^2
    - \boldsymbol{r}_t^\top \Sigma_{t-1}^{-1}\boldsymbol{r}_t
    =
    \var(z_t \vert z_{1:t-1}).
\end{equation}
The $AR(p)$ Markov property and the innovation equation imply
\begin{equation}
    \var(z_t \vert z_{1:t-1})
    = \var(z_t \vert z_{t-p:t-1})
    = \sigma_{\epsilon}^2,
    \qquad t=p+1,\dots,T.
\end{equation}
Therefore,
\begin{equation}
    \vert \Sigma_t \vert
    = \vert \Sigma_{t-1} \vert \cdot \sigma_{\epsilon}^2,
    \qquad t=p+1,\dots,T.
\end{equation}
Iterating this algebraic identity from $t=p+1$ to $T$ gives
\begin{equation}
    \vert \Sigma \vert = \vert \Sigma_p \vert \cdot (\sigma_{\epsilon}^2)^{T-p} .
\end{equation}

For a stationary process, the marginal variance $\sigma_z^2$ is constant. The covariance matrix $\Sigma$ and correlation matrix $R$ are related by $\Sigma = \sigma_z^2 R$. Using the determinant property $\vert \lambda A \vert = (\lambda)^k \vert A \vert$ for a $k \times k$ matrix $A$, we have
\begin{equation}
\begin{dcases}
    \vert \Sigma \vert = (\sigma_z^2)^T \vert R \vert
    \\
    \vert \Sigma_p \vert = (\sigma_z^2)^p \vert R_p \vert .
\end{dcases}
\end{equation}

Substituting these into the covariance identity:
\begin{equation}
    \vert R \vert = \vert R_p \vert \cdot \left( \frac{\sigma_{\epsilon}^2}{\sigma_z^2} \right)^{T-p} .
\end{equation}

This can also be expressed using the SSNR ($\ssnr = \sigma_z^2 / \sigma_{\epsilon}^2$):
\begin{equation}
    \vert R \vert = \vert R_p \vert \cdot \left( \frac{1}{\ssnr} \right)^{T-p} .
\end{equation}

\end{proof}

\newpage
\section{Nonlinear Extension of Stochastic EOB}
\label{app:E}
\label{app:GMM_modeling}

To extend our analysis beyond the linear domain, we use the \textbf{Gaussian Mixture Model (GMM)} to decompose a nonlinear stochastic process into several local covariance-stationary Gaussian regimes. 

\begin{proof}
Let $c \in \{1,\dots,K\}$ be a latent regime variable with $\Pr(c=k)=\pi_k$. Conditional on regime $k$, the finite window follows
\begin{equation}
    z_{1:T} \mid c=k
    \sim p_k(z_{1:T})
    = \mathcal{N}(\boldsymbol{\mu}_k,\Sigma_k)
\end{equation}
where each component regime is covariance-stationary, meaning that its mean is constant and its autocovariance depends only on the time lag. The resulting mixture distribution is
\begin{equation}
    p(z_{1:T}) = \sum_{k=1}^K \pi_k p_k(z_{1:T}),
    \qquad
    p(z_t) = \sum_{k=1}^K \pi_k p_{k,t}(z_t),
\end{equation}
where $p_{k,t}$ is the marginal density of timestamp $t$ under component $k$. The mixture weights are time-invariant, so the same regime prior is used for both the full window and its marginals.

Directly computing the stochastic EOB for a GMM is analytically intractable due to the logarithm-of-sums form, $\log \left( \sum_{k=1}^K \pi_k p_k(z_{1:T}) \right)$, in its log-likelihood. We therefore reformulate the EOB using differential entropy, $H(X)=\expt[-\log p(X)]$.

The stochastic EOB can be written as the total correlation of the finite window:
\begin{equation}
\begin{aligned}
    \expt[\bias_z] 
    &= \expt_{p} \left[ \log \frac{p(z_{1:T})}{\prod_{t=1}^T p(z_t)} \right]
     = \expt_{p}[\log p(z_{1:T})] - \sum_{t=1}^T \expt_{p}[\log p(z_t)]
    \\
    &= - H(z_{1:T}) + \sum_{t=1}^T H(z_t) .
\end{aligned}
\end{equation}

While the differential entropy of GMM lacks a closed-form expression, it admits Jensen-type bounds:
\begin{equation}
    \sum_{k=1}^K \pi_k H(X_k)
    \le H(X)
    \le \sum_{k=1}^K \pi_k H(X_k) + H(\pi)
\end{equation}
where $X \sim \sum_k \pi_k p_k$, $X_k \sim p_k$, and $H(\pi) = - \sum_{k=1}^K \pi_k \log \pi_k$ is the Shannon entropy of the mixture weights.

Applying these bounds to the joint window and to each marginal gives
\begin{equation}
\begin{dcases}
    -H(z_{1:T}) 
    \ge - \sum_{k=1}^K \pi_k H(z_{k, 1:T}) - H(\pi)
    \\
    H(z_t) 
    \ge \sum_{k=1}^K \pi_k H(z_{k,t}) .
\end{dcases}
\end{equation}

Substituting these bounds into the EOB expression and regrouping terms by component yields
\begin{equation}
\begin{aligned}
    \expt[\bias_z]
    &\ge \sum_{k=1}^K \pi_k \left[ \sum_{t=1}^T H(z_{k, t}) - H(z_{k, 1:T}) \right] - H(\pi) 
    \\
    &= \sum_{k=1}^K \pi_k
    \expt_{p_k} \left[
        \log \frac{p_k(z_{1:T})}{\prod_{t=1}^T p_{k,t}(z_t)}
    \right] - H(\pi)
    \\
    &= \sum_{k=1}^K \pi_k \expt[\bias_k] - H(\pi) .
\end{aligned}
\end{equation}
Here $\expt[\bias_k]$ denotes the stochastic EOB of the $k$-th covariance-stationary Gaussian component, quantified by its component-level SSNR through Eq.~\eqref{eq:EOB_general}.

\paragraph{Tightness of the bound.}
The gap between the true stochastic EOB and the lower bound is exactly the sum of the entropy-bound gaps:
\begin{equation}
\begin{aligned}
    \expt[\bias_z]
    - \left( \sum_{k=1}^K \pi_k \expt[\bias_k] - H(\pi) \right)
    &=
    \sum_{t=1}^T
    \left[
        H(z_t) - \sum_{k=1}^K \pi_k H(z_{k,t})
    \right]
    \\
    &\quad+
    \left[
        \sum_{k=1}^K \pi_k H(z_{k,1:T}) + H(\pi) - H(z_{1:T})
    \right].
\end{aligned}
\end{equation}

\end{proof}

Thus the bound is tight when the Jensen gaps for both the marginal distributions and the full-window distribution are small. This condition depends on how the latent regime is expressed at the marginal and joint levels, rather than on pairwise component separation alone.

\newpage
\section{Discussion}
\label{app:F}
\label{app:discussion}

\subsection{Systematic Explanation of Existing Loss Studies}
\label{app:F1}

The EOB perspective clarifies which kinds of methods can actually mitigate the bias induced by point-wise supervision. 
The key distinction is whether a method changes the \textit{final supervised target and loss} or only changes the \textit{model representation}. 
A decomposition module~\citep{11314178}, frequency block~\citep{10.1145/3711896.3736952, xu2024fits}, unsupervised pre-training~\citep{10600455, 10496248}, or transformed input~\citep{11169478} used inside the architecture does not reduce EOB if the final constraint is still temporal point-wise loss function on the original sequence. 
In that case, the induced surrogate remains the same factorized law over the temporal target, and the EOB is unchanged. 
Such methods may improve expressiveness or optimization efficiency, but they do not remove the objective-level mismatch analyzed in this work.

\paragraph{Auxiliary Statistical Constraints.}
Methods such as PMLF~\citep{chen2026pmlf}, Loss Shaping Constraints~\citep{pmlr-v235-hounie24a}, PSLoss~\citep{pmlr-v267-kudrat25a}, DBLoss~\citep{qiu2025DBLoss}, and DistDF~\citep{wang2026distdf} keep the temporal prediction target but add auxiliary structural constraints beyond point-wise errors. 
Loss shaping constrains step-wise errors to shape their distribution across the forecasting window.
PSLoss constrains patch-level correlation, variance, and mean; DBLoss separately constrains trend and seasonal components within the forecasting horizon; DistDF adds a joint-distribution Wasserstein constraint to align forecast and label distributions. 
From the EOB Theory view, these methods work because they reintroduce statistical information discarded by the factorized point-wise surrogate. 
They do not eliminate the full temporal correction term, but they partially compensate for it via horizon-wise constraints, explicit statistics, or distributional moments.

\paragraph{Correlation-decoupled Objectives.}
Methods such as FreDF~\citep{wang2025fredf}, Time-o1~\citep{wang2025timeo1}, and QDF~\citep{wang2026quadratic} instead redesign the supervised sequence or error geometry to decouple temporal correlations. 
FreDF aligns prediction and label sequences in the frequency domain; Time-o1 learns transformed label components with weaker dependence and ordered significance; QDF replaces identity-weighted MSE with a learned quadratic-form objective whose off-diagonal terms account for label autocorrelation. 
From the EOB Theory view, these methods work by reducing the dependence that the point-wise surrogate must ignore, or by replacing the identity factorization with a correlation-aware error geometry. 
This explains why correcting label correlation or reducing multi-task coupling improves performance: label correlation is a manifestation of SSNR, while multi-task difficulty reflects the accumulation of EOB with optimization length.

\paragraph{Implication.}
This classification shows that existing loss studies are not isolated heuristics. 
They work when they either add back structural information discarded by the point-wise factorization or change the supervised target so that less temporal dependence is discarded in the first place. 
Our EOB theory makes this principle explicit by identifying the factorized surrogate as the source of the bias and by quantifying its intrinsic drivers through SSNR and optimization length.

\subsection{Scope of EOB Theory}
\label{app:F2}
The definition of EOB is distributional and does not rely on linearity, Gaussianity, or stationarity: it is the KL gap between the true sequential law and the factorized surrogate induced by point-wise supervision. 
These assumptions are only required for the closed-form quantification in Theorem~\ref{thm:quan_EOB_general}.

Covariance stationarity is adopted by design to isolate the bias induced by point-wise loss from confounding distribution shift. 
Gaussianity provides a maximum-entropy analytical baseline under fixed second-order statistics, making the result conservative with respect to additional non-Gaussian structure. 
Linearity gives the cleanest analytical foundation, while the GMM lower bound in Section~\ref{sec:non_linear_modeling} and Eq.~\eqref{eq:EOB_GMM} extends the same logic to nonlinear mixture regimes.

Therefore, the exact formula requires the linear, Gaussian, and covariance-stationary setting, but the qualitative insight is broader: stronger temporal predictability and longer optimization horizons increase the bias induced by point-wise factorization. 
This trend is empirically validated across architectures and non-Gaussian innovation distributions in Section~\ref{sec:EOB_verif} and Appendix~\ref{app:J1}.

\newpage
\section{Characteristic Analysis of DFT, DWT, and Harmonized $\ell_p$ Norm}
\label{app:G}

\subsection{Characteristic of Discrete Fourier Transform}
\label{app:G1}
\label{app:adv_analysis_DFT}

To instantiate structural orthogonalization, we select the Discrete Fourier Transform (DFT) as a principled and effective implementation. 
For a time series sample $\boldsymbol{x} \in \mathbb{R}^{L}$, the full DFT is a linear projection via a unitary matrix $\mathbf{U} \in \mathbb{C}^{L \times L}$ that maps the signal to its complex spectrum $\boldsymbol{f} \in \mathbb{C}^{L}$
\begin{equation}
    \boldsymbol{f} = \mathbf{U}\boldsymbol{x}.
\end{equation}
where the element of $\mathbf{U}$ is given by $u_{k, l} = \frac{1}{\sqrt{L}} \exp(-j 2\pi k l /L)$. 

As a linear, differentiable, and unitary operation, its inverse is simply its conjugate transpose, ensuring a well-defined gradient for backpropagation:
\begin{equation}
    \frac{\partial \mcL}{\partial \boldsymbol{x}} = \mathbf{U}^{\herm} \frac{\partial \mcL}{\partial \boldsymbol{f}} .
\end{equation}

The DFT is suited to our framework for two key advantages that directly align with structural orthogonalization:
\begin{itemize}
    \item \textbf{Approximate Orthogonality Decomposition:}
    The DFT asymptotically decorrelates many covariance-stationary processes into approximately uncorrelated frequency components. 
    For finite windows, this transformation reduces the effective SSNR seen by the point-wise surrogate, making the transformed-domain objective better aligned with the sequential law.

    \item \textbf{Exceptional Computational Efficiency:} 
    The DFT and its inverse are computed with remarkable efficiency using the Fast Fourier Transform (FFT) algorithm, which has a low computational complexity of $\mathcal{O}(L \log L)$. This makes the entire debiasing framework practical even for very long sequences.
\end{itemize}

In summary, the DFT provides a canonical way to target the structural driver of EOB by reducing effective temporal correlation. 
While the DFT is a natural choice for global periodic structure, other orthogonal transforms such as wavelet or Chebyshev transforms also represent promising alternatives.

\subsection{Characteristic of Discrete Wavelet Transform}
\label{app:G2}
\label{app:adv_analysis_DWT}

Discrete Wavelet Transform (DWT) is a localized implementation of structural orthogonalization. 
For a given time series sample $\boldsymbol{x} \in \mathbb{R}^{L}$, the DWT performs a multi-resolution analysis via an orthogonal matrix $\mathbf{W} \in \mathbb{R}^{L \times L}$ that maps the signal to a set of wavelet coefficients $\boldsymbol{w} \in \mathbb{R}^{L}$:
\begin{equation}
    \boldsymbol{w} = \mathbf{W}\boldsymbol{x}
\end{equation}
where $\mathbf{W}$ is constructed from a chosen mother wavelet (\eg Haar, Daubechies), and satisfies $\mathbf{W}^{\top} \mathbf{W} = I$. 

As an orthogonal operation, its inverse is its transpose ($\mathbf{W}^{\top}$), which facilitates seamless backpropagation during training:
\begin{equation}
    \frac{\partial \mcL}{\partial \boldsymbol{x}} = \mathbf{W}^{\top} \frac{\partial \mcL}{\partial \boldsymbol{w}}
\end{equation}

The DWT serves as a practical vehicle for our debiasing framework based on the following structural advantages:
\begin{itemize}
    \item \textbf{Multi-scale Orthogonal Decomposition:}
    The DWT decomposes a time series into coefficients indexed by scale and shift. 
    Unlike the DFT, which uses global trigonometric functions, the DWT uses localized basis functions. 
    This can reduce local and multi-scale dependence in the supervised target, thereby lowering the effective SSNR faced by point-wise optimization.

    \item \textbf{Capture of Non-Stationary Dynamics:}
    Time series data often exhibit non-stationary behavior where statistical properties change over time. The DWT provides time-frequency localization, allowing the debiasing program to isolate bias not just in frequency bands, but also in specific temporal windows. This makes the framework more robust to structural breaks or transient patterns that a global DFT might smooth over.

    \item \textbf{Linear Computational Complexity:}
    The DWT and its inverse can be computed with extreme efficiency using the pyramidal algorithm (Mallat's algorithm), which has a computational complexity of $\mathcal{O}(L)$. This is even more efficient than the $\mathcal{O}(L \log L)$ complexity of the FFT, making it the most computationally lean choice for real-time debiasing in long-sequence forecasting.
\end{itemize}

\paragraph{Summary.}
The DWT provides a localized orthogonal solution that targets the structural driver of optimization bias with higher granularity than the DFT. 
By reducing temporal dependence across multiple scales, it can lower the effective SSNR while preserving local signal structure.

\subsection{Duality of Gradient Flaws in $\ell_p$ Norms}
\label{app:G3}
\label{app:grad_l_p_analysis}

\paragraph{Gradient Dominance in $\ell_2$ Norm.}
The gradient of $\ell_2$ norm is driven by absolute error. For a single predicted component $\hat{x}_k$, the gradient is:
\begin{equation}
    \frac{\partial}{\partial \hat{x}_k} \Vert x_k - \hat{x}_k \Vert_2^2 
    = -2 (x_k - \hat{x}_k) .
\end{equation}

To understand the scale dependence, we can express the absolute error as the product of the relative error ($\hat{e}_k = (x_k - \hat{x}_k) / \Vert x_k \Vert \in [-1, 1]$) and the true magnitude ($\Vert x_k \Vert$). The gradient's magnitude is therefore proportional to $\Vert x_k \Vert$.
Consequently, high-magnitude components generate overwhelmingly strong gradients, creating \textit{gradient dominance}, while low-magnitude components receive weak optimization signals.

\paragraph{Gradient Fatigue in $\ell_1$ Norm.}
Conversely, the gradient of $\ell_1$ norm depends only on the \textit{sign} of the error, not its magnitude:
\begin{equation}
    \frac{\partial}{\partial \hat{x}_k} \Vert x_k - \hat{x}_k \Vert_1 
    = - \sgn( x_k - \hat{x}_k ) .
\end{equation}

The gradient's magnitude is always 1 (or 0). This uniform ``push" is too weak for large-magnitude components, which require substantial updates to converge, leading to what we term \textit{gradient fatigue}. Simultaneously, this constant force can be too aggressive for low-magnitude components, causing overshooting and instability.

\subsection{Statistical Analysis of Harmonized $\ell_p$ Norm}
\label{app:G4}
\label{app:stat_harm}

\subsubsection{Component-wise Optimum Preservation}
\label{app:harm_point_est}
First and foremost, the re-weighting mechanism should alter only the optimization dynamics without shifting the component-wise optimum of the transformed-domain target.

Consider the Harmonized MSE ($\mcL_{\text{Harm}, \ell_2}$) applied to a specific component $k$. The loss is defined as the weighted squared Euclidean distance:
\begin{equation}
    \mcL_{\text{Harm}, \ell_2}^k 
    = w_k \Vert f_k - \hat{f}_k \Vert_2^2, \quad 
    \text{where} \quad w_k = 1 + \frac{\gamma}{\bar{f}_k + \epsilon}
\end{equation}
where $f_k$ represents the ground truth and $\hat{f}_k$ denotes the estimation. 

Since the weight $w_k$ is strictly positive ($w_k > 0$) and independent of the instantaneous estimation error $(f_k - \hat{f}_k)$, calculating the gradient with respect to the estimator $\hat{f}_k$ yields:
\begin{equation}
    \frac{\partial \mcL_{\text{Harm}, \ell_2}^k}{\partial \hat{f}_k}
    = -2 w_k (f_k - \hat{f}_k)
\end{equation}

Setting the gradient to zero to find the optimal estimator $\hat{f}_k^*$:
\begin{equation}
    -2 w_k (f_k - \hat{f}_k) = 0 
    \quad \Longrightarrow \quad 
    \hat{f}_k^* = f_k
\end{equation}

If the ground truth $f_k$ is a realization of the true underlying signal $f_k^{\text{true}}$ plus zero-mean noise $\xi$, \ie $f_k = f_k^{\text{true}} + \xi$, taking the expectation yields:
\begin{equation}
    \expt [\hat{f}_k^*] 
    = \expt [f_k] 
    = f_k^{\text{true}}
\end{equation}

For Harmonized MAE ($\mcL_{\text{Harm}, \ell_1}$), the positive weight $w_k$ also preserves the component-wise minimizer. 
For a single realization, the subgradient condition is satisfied at $\hat{f}_k=f_k$ because $0 \in \partial |f_k-\hat{f}_k|$:
\begin{gather}
    \mcL_{\text{Harm}, \ell_1}^k 
    = w_k \Vert f_k - \hat{f}_k \Vert_1, \quad \text{where} \quad w_k = 1 + \gamma \bar{f}_k
    \\
    \partial_{\hat{f}_k} \mcL_{\text{Harm}, \ell_1}^k
    = w_k \, \partial \vert \hat{f}_k - f_k \vert
    \\
    0 \in \partial_{\hat{f}_k} \mcL_{\text{Harm}, \ell_1}^k
    \quad \Longrightarrow \quad
    \hat{f}_k^* = f_k .
\end{gather}
For a random target, the population minimizer of $\ell_1$ is the conditional median. 
When $\xi$ is symmetric about zero, this median equals $f_k^{\text{true}}$, so the point-estimation consistency argument carries through.

\textbf{Conclusion:} 
The Harmonized weighting term $w_k$ scales the gradient magnitude, thereby resolving gradient dominance / fatigue, but it does not shift the component-wise optimum.

\subsubsection{Element Bias: Balancing Training Stability and Dynamics Rectification}
\label{app:harm_element_bias}

We generalize Spectral Bias to \textbf{Element Bias} to describe the phenomenon where optimization dynamics are dominated by high-magnitude components in any structural orthogonalization (\eg DFT, DWT, or SVD). 
While a strictly scale-normalized loss (\ie spectral whitening, $\mcL_{\text{Scale}, \ell_2} = \sum_k^K \Vert f_k - \hat{f}_k \Vert_2^2 / \bar{f}_k^2$ and $\mcL_{\text{Scale}, \ell_1} = \sum_k^K \Vert f_k - \hat{f}_k \Vert_1 / \bar{f}_k$) theoretically improves uniform gradient sensitivity, it is practically unstable under uneven Signal-to-Noise Ratio (SNR) across components. 
Strict whitening assigns aggressive weights to noise-dominated low-amplitude components, leading to \textbf{training instability} and noise overfitting, while reducing the relative gradient contribution of high-amplitude main structures.

Our Harmonized Loss resolves this dichotomy by establishing a strategic equilibrium between \textbf{Optimization Dynamics Rectification} and \textbf{Training Stability} through an adaptive soft-switching mechanism:
\begin{itemize}
    \item \textbf{Stability Anchor (Dominant Regime):} 
    For large-amplitude components, the harmonizing weight stays close to the standard norm scale, retaining sensitivity to \textbf{amplitude}. 
    This regime anchors the optimization trajectory on high-energy structures and avoids destabilization by noise-dominated components.

    \item \textbf{Dynamics Rectification (Weak Regime):} 
    For small-amplitude components, the harmonizing weight increases their relative gradient strength without the singular behavior of strict whitening. 
    This rectifies \textit{gradient fatigue}, ensuring that subtle but informative details receive sufficient gradient flow.
\end{itemize}

\textbf{Conclusion:} 
Element Bias is not blindly eliminated but \textbf{adaptively managed}. The Harmonized Loss achieves a Pareto-efficient balance, allowing the model to capture fine-grained details without compromising its grasp on global patterns or succumbing to numerical instability.

\subsubsection{Information-Theoretic Consistency via EOB Theory}
\label{app:harm_EOB_bias}

Finally, we connect the proposed Harmonized $\ell_p$ Norm back to our foundational EOB Theory to demonstrate its consistency with the transformed-domain objective.

Recall that EOB is formalized as the KL divergence between the true sequential law and the factorized surrogate induced by point-wise supervision:
\begin{equation}
    \expt[\bias]
    =
    D_{\textit{KL}}\!\left(p(x_{1:T}) \Vert q(x_{1:T})\right),
    \quad
    q(x_{1:T}) = \prod_{t=1}^{T} p(x_t).
\end{equation}
Structural orthogonalization replaces the supervised target by $x_o=\mathcal{T}_{\text{ortho}}(x)$, whose transformed law is closer to a factorized surrogate,
\begin{equation}
    q_o(x_o) = \prod_{k=1}^{T} p(x_{o,k}),
    \quad
    \cov\!\left(\mathcal{T}_{\text{ortho}}(X)\right)
    \approx \diag(\sigma_1^2,\dots,\sigma_T^2).
\end{equation}

\textbf{1. Consistency of the Component-wise Optimum} \\
Although the Harmonized $\ell_p$ Norm introduces spectral weights $w_k$, these weights are derived from target-side spectral statistics and are independent of the model parameters during each update. 
Because the weights are strictly positive, they do not change the component-wise zero-error minimizer after structural orthogonalization. 
The Harmonized $\ell_p$ Norm therefore reshapes the \textit{optimization manifold} in the transformed domain without changing the factorized target $q_o$.

\textbf{2. Asymptotic Unbiasedness at Unit SSNR} \\
A critical test for any debiasing framework is its behavior at the theoretical limit where bias should naturally vanish. 
According to our quantification of stochastic EOB (Theorem~\ref{thm:quan_EOB_general}), the lower bound of EOB is monotonically related to the Structural Signal-to-Noise Ratio: $\expt[\bias_z] \propto \log(\ssnr)$.

When $\ssnr = 1$, the stochastic component is effectively white or perfectly whitened, so temporal correlations are absent and the i.i.d. assumption becomes valid. 
For stationary white processes, spectral energy is uniform in expectation, \ie $\expt[\bar{f}_k] \approx C$ across $k$, and theoretically $\expt[\bias_z] \to 0$.

Under this condition, the adaptive weights in our Harmonized $\ell_p$ Norm become uniform:
\begin{equation}
    w_k = 1 + \frac{\gamma}{\bar{f}_k + \epsilon} \approx 1 + \frac{\gamma}{C + \epsilon} = \text{const}.
\end{equation}
Consequently, the harmonized squared loss degenerates to a scaled version of the standard MSE, $\mcL_{\text{Harm}, \ell_2} \propto \mcL_{\text{MSE}}$, with an analogous uniform-weight limit for $\mcL_{\text{Harm}, \ell_1}$.

\textbf{Conclusion:} 
This shows that the Harmonized $\ell_p$ Norm introduces no additional target mismatch in the unit-SSNR limit. 
Its role is to stabilize transformed-domain optimization when structural redundancy creates uneven component magnitudes.

\newpage
\section{Gradient Analysis of Discrete Fourier Transform}
\label{app:H}
\label{app:grad_analysis}

In this section, we analyze the gradients induced by several DFT-based objectives.

\subsection{Point-wise Loss Function Based on $\ell_2$ Norm}
\label{app:H1}
We begin by analyzing point-wise loss functions based on the ubiquitous $\ell_2$ norm.

\subsubsection{Decoupled Optimization of Real and Imaginary Components}
\label{app:ind_opt_real_imag}

Consider a time series sample $\boldsymbol{x} \in \mathbb{R}^{T}$. 
The DFT can be expressed as a linear projection via a unitary matrix $\mathbf{U} \in \mathbb{C}^{T \times T}$:
\begin{equation}
    \boldsymbol{f} = \mathbf{U}\boldsymbol{x}
\end{equation}
where $\boldsymbol{f} \in \mathbb{C}^{T}$ is the complex spectrum. 
The elements of the DFT matrix $\mathbf{U}$ are given by $u_{k, \tau} = \frac{1}{\sqrt{T}} \exp(-j 2\pi k\tau /T)$. 
Notably, $\mathbf{U}$ satisfies the unitary property $\mathbf{U}^{\herm}\mathbf{U} = \mathbf{I}$, where $\mathbf{U}^{\herm}$ denotes the conjugate transpose.
We write $\mathbf{U}_r = \Re(\mathbf{U})$ and $\mathbf{U}_i = \Im(\mathbf{U})$.

We can decompose the spectrum into real and imaginary parts, $\boldsymbol{f} = \boldsymbol{f}_r + j \boldsymbol{f}_i$. 
A common intuitive strategy for frequency-domain optimization is to separately penalize the errors in these components:
\begin{equation}
    \mcL_{\text{freq}, \ell_2} 
    \triangleq \Vert \Re(\boldsymbol{f} - \hat{\boldsymbol{f}}) \Vert_2^2 + \Vert \Im(\boldsymbol{f} - \hat{\boldsymbol{f}}) \Vert_2^2
    = \Vert \mathbf{U}_r (\boldsymbol{x} - \hat{\boldsymbol{x}}) \Vert_2^2 + \Vert \mathbf{U}_i (\boldsymbol{x} - \hat{\boldsymbol{x}}) \Vert_2^2
\end{equation}
where $\hat{\boldsymbol{x}}$ is the model's prediction and $\hat{\boldsymbol{f}} = \mathbf{U} \hat{\boldsymbol{x}}$.

However, a closer examination reveals that this objective is mathematically redundant under the $\ell_2$ norm. 
Leveraging the property that for any complex vector $\boldsymbol{z}$, $\Vert \boldsymbol{z} \Vert_2^2 = \Vert \Re(\boldsymbol{z}) \Vert_2^2 + \Vert \Im(\boldsymbol{z}) \Vert_2^2$, and the unitary nature of $\mathbf{U}$, we can simplify $\mcL_{\text{freq}}$:
\begin{equation}
\begin{aligned}
    \mcL_{\text{freq}, \ell_2}
    &= \Vert \boldsymbol{f} - \hat{\boldsymbol{f}} \Vert_2^2 
    \\
    &= \left(\mathbf{U}(\boldsymbol{x} - \hat{\boldsymbol{x}})\right)^{\herm} \left(\mathbf{U}(\boldsymbol{x} - \hat{\boldsymbol{x}})\right)
    \\
    &= \left(\boldsymbol{x} - \hat{\boldsymbol{x}}\right)^{\top} \mathbf{U}^{\herm}\mathbf{U} \left(\boldsymbol{x} - \hat{\boldsymbol{x}}\right)
    \\
    &= \Vert \boldsymbol{x} - \hat{\boldsymbol{x}} \Vert_2^2 
    \equiv \mcL_{\text{temp}, \ell_2} .  
\end{aligned}
\end{equation}

This derivation proves that minimizing the squared $\ell_2$ error in the frequency domain is exactly equivalent to minimizing it in the time domain. Consequently, their gradients are identical:
\begin{equation}
    \frac{\partial \mcL_{\text{freq}, \ell_2}}{\partial \hat{\boldsymbol{x}}}
    \equiv \frac{\partial \mcL_{\text{temp}, \ell_2}}{\partial \hat{\boldsymbol{x}}} 
    = -2 (\boldsymbol{x}- \hat{\boldsymbol{x}}) .
\end{equation}
This result is a direct manifestation of \textit{Parseval's Theorem}, which guarantees the preservation of energy (and thus Euclidean distance) under unitary transformations.

\begin{tcolorbox}[tcbset]
\begin{theorem}
\label{thm:mse_invariance}
    \textbf{(Invariance of $\ell_2$ Optimization under Unitary Transform)} 
    Let $\mathbf{U}$ be a unitary operator (\eg DFT matrix) such that $\mathbf{U}^{\herm}\mathbf{U} = \mathbf{I}$. 
    The squared $\ell_2$ objective and its gradient with respect to the prediction are invariant under the transformation $\mathbf{U}$.
\end{theorem}
\end{tcolorbox}

\textit{Remark:} \textbf{(Reconciliation with Structural Orthogonalization)}
Theorem~\ref{thm:mse_invariance} demonstrates that simply changing the basis does not alter the optimization signal if the loss remains the squared point-wise $\ell_2$ objective, since the gradient dynamics remain isometric.
The debiasing program in Section~\ref{sec:debiasing_prog} therefore combines the transform with a non-isotropic loss, so that component-wise supervision can break this rotational invariance and alter the effective SSNR experienced by the optimizer.

\subsubsection{Decoupled Optimization of Amplitude and Phase}
An alternative approach is to disentangle the objectives by penalizing errors in the amplitude and phase spectra separately. 
Let $\boldsymbol{A}=|\mathbf{U}\boldsymbol{x}|$ and $\boldsymbol{\theta}=\arg(\mathbf{U}\boldsymbol{x})$ denote the target amplitude and phase, and let $\hat{\boldsymbol{A}} = |\mathbf{U}\hat{\boldsymbol{x}}|$ and $\hat{\boldsymbol{\theta}} = \arg(\mathbf{U}\hat{\boldsymbol{x}})$ denote their predicted counterparts.
The decoupled objectives are defined as:
\begin{equation}
    \mcL_{A, \ell_2} 
    = \Vert \boldsymbol{A} - \hat{\boldsymbol{A}} \Vert_2^2 , \quad 
    \mcL_{\theta, \ell_2} 
    = \Vert \boldsymbol{\theta} - \hat{\boldsymbol{\theta}} \Vert_2^2 .
\end{equation}

The gradients of these loss functions with respect to the prediction $\hat{\boldsymbol{x}}$ are derived via the chain rule. 
We can simplify the expressions by noting that $\mathbf{U}_r \hat{\boldsymbol{x}} / \hat{\boldsymbol{A}} = \cos\hat{\boldsymbol{\theta}}$ and $\mathbf{U}_i \hat{\boldsymbol{x}} / \hat{\boldsymbol{A}} = \sin\hat{\boldsymbol{\theta}}$:
\begin{align}
    \frac{\partial \mcL_{A, \ell_2}}{\partial \hat{\boldsymbol{x}}} 
    &=
    - 2 \mathbf{U}_r^{\top} \left( (\boldsymbol{A} - \hat{\boldsymbol{A}}) \odot \cos\hat{\boldsymbol{\theta}} \right)
    - 2 \mathbf{U}_i^{\top} \left( (\boldsymbol{A} - \hat{\boldsymbol{A}}) \odot \sin\hat{\boldsymbol{\theta}} \right)
    \\
    \frac{\partial \mcL_{\theta, \ell_2}}{\partial \hat{\boldsymbol{x}}} 
    &=
    + 2 \mathbf{U}_r^{\top} \left( (\boldsymbol{\theta} - \hat{\boldsymbol{\theta}}) \odot \frac{\sin\hat{\boldsymbol{\theta}}}{\hat{\boldsymbol{A}}} \right)
    - 2 \mathbf{U}_i^{\top} \left( (\boldsymbol{\theta} - \hat{\boldsymbol{\theta}}) \odot \frac{\cos\hat{\boldsymbol{\theta}}}{\hat{\boldsymbol{A}}} \right)
\end{align}
where $\odot$ denotes the Hadamard product, divisions are element-wise, and phase derivatives are taken on a fixed local branch away from zero-amplitude entries.

\textit{Remark:} The term $\hat{\boldsymbol{A}}$ in the denominator of the phase gradient reveals a critical numerical instability: for frequency components with near-zero amplitude, the phase gradient tends to explode.

\subsubsection{Decoupled Optimization of Error Amplitude and Phase}
We now examine the properties of the error signal itself in the frequency domain. 
Let the error vector be $\boldsymbol{e} = \boldsymbol{x} - \hat{\boldsymbol{x}}$, and its complex spectrum be $\boldsymbol{f}_e = \mathbf{U}\boldsymbol{e}$. 
We define the error amplitude $\bar{\boldsymbol{A}} \in \mathbb{R}^T$ and error phase $\bar{\boldsymbol{\theta}} \in \mathbb{R}^T$ as:
\begin{gather}
    \bar{\boldsymbol{A}} = \vert \boldsymbol{f}_e \vert = \vert \mathbf{U}(\boldsymbol{x} - \hat{\boldsymbol{x}}) \vert, 
    \quad
    \bar{\boldsymbol{\theta}} = \arg(\boldsymbol{f}_e) .
\end{gather}
The corresponding squared $\ell_2$ objectives are:
\begin{equation}
    \mcL_{\bar{A}, \ell_2} = \Vert \bar{\boldsymbol{A}} \Vert_2^2, 
    \quad
    \mcL_{\bar{\theta}, \ell_2} = \Vert \bar{\boldsymbol{\theta}} \Vert_2^2 .
\end{equation}

\textbf{1. The Redundancy of Error Amplitude:}
Analyzing the gradient of the error amplitude loss reveals a fundamental equivalence. By \textit{Parseval's theorem}, the energy of the error spectrum magnitude is identical to the energy of the time-domain error:
\begin{equation}
    \mcL_{\bar{A}, \ell_2} 
    = \Vert \mathbf{U}(\boldsymbol{x} - \hat{\boldsymbol{x}}) \Vert_2^2 
    = \Vert \boldsymbol{x} - \hat{\boldsymbol{x}} \Vert_2^2
    \equiv \mcL_{\text{temp}, \ell_2} .
\end{equation}
Consequently, its gradient collapses back to the standard residual:
\begin{equation}
    \frac{\partial \mcL_{\bar{A}, \ell_2}}{\partial \hat{\boldsymbol{x}}} 
    = -2 (\boldsymbol{x} - \hat{\boldsymbol{x}}) .
\end{equation}

\begin{tcolorbox}[tcbset]
\textbf{Insight:} Optimizing the amplitude of the frequency-domain error under the $\ell_2$ norm yields no new optimization signal, as it is mathematically identical to the temporal squared $\ell_2$ objective.
\end{tcolorbox}

\textbf{2. The Distinctness of Error Phase:}
In contrast, the gradient for the phase of the error, $\mcL_{\bar{\theta}}$, provides a mathematically distinct learning signal:
\begin{equation}
    \frac{\partial \mcL_{\bar{\theta}, \ell_2}}{\partial \hat{\boldsymbol{x}}} =
    + 2 \mathbf{U}_r^{\top} \left( \bar{\boldsymbol{\theta}} \odot \frac{\mathbf{U}_i \boldsymbol{e}}{\bar{\boldsymbol{A}}^2} \right) 
    - 2 \mathbf{U}_i^{\top} \left( \bar{\boldsymbol{\theta}} \odot \frac{\mathbf{U}_r \boldsymbol{e}}{\bar{\boldsymbol{A}}^2} \right) .
\end{equation}
This confirms that while penalizing error amplitude is redundant, penalizing error phase offers a distinct optimization path from the temporal squared $\ell_2$ objective.
The same element-wise and fixed-branch convention applies to this phase gradient.

\textit{Remark (Gradient Instability):} Note the term $\bar{\boldsymbol{A}}^2$ in the denominator. As the model converges and the error amplitude $\bar{\boldsymbol{A}} \to 0$, the gradient for the phase objective approaches infinity. This \textit{singularity} suggests that pure phase optimization is numerically unstable near convergence.

\subsection{Point-wise Loss Function Based on $\ell_1$ Norm}
\label{app:H2}

We now investigate the properties of the $\ell_1$ norm in the frequency domain. Unlike the $\ell_2$ norm, which induces isotropic gradients leading to the time-frequency equivalence proven in Theorem~\ref{thm:mse_invariance}, the $\ell_1$ norm is \textit{anisotropic}. Its subgradient (\ie the sign function) provides update signals that are independent of the error magnitude, offering distinct robustness and sparsity-inducing characteristics.

\subsubsection{Decoupled Optimization of Real and Imaginary Components}
The $\ell_1$ analogue to the standard spectral loss is defined by separately penalizing the real and imaginary components:
\begin{equation}
    \mcL_{\text{freq}, \ell_1} 
    = \Vert \Re(\boldsymbol{f} - \hat{\boldsymbol{f}}) \Vert_1 + \Vert \Im(\boldsymbol{f} - \hat{\boldsymbol{f}}) \Vert_1
    = \Vert \mathbf{U}_r (\boldsymbol{x} - \hat{\boldsymbol{x}}) \Vert_1 + \Vert \mathbf{U}_i (\boldsymbol{x} - \hat{\boldsymbol{x}}) \Vert_1 .
    \label{eq:real_imag_l1}
\end{equation}

With $\boldsymbol{e} = \boldsymbol{x} - \hat{\boldsymbol{x}}$, its subgradient with respect to the temporal prediction $\hat{\boldsymbol{x}}$ is:
\begin{equation}
    \partial_{\hat{\boldsymbol{x}}} \mcL_{\text{freq}, \ell_1}
    =
    - \mathbf{U}_r^{\top} \sgn(\mathbf{U}_r \boldsymbol{e})
    - \mathbf{U}_i^{\top} \sgn(\mathbf{U}_i \boldsymbol{e})
\end{equation}

\begin{tcolorbox}[tcbset]
\textbf{The Breaking of Equivalence:} 
Crucially, unlike the $\ell_2$ case where Parseval's theorem guarantees $\Vert \mathbf{U}\boldsymbol{x} \Vert_2 = \Vert \boldsymbol{x} \Vert_2$, the $\ell_1$ norm does not satisfy rotational invariance (\ie $\Vert \mathbf{U}\boldsymbol{x} \Vert_1 \neq \Vert \boldsymbol{x} \Vert_1$). 
Consequently, the gradient in Eq.~\eqref{eq:real_imag_l1} \textbf{does not collapse} to the temporal $\ell_1$ update. Frequency-domain $\ell_1$ optimization therefore gives a mathematically distinct trajectory that is sensitive to spectral sparsity.
\end{tcolorbox}

\subsubsection{Decoupled Optimization of Amplitude and Phase}

By decomposing the complex spectrum into polar coordinates, we can apply the $\ell_1$ norm to the magnitude envelope and phase alignment independently. 
Let $\boldsymbol{A}=|\mathbf{U}\boldsymbol{x}|$, $\boldsymbol{\theta}=\arg(\mathbf{U}\boldsymbol{x})$, $\hat{\boldsymbol{A}} = |\mathbf{U}\hat{\boldsymbol{x}}|$, and $\hat{\boldsymbol{\theta}} = \arg(\mathbf{U}\hat{\boldsymbol{x}})$. The objectives are:
\begin{equation}
    \mcL_{A, \ell_1} = \Vert \boldsymbol{A} - \hat{\boldsymbol{A}} \Vert_1 , \quad 
    \mcL_{\theta, \ell_1} = \Vert \boldsymbol{\theta} - \hat{\boldsymbol{\theta}} \Vert_1 .
\end{equation}

The corresponding subgradients are derived via the chain rule. We can simplify the expressions by noting that $\mathbf{U}_r \hat{\boldsymbol{x}} / \hat{\boldsymbol{A}} = \cos\hat{\boldsymbol{\theta}}$ and $\mathbf{U}_i \hat{\boldsymbol{x}} / \hat{\boldsymbol{A}} = \sin\hat{\boldsymbol{\theta}}$:
\begin{align}
    \partial_{\hat{\boldsymbol{x}}} \mcL_{A, \ell_1}
    &= 
    - \mathbf{U}_r^{\top} \left( \sgn(\boldsymbol{A} - \hat{\boldsymbol{A}}) \odot \cos\hat{\boldsymbol{\theta}} \right) 
    - \mathbf{U}_i^{\top} \left( \sgn(\boldsymbol{A} - \hat{\boldsymbol{A}}) \odot \sin\hat{\boldsymbol{\theta}} \right)
    \\
    \partial_{\hat{\boldsymbol{x}}} \mcL_{\theta, \ell_1}
    &= 
    + \mathbf{U}_r^{\top} \left( \sgn(\boldsymbol{\theta} - \hat{\boldsymbol{\theta}}) \odot \frac{\sin\hat{\boldsymbol{\theta}}}{\hat{\boldsymbol{A}}} \right) 
    - \mathbf{U}_i^{\top} \left( \sgn(\boldsymbol{\theta} - \hat{\boldsymbol{\theta}}) \odot \frac{\cos\hat{\boldsymbol{\theta}}}{\hat{\boldsymbol{A}}} \right) .
\end{align}
Again, divisions are element-wise and phase subgradients are taken on a fixed local branch away from zero-amplitude entries.

\textbf{Geometric Interpretation:}
\begin{itemize}
    \item \textbf{Amplitude Gradient:} The update vector is directed along the current phase angle $\hat{\boldsymbol{\theta}}$ with a constant magnitude determined by $\sgn(\boldsymbol{A}-\hat{\boldsymbol{A}})$.
    \item \textbf{Phase Gradient:} The update is orthogonal to the amplitude direction and scales with $1/\hat{\boldsymbol{A}}$.
\end{itemize}

\textit{Remark:} The presence of $1/\hat{\boldsymbol{A}}$ in the phase gradient again highlights the \textit{Phase Singularity} problem: for frequency components where the model predicts near-zero energy ($\hat{\boldsymbol{A}} \approx 0$), the phase gradient becomes numerically unstable.

\subsubsection{Decoupled Optimization of Error Amplitude and Phase}
Finally, we consider the $\ell_1$ norm of the error's amplitude and phase. Let $\bar{\boldsymbol{A}} = |\mathbf{U}\boldsymbol{e}|$ and $\bar{\boldsymbol{\theta}}=\arg(\mathbf{U}\boldsymbol{e})$ be the amplitude and phase of the error spectrum.
\begin{equation}
    \mcL_{\bar{A}, \ell_1} = \Vert \bar{\boldsymbol{A}} \Vert_1 , \quad 
    \mcL_{\bar{\theta}, \ell_1} = \Vert \bar{\boldsymbol{\theta}} \Vert_1 .
\end{equation}

\textbf{Gradient of Error Amplitude:}
Since the error amplitude $\bar{\boldsymbol{A}}$ is intrinsically non-negative, minimizing its $\ell_1$ norm is equivalent to minimizing the sum of magnitudes. The subgradient derivation simplifies compared to the $\ell_2$ case:
\begin{equation}
\begin{aligned}
    \partial_{\hat{\boldsymbol{x}}} \mcL_{\bar{A}, \ell_1}
    &=
    - \mathbf{U}_r^{\top} \left( \frac{\mathbf{U}_r \boldsymbol{e}}{\bar{\boldsymbol{A}}} \right)
    - \mathbf{U}_i^{\top} \left( \frac{\mathbf{U}_i \boldsymbol{e}}{\bar{\boldsymbol{A}}} \right)
    \\
    &= 
    - \mathbf{U}_r^{\top} \cos \bar{\boldsymbol{\theta}}
    - \mathbf{U}_i^{\top} \sin \bar{\boldsymbol{\theta}}
    \\
    &= -\Re\!\left(\mathbf{U}^{\herm} \frac{\mathbf{U}\boldsymbol{e}}{|\mathbf{U}\boldsymbol{e}|}\right)
    = -\Re\!\left(\mathbf{U}^{\herm} e^{j \bar{\boldsymbol{\theta}}}\right).
\end{aligned}
\end{equation}
where division is element-wise and $\bar{\boldsymbol{\theta}}$ is the phase of the error spectrum.

\textit{Interpretation:} 
The gradient depends on the phase of the error spectrum and normalizes out its magnitude. 
This applies more uniform gradient pressure across frequencies and prevents high-energy components from fully dominating the update, a key distinction from $\ell_2$-based losses.

\textbf{Gradient of Error Phase:}
Similarly, the gradient for the error phase focuses purely on rotational alignment:
\begin{equation}
    \partial_{\hat{\boldsymbol{x}}} \mcL_{\bar{\theta}, \ell_1} = 
    + \mathbf{U}_r^{\top} \left( \sgn(\bar{\boldsymbol{\theta}}) \odot \frac{\mathbf{U}_i \boldsymbol{e}}{\bar{\boldsymbol{A}}^2} \right)
    - \mathbf{U}_i^{\top} \left( \sgn(\bar{\boldsymbol{\theta}}) \odot \frac{\mathbf{U}_r \boldsymbol{e}}{\bar{\boldsymbol{A}}^2} \right) .
\end{equation}
The same element-wise and fixed-branch convention applies here.
Both gradients provide non-trivial and distinct learning signals compared to a simple temporal $\ell_1$ loss, showing that the frequency-domain $\ell_1$ norm changes the optimization geometry.

\textit{Notably, the structure of these $\ell_1$ gradients is analogous to their $\ell_2$ counterparts, but the linear residual terms are replaced by sign-based terms, altering the optimization dynamics significantly.}

\newpage
\section{Experimental Settings}
\label{app:I}

All experiments are conducted on a workstation equipped with four NVIDIA RTX 4090 GPUs, running Ubuntu 22.04 with PyTorch v2.8 and Python 3.9.

\subsection{Synthetic Dataset Generation Mathematical Mechanism}
\label{app:I1}
\label{app:synthetic_dataset}
To empirically validate our theoretical findings, particularly the Paradigm Paradox (Theorem~\ref{thm:paradigm_paradox}), we require a process with controllable deterministic and stochastic structural components. We therefore construct a synthetic hybrid process $x_t$ by combining a strong, multi-frequency deterministic signal $v_t$ (a sum of sinusoids) with a stochastic autoregressive process $z_t$ (an $AR(p)$ model):
\begin{equation} 
    x_t 
    = v_t + z_t 
    =   \sum_{i=1}^{K} A_i \sin \left( \frac{2 \pi k_i}{T} t + \varphi_i \right) 
        + \sum_{j=1}^p \phi_j z_{t-j} + \epsilon_t 
\end{equation}
where $v_t$ represents the deterministic component with $K$ harmonics, and $z_t$ is the $AR(p)$ stochastic component driven by innovation noise $\epsilon_t$. $A_i$, $k_i$, and $\varphi_i$ are the amplitude, frequency, and phase of the $i$-th harmonic, respectively.

Assuming the independence between deterministic $\{ v_t \}$ and stochastic $\{ z_t \}$ processes, the total variance of the hybrid process $\sigma_x^2 = \var(x)$ is the sum of the individual variances:
\begin{equation}
    \sigma_x^2
    = \var(v) + \var(z)
    = \sigma_v^2 + \sigma_z^2
\end{equation}
where $\sigma_v^2$ and $\sigma_z^2$ are the variance of deterministic and stochastic processes, respectively.

We now extend our core metric (Definition~\ref{def:ssnr}) to this hybrid process. We define the total SSNR of the process, $\ssnr_x$, as the ratio of the total process variance $\sigma_x^2$ to the variance of the irreducible innovation noise $\sigma_\epsilon^2$ (\ie one-step-ahead optimal prediction variance). Thus, $\ssnr_x$ is quantified as: 
\begin{equation}
    \ssnr_x 
    = \frac{\sigma_x^2}{\sigma_{\epsilon}^2}
    = \frac{\sigma_v^2 + \sigma_z^2}{\sigma_{\epsilon}^2}
    = \ssnr_v + \ssnr_z
\end{equation}
where $\ssnr_v$ and $\ssnr_z$ are the SSNRs of deterministic and stochastic components in the hybrid process, respectively.

\subsubsection{Statistical Property of Deterministic Component}
\label{app:stat_prop_det_comp}
We assume the phases $\varphi_i$ are independent random variables drawn from a uniform distribution $U[0, 2\pi]$. This ensures the deterministic signal $v_t$ has an expected value of zero:
\begin{equation}
    \expt [v_t]
    = \expt \left[ 
        \sum_{i=1}^{K} A_i \sin \left( \frac{2 \pi k_i}{T} t + \varphi_i \right) 
    \right]
    = \sum_{i=1}^{K} A_i \expt \left[ \sin \left( \frac{2 \pi k_i}{T} t + \varphi_i \right) \right]
    = 0 .
\end{equation}

The variance of a single sinusoidal component, when averaged over the random phase $\varphi_i$, is $\frac{1}{2}$:
\begin{equation}
    \var \left( \sin \left( \frac{2 \pi k_i}{T} t + \varphi_i \right) \right)
    = \expt \left[ \sin^2 \left( \frac{2 \pi k_i}{T} t + \varphi_i \right) \right]
    = \frac{k_i}{T} \int_{0}^{\frac{T}{k_i}} \frac{1 - \cos \left( \frac{2 \pi k_i}{T} t + \varphi_i \right)}{2} \diff t
    = \frac{1}{2} .
\end{equation}

With distinct Fourier frequencies over the generation window, the $K$ sinusoidal components are orthogonal; otherwise, their cross terms vanish in expectation over random phases. Thus, the variance of the deterministic process $\sigma_v^2$ is the sum of the variances of sinusoidal components.
\begin{equation}
    \sigma_v^2
    = \var \left( \sum_{i=1}^{K} A_i \sin \left( \frac{2 \pi k_i}{T} t + \varphi_i \right) \right)
    = \sum_{i=1}^K A_i^2 \var \left( \sin \left( \frac{2 \pi k_i}{T} t + \varphi_i \right) \right)
    = \frac{1}{2} \sum_{i=1}^K A_i^2 .
\end{equation}

In our experiments, we employ an adaptive amplitude adjustment technique to enhance the low-frequency component.
\begin{equation}
    A_i 
    = \frac{A \sqrt{K}}{k_i \sqrt{\sum_{j=1}^K 1/k_j^2}} 
\end{equation}
where $A$ denotes the fundamental amplitude and $k_i$ is the randomly selected frequency.

Using the equality for the sum of squared amplitudes, $\ssnr_v$ is calculated as follows:
\begin{gather}
    \sum_{i=1}^K A_i^2 
    = \sum_{i=1}^K \frac{K A^2 / k_i^2 }{\sum_{j=1}^K 1/k_j^2} = K A^2 ,
    \\
    \ssnr_v = \frac{K A^2}{2 \sigma_{\epsilon}^2} .
\end{gather}

\subsubsection{Statistical Property of Stochastic Component}
We set an autoregressive process with $p=1$ as the stochastic component. For a covariance-stationary AR(1) process with zero-mean finite-variance innovations and $|\phi_1|<1$, the stochastic variance $\sigma_z^2$ is controlled by autoregressive coefficient ($\phi_1$) and variance of innovation noise ($\sigma_{\epsilon}^2$):
\begin{equation}
    \sigma_z^2 = \frac{\sigma_{\epsilon}^2}{1 - \phi_1^2} .
\end{equation}

We control the autoregressive coefficient ($\phi_1$) to adjust $\ssnr_z$ with arbitrarily distributed noise,
\begin{equation}
    \ssnr_z = \frac{\sigma_z^2}{\sigma_{\epsilon}^2} = \frac{1}{1 - \phi_1^2}
    \Rightarrow 
    \phi_1 = \sqrt{\frac{\ssnr_z - 1}{\ssnr_z}} .
\end{equation}
 
Consequently, we have the precise formula of $\ssnr_x$:
\begin{equation}
    \ssnr_x 
    = \ssnr_v + \ssnr_z
    = \frac{K A^2}{2 \sigma_{\epsilon}^2} + \frac{1}{1 - \phi_1^2} .
\end{equation}

\begin{table}[t]
    \centering
    \caption{Benchmark Statistical Description.}
    \label{tab:benchmark}
    \resizebox{\linewidth}{!}{%
    \begin{tabular}{cccccc}
        \bottomrule 
        
        \toprule
        Dataset & Channels & Forecast Horizon  & Train / Validation / Test & Frequency & Description     \\ \midrule
        ETTh1   & 7        & 96, 192, 336, 720 & 8545 / 2881 / 2881        & Hourly    & Oil Temperature \\ \midrule
        ETTh2   & 7        & 96, 192, 336, 720 & 8545 / 2881 / 2881        & Hourly    & Oil Temperature \\ \midrule 
        ETTm1   & 7        & 96, 192, 336, 720 & 34465 / 11521 / 11521     & 15 min    & Oil Temperature \\ \midrule 
        ETTm2   & 7        & 96, 192, 336, 720 & 34465 / 11521 / 11521     & 15 min    & Oil Temperature \\ \midrule 
        ECL     & 321      & 96, 192, 336, 720 & 18317 / 2633 / 5261       & Hourly    & Electricity Consumption \\ \midrule 
        Traffic & 862      & 96, 192, 336, 720 & 12185 / 1757 / 3509       & Hourly    & Road Occupancy Rates \\ \midrule 
        Weather & 21       & 96, 192, 336, 720 & 36792 / 5271 / 10540      & 10 min    & $CO_2$ Concentration \\ \midrule
        Exchange& 8        & 96, 192, 336, 720 & 5120 / 665 / 1422         & Daily     & Exchange rate \\ \midrule
        PJM     & 3        & 96, 192, 336, 720 & 36500 / 5147 / 10388      & Hourly    & Electricity Price \\ \midrule
        BE      & 3        & 96, 192, 336, 720 & 36500 / 5147 / 10388      & Hourly    & Electricity Price \\ \midrule
        FR      & 3        & 96, 192, 336, 720 & 36500 / 5147 / 10388      & Hourly    & Electricity Price \\ 
        \bottomrule
    
        \toprule
    \end{tabular}%
    }
\end{table}

\subsection{Real-world Benchmark Description}
\label{app:I2}
\label{app:actual_dataset}
Our empirical validation relies on a diverse suite of real-world benchmarks, detailed in Table~\ref{tab:benchmark}. These datasets were selected to represent a wide range of domains, temporal resolutions, and challenging structural characteristics:
\begin{itemize}[leftmargin=*]
    \item \textbf{ETT} (Electricity Transformer Temperature)~\citep{informer2021}: The ETT dataset includes two hourly-level datasets (ETTh1 and ETTh2) and two 15-minute-level datasets (ETTm1 and ETTm2). Each one includes 7 oil and load features of electricity transformers from July 2016 to July 2018.

    \item \textbf{ECL} (Electricity Consumption Load)~\citep{autoformer2021}: The ECL dataset records the hourly electricity consumption of 321 distinct clients from 2012 to 2014.

    \item \textbf{Traffic}~\citep{autoformer2021}: The Traffic dataset contains hourly road occupancy rates obtained from sensors located on San Francisco freeways from 2015 to 2016.

    \item \textbf{Weather}~\citep{autoformer2021}: The Weather dataset contains 21 indicators of weather (\eg air temperature and humidity), which are collected in Germany. The data is recorded at a high frequency of every 10 minutes.

    \item \textbf{Exchange}~\citep{10.1145/3209978.3210006}: The Exchange dataset records the daily exchange rates of eight different countries relative to the US dollar, ranging from 1990 to 2016.

    \item \textbf{PJM}~\citep{LAGO2021116983}: The PJM dataset represents the Pennsylvania-New Jersey-Maryland market, containing the zonal electricity price within the Commonwealth Edison (COMED) zone, and the corresponding System load and COMED load forecast from 2013-01-01 to 2018-12-24.

    \item \textbf{BE}~\citep{LAGO2021116983}: The BE dataset represents the Belgium electricity market, recording the hourly electricity price, the load forecast in Belgium and the generation forecast in France from 2011-01-09 to 2016-12-31.

    \item \textbf{FR}~\citep{LAGO2021116983}: The FR dataset represents the electricity market in France, recording the hourly prices, and corresponding load and generation forecast from 2012-01-09 to 2017-12-31.
\end{itemize}

\subsection{Backbone Description}
\label{app:I3}
\label{app:backbone}

Transformer-based methods:
\begin{itemize}[leftmargin=*]
    \item \textbf{iTransformer}~\citep{liu2024itransformer} introduces an “inverted” transformer architecture that swaps the roles of queries, keys, and values to simplify and accelerate time series modeling.
    
    \item \textbf{PatchTST}~\citep{nie2023a} treats a time series as a sequence of fixed-length patches and applies Transformer-based patch-wise modeling to capture long-range dependencies.

    \item \textbf{Pyraformer}~\citep{liu2022pyraformer} proposes the pyramidal attention module (PAM) to capture multi-scale temporal dependencies with $O(N)$ linear complexity, effectively handling long sequences.
    
    \item \textbf{FEDformer}~\citep{zhou2022fedformer} leverages frequency-enhanced decomposition within a transformer framework to efficiently model and forecast long-term periodic patterns.
    
    \item \textbf{Autoformer}~\citep{autoformer2021} replaces the traditional self-attention mechanism with the Auto-Correlation mechanism to discover period-based dependencies and incorporates a basic inner block of deep models.
    
    \item \textbf{Transformer}~\citep{NIPS2017_3f5ee243} utilizes a self-attention mechanism to capture global dependencies across the entire sequence, serving as a fundamental baseline for sequence modeling.
\end{itemize}

MLP-based methods:
\begin{itemize}[leftmargin=*]
    \item \textbf{TimeMixer}~\citep{wang2024timemixer} introduces Past-Decomposable-Mixing (PDM) and Future-Multipredictor-Mixing (FMM) blocks, effectively handling multiscale series.
    
    \item \textbf{TSMixer}~\citep{TSMxier2023} operates in both the time and feature dimensions to extract information, efficiently utilizing cross-variate and auxiliary information.

    \item \textbf{DLinear}~\citep{Dlinear2023} decomposes the series into trend and seasonal components, fits each with a simple linear model, and then recombines them for forecasting.
    
    \item \textbf{FreTS}~\citep{FreTS2023} utilizes MLPs in the frequency domain, capturing global dependencies with lower computational complexity.
    
\end{itemize}

CNN-based methods:
\begin{itemize}[leftmargin=*]
    \item \textbf{TimesNet}~\citep{wu2023timesnet} constructs a 2D temporal-variation representation and applies joint time–frequency convolutions to capture general patterns in time series data.
    
    \item \textbf{MICN}~\citep{wang2023micn} designs a multi-scale isometric convolution network that combines down-sampled convolution with isometric convolution to capture local features and global correlations simultaneously.
\end{itemize}

Others:
\begin{itemize}[leftmargin=*]
    \item \textbf{FreDF}~\citep{wang2025fredf} uses Fourier Transform to reduce label autocorrelation by supervising forecasts in the frequency domain.
\end{itemize}

\subsection{Implementation Details}
\label{app:I4}
\label{app:implementation}

\subsubsection{For Verification Experiment of EOB Theory}
\label{app:impl_EOB_theory}
Following the synthetic generation mechanism above, we fix the stochastic component $z_t$ (constant $\sigma_z^2$) and increase the amplitude $A$ of the deterministic component $v_t$. While this effectively increases SSNR, it also increases the total variance $\sigma_x^2$. This protocol highlights the model's inability to converge to the theoretical optimum $\mse^{\text{opt.}}$ as the deterministic signal grows.

We set $A^2 = [0, 1, 2, 3, 4, 5, 6, 7, 8, 9]$ to ensure comprehensive coverage. Correspondingly, the total SSNR, $\ssnr_x$, ranges from 32 to 320 with an interval of 32. Specifically, when $\ssnr_x = 32$, series reduces to purely stochastic $AR(1)$ process. 

For each $\ssnr_x$ value, we generate 100 time series of length 5000, with a training/test split ratio set to 0.7/0.3. We randomly select one time series for experimental validation and sample sub-series with a stride of 1 to construct the dataset. 

The history windows size is fixed at $H = 128$, and the forecast horizon is set to multiples of the history windows size, \ie $h = 128 \times n$ for $n=1, 2, \dots, 10$.

\paragraph{Linear-optimal baseline.}
Since the deterministic component is predictable under the data-generating process, the irreducible forecasting error comes from the stochastic $AR(p)$ component.
Let $\psi_0=1$ and $\{\psi_\ell\}_{\ell\geq0}$ be the impulse-response coefficients of the $AR(p)$ process.
For a direct $h$-step prediction window, the linear-optimal actual MSE is
\begin{equation}
    \mse^{\text{opt.}}_{\text{act.}}(h)
    = \frac{\sigma_\epsilon^2}{h}
    \sum_{r=1}^{h} \sum_{\ell=0}^{r-1} \psi_\ell^2
    = \frac{\sigma_\epsilon^2}{h}
    \sum_{\ell=0}^{h-1} (h-\ell)\psi_\ell^2 .
\end{equation}
In our synthetic experiments, $z_t=\phi_1 z_{t-1}+\epsilon_t$, so $\psi_\ell=\phi_1^\ell$ and
\begin{equation}
    \mse^{\text{opt.}}_{\text{act.}}(h)
    = \frac{\sigma_\epsilon^2}{h}
    \sum_{r=1}^{h} \frac{1-\phi_1^{2r}}{1-\phi_1^2}
    = \frac{\sigma_\epsilon^2}{1-\phi_1^2}
    \left(
        1 - \frac{\phi_1^2(1-\phi_1^{2h})}{h(1-\phi_1^2)}
    \right).
\end{equation}
Thus $\mse^{\text{opt.}}_{\text{act.}}(1)=\sigma_\epsilon^2$ and $\mse^{\text{opt.}}_{\text{act.}}(h) \to \sigma_z^2 = \sigma_{\epsilon}^2 / (1-\phi_1^2)$ as $h$ increases.
The optimal relative MSE used in Figure~\ref{fig:simulation}(c) is $\mse^{\text{opt.}}_{\text{rel.}}(h)=\mse^{\text{opt.}}_{\text{act.}}(h)/\sigma_x^2$.

To ensure statistical reliability, each result is the average of 25 independent experiments with different random seeds.

We adopt the \textit{Adam Optimizer} for model parameter optimization and the \textit{early stop technique} to mitigate overfitting. 

\textit{\textbf{To effectively simulate actual data distributions and comprehensively validate our EOB Theory, we conducted grid experiments across 5 kinds of foundational deep learning architectures and 6 categories of innovation noise distributions.}}

Foundational deep learning architectures: CNN, LSTM, MLP, ModernTCN, and Transformer.

All innovation samples are centered to zero mean and rescaled to variance $\sigma_{\epsilon}^{2}$ before being injected into the AR process.

\begin{table}[]
    \centering
    \caption{Innovation Noise Distribution Parameter Settings. Variance of innovation noise is $\sigma_{\epsilon}^2=0.25$.}
    \label{tab:noise_setting}
    \resizebox{\linewidth}{!}{%
    \begin{tabular}{c|ccccc|c}
        \bottomrule
    
        \toprule
        Innovation Noise  & Continuity & Symmetry      & Support Set           & Tail   & Peak Degree & Parameters \\ \midrule
        Binomial          & \ding{55}  & Approx. Sym.  & $\{0, 1, \dots, n \}$ & None   & $<3$        & $n = 1$, $p = 0.5$                    \\
        Geometric         & \ding{55}  & Right-leaning & $\{1, 2, \dots \}$    & Heavy  & $-$         & $p = 2 (\sqrt{2} - 1)$                \\
        Gaussian          & \ding{51}  & \ding{51}     & $(-\infty, \infty)$   & Middle & $3$         & $\mu = 0$, $\sigma = 0.5$             \\
        Poisson           & \ding{55}  & Approx. Sym.  & $\{0, 1, \dots \}$    & Middle & $\approx 3$ & $\lambda = 0.25$                       \\
        Student’s $t$     & \ding{51}  & \ding{51}     & $(-\infty, \infty)$   & Heavy  & $>3$        & $\nu = 5$, $\alpha = \sqrt{15} / 10$  \\
        Cont. Uniform     & \ding{51}  & \ding{51}     & $[a, b]$              & None   & $1.8$       & $b = -a = \sqrt{3}/2$                 \\
        \bottomrule
    
        \toprule
    \end{tabular}%
    }
\end{table}

Innovation noise is modeled by a distribution with variance $\sigma_{\epsilon}^2$, which can be categorized into discrete vs. continuous, symmetrical vs. asymmetrical, and heavy-tailed vs. light-tailed types. The characteristics and parameter settings of each distribution are summarized in Table~\ref{tab:noise_setting}.
\begin{itemize}[leftmargin=*]
    \item \textbf{Binomial Distribution} is the discrete probability distribution of the number of successes in a sequence of $n$ independent experiments, each asking a yes–no question.
    
    Probability Mass Function:
    \[
        P(X = k) = \binom{n}{k} p^k (1-p)^{n-k},
        \quad k=0,1,\dots,n .
    \]
    Statistical Properties:
    \[
        \expt[X] = np; \quad \var(X) = np(1-p)
    \]
    Configure Parameters:
    \[
        n = 1; \quad p = \frac{1 \pm \sqrt{1 - 4 \sigma_{\epsilon}^2}}{2}
    \]

    \item \textbf{Geometric Distribution} gives the probability that the first occurrence of success requires $k$ independent trials, each with success probability $p$.

    Probability Mass Function:
    \[
        P(X = k) = (1-p)^{k-1}p, \quad k = 1, 2, \dots
    \]
    Statistical Properties:
    \[
        \expt[X] = \frac{1}{p}; \quad \var(X) = \frac{1-p}{p^2}
    \]
    Configure Parameters:
    \[
        p = \frac{-1 + \sqrt{1 + 4\sigma_{\epsilon}^2}}{2\sigma_{\epsilon}^2}
    \]

    \item \textbf{Gaussian Distribution} is important in statistics and is often used in the natural and social sciences to represent real-valued random variables whose distributions are not known.
    
    Probability Density Function:
    \[
        p_X(x) = \frac{1}{\sqrt{2\pi \sigma^2}} \exp \left(
            - \frac{(x - \mu)^2}{2 \sigma^2} 
        \right)
    \]
    Configure Parameters:
    \[
        \mu = 0; \quad \sigma = \sigma_{\epsilon}
    \]

    \item \textbf{Poisson Distribution} is a discrete probability distribution that expresses the probability of a given number of events occurring in a fixed interval of time if these events occur with a known constant mean rate and independently of the time since the last event.
    
    Probability Mass Function:
    \[
        P(X = k) = 
        \begin{cases} 
            \displaystyle \frac{\lambda^k \mathrm{e}^{-\lambda}}{k!}, & k = 0, 1, 2, \dots, \\
            0, & \text{otherwise},
        \end{cases}
    \]
    Statistical Properties:
    \[
        \expt[X] = \lambda; \quad \var(X) = \lambda
    \]
    Configure Parameters:
    \[
        \lambda = \sigma_{\epsilon}^{2}
    \]

    \item \textbf{Student's $t$ Distribution} is a continuous probability distribution that generalizes the standard normal distribution, but has heavier tails, and the amount of probability mass in the tails is controlled by the parameter $\nu$.
    
    Probability Density Function:
    \[
        p_X(x; \nu) 
        = \frac{\Gamma(\frac{\nu+1}{2})}{\sqrt{\pi \nu} \Gamma(\frac{\nu}{2})} \left( 
            1 + \frac{x^2}{\nu}    
        \right)^{-(\nu+1)/2}
    \]
    where $\Gamma(\cdot)$ is the gamma function.
    
    Statistical Properties:
    \[
        \expt[X] = 
        \begin{dcases}
            0, & \nu > 1    \\
            \text{Does not exist}, & \nu < 1
        \end{dcases}, \quad
        \var(X) = 
        \begin{dcases}
            \frac{\nu}{\nu - 2}, & \nu > 2  \\
            \text{Does not exist}, & \nu < 2
        \end{dcases}
    \]
    Configure Parameters:
    \[
        \nu = 5, \quad 
        \alpha = \sigma_{\epsilon} \sqrt{\frac{\nu - 2}{\nu}}, \quad
        X' = \alpha X
    \]

    \item \textbf{Continuous Uniform Distribution} describes an experiment where there is an arbitrary outcome that lies between certain bounds. The bounds are defined by the parameters, $a$ and $b$, which are the minimum and maximum values.
    
    Probability Density Function:
    \[
        p_X(x) = 
        \begin{dcases}
            \frac{1}{b - a}, & \text{if} \; a \le x \le b  \\
            0, &\text{otherwise}
        \end{dcases}
    \]
    Statistical Properties:
    \[
        \expt[X] = \frac{a+b}{2}, \quad \var(X) = \frac{(b - a)^2}{12}
    \]
    Configure Parameters:
    \[
        b = -a = \sqrt{3} \sigma_{\epsilon}
    \]
\end{itemize}

\subsubsection{For Mechanistic Insight on Trigonometric Series}
\label{app:impl_insight_exp}

The generated trigonometric series follows the dynamic mechanism described in Appendix~\ref{app:stat_prop_det_comp}. 
We randomly select $k \in \{ 1, 2, 3, 4, 5 \}$ components with a maximum cutoff frequency of $f_{\max} = 15$ and a random phase $\theta$. 
The original series is split into training and testing sets with a ratio 7:3. 
Both the historical context and the forecasting horizon are set to 200 points. 
A Transformer architecture is employed as the model backbone.

\subsubsection{For Long-term Forecasting Tasks}
\label{app:impl_forecasting}

The baseline models are reproduced using the iTransformer benchmark scripts~\citep{liu2024itransformer}, which employ Adam optimizer to minimize the target loss. Datasets are split chronologically into training, validation, and test sets with the ratio of 6:2:2. Following the protocol outlined in the comprehensive benchmark~\citep{qiu2024tfb}, the dropping-last trick is disabled during the test phase.

Specifically, we adopt proposed $\mcL_{\text{Harm}, \ell_p}$ with the $\ell_1$ norm, setting the hyperparameters to $\gamma=0.5$ and $\beta=0.3$ and excluding the temporal MSE term. For structural orthogonality, dataset-specific strategies are implemented as variants of DFT and DWT.

Results in Tables~\ref{tab:longterm_app} and~\ref{tab:longterm_app_part2} are averaged over 20 random seeds.
Detailed analysis is available at Appendix~\ref{app:long_term_forecast}.

\subsubsection{For Missing Data Imputation Tasks}
\label{app:impl_imputation}

All models are trained using an autoencoding approach: they are optimized to reconstruct observed entries from incomplete input sequences during training, and are subsequently leveraged to impute actual missing values during inference.

Specifically, we adopt proposed $\mcL_{\text{Harm}, \ell_p}$ with the $\ell_1$ norm, setting the hyperparameters to $\gamma=0.3$ and $\beta=0.3$. For structural orthogonality, dataset-specific strategies are implemented as variants of DFT and DWT.

Results in Table~\ref{tab:imp_app} are averaged over 20 random seeds.
Detailed analysis is available at Appendix~\ref{app:imputation}.

\newpage
\section{Extensive Experimental Analysis}
\label{app:J}

\subsection{Empirical Verification of EOB Theory}
\label{app:J1}
\label{app:EOB_simulation}

\begin{figure}
    \centering
    \includegraphics[width=0.98\linewidth]{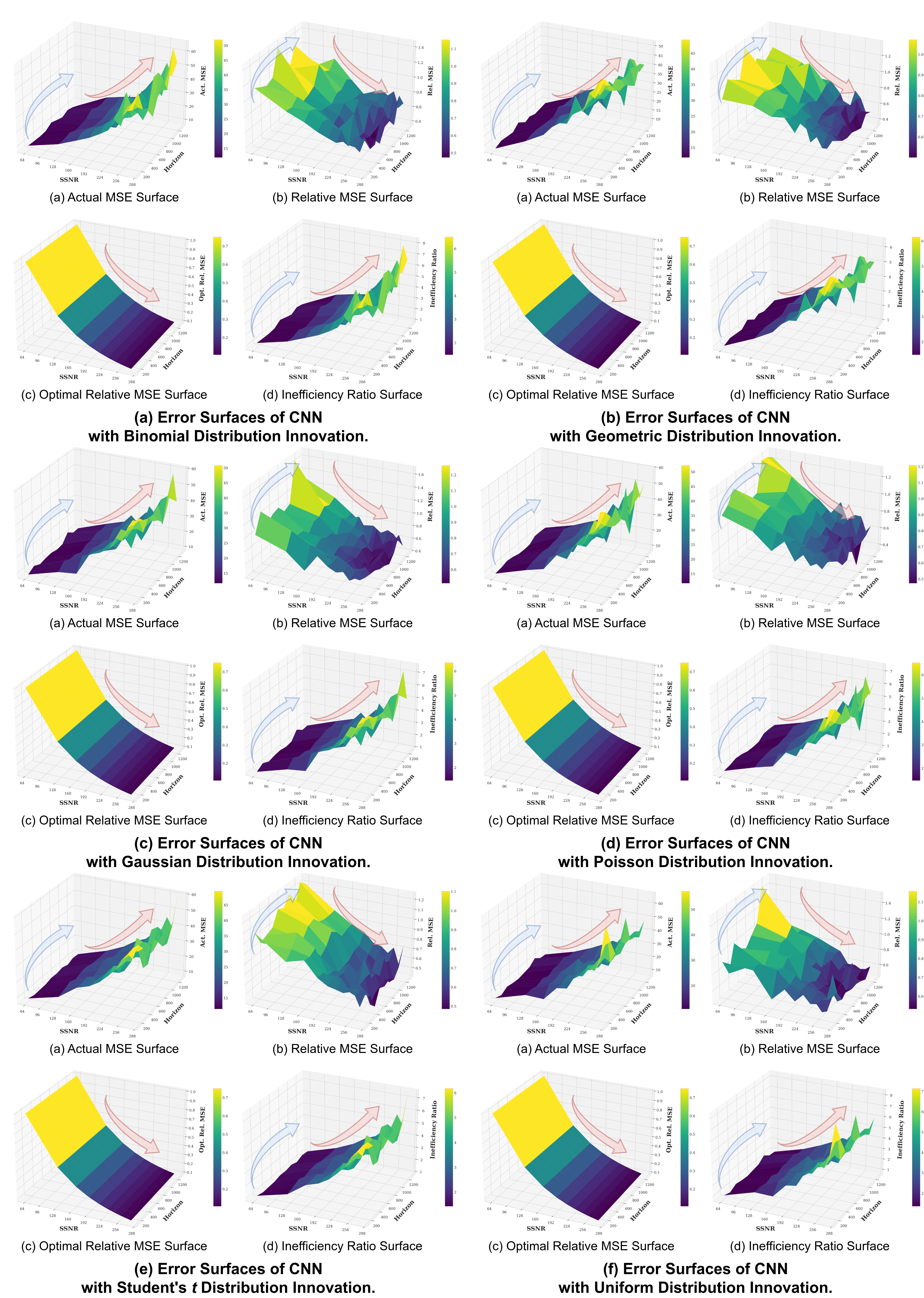}
    \caption{
        Empirical verification of EOB Theory via CNN model. The blue and red arrows indicates the surface variation trend along horizon ($h$) and the total SSNR ($\ssnr_x$), respectively.
    }
    \label{app_fig:CNN_err_surf}
\end{figure}

\begin{figure}
    \centering
    \includegraphics[width=0.98\linewidth]{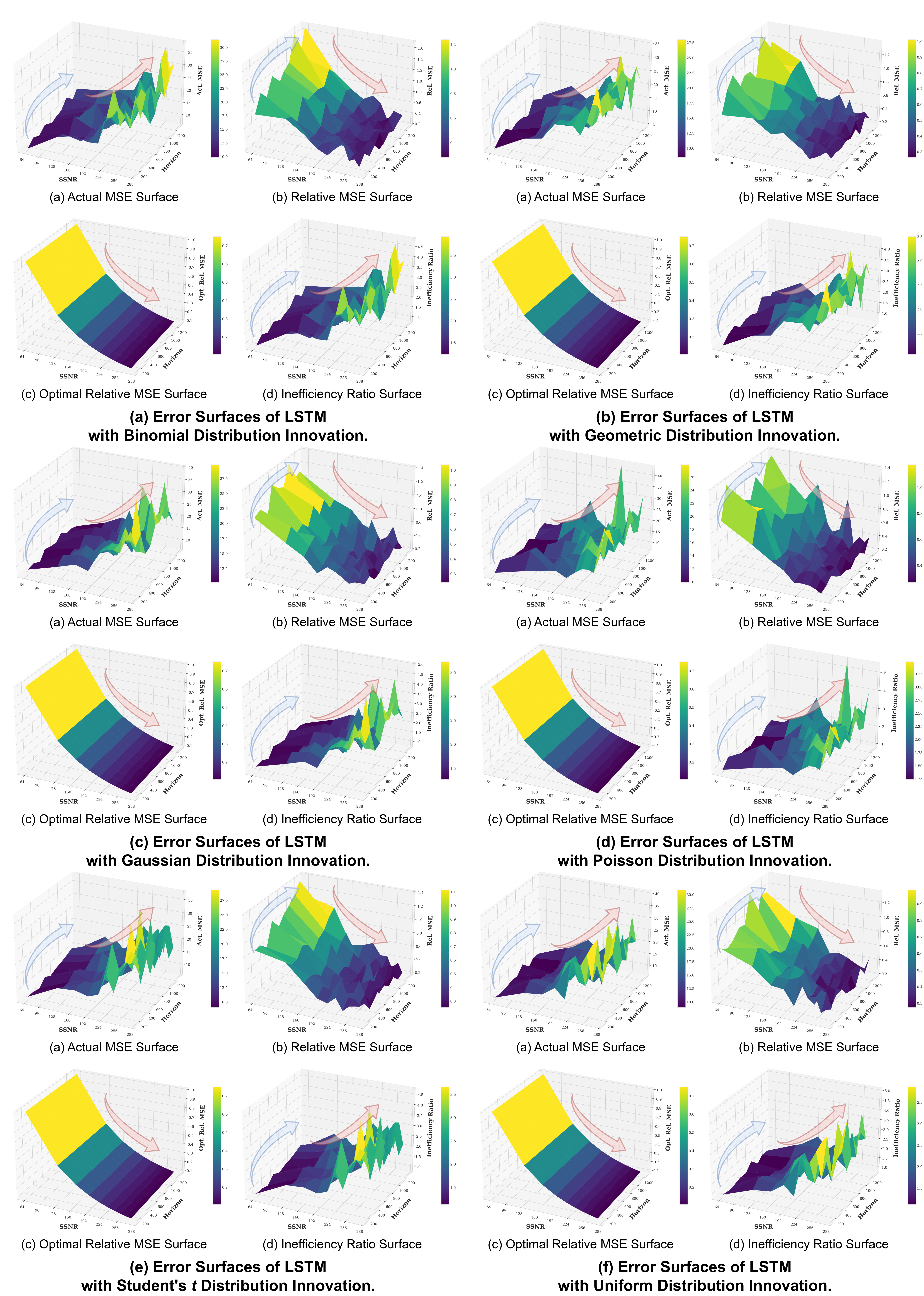}
    \caption{
        Empirical verification of EOB Theory via LSTM model. The blue and red arrows indicates the surface variation trend along horizon ($h$) and the total SSNR ($\ssnr_x$), respectively.
    }
    \label{app_fig:LSTM_err_surf}
\end{figure}

\begin{figure}
    \centering
    \includegraphics[width=0.98\linewidth]{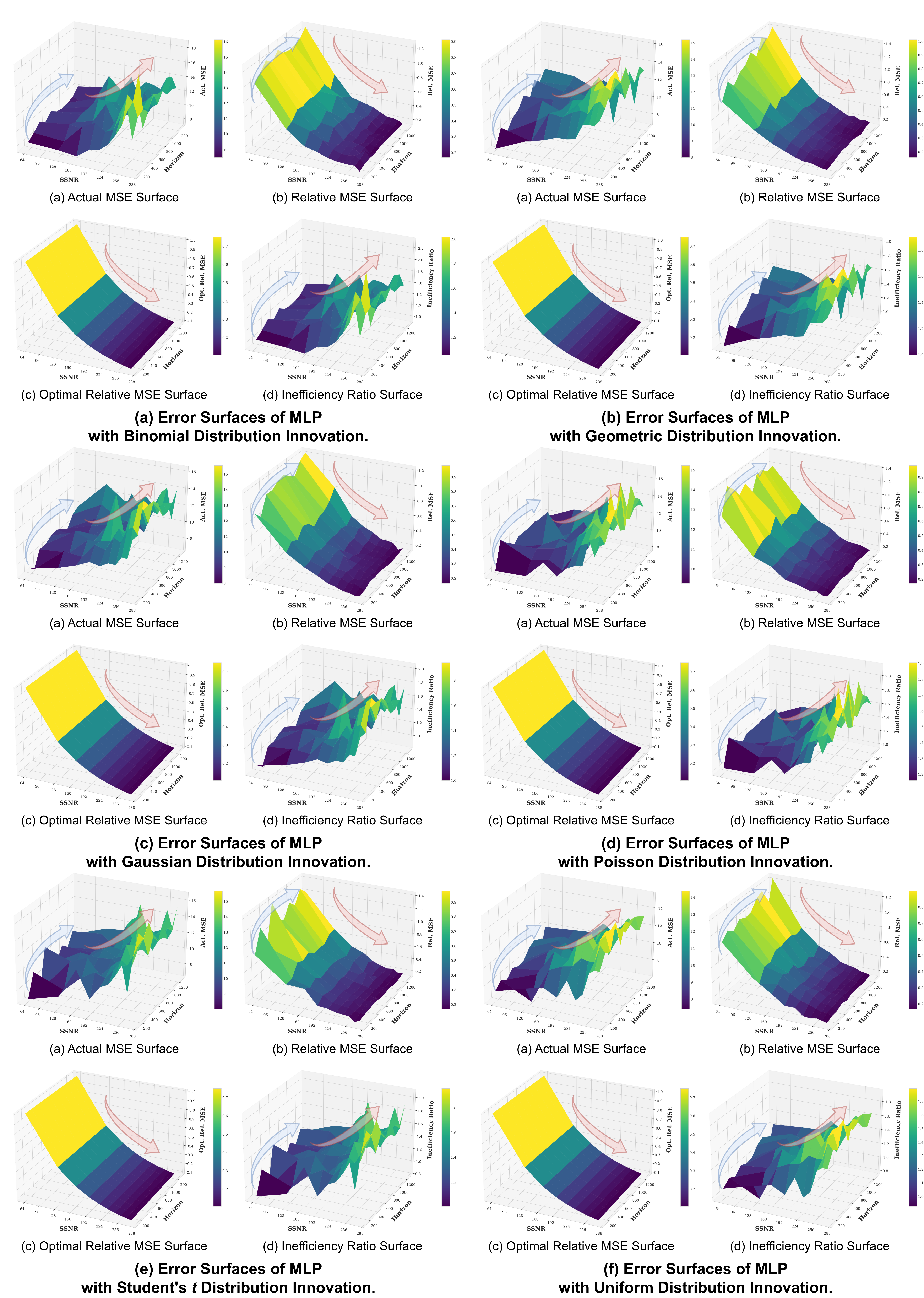}
    \caption{
        Empirical verification of EOB Theory via MLP model. The blue and red arrows indicates the surface variation trend along horizon ($h$) and the total SSNR ($\ssnr_x$), respectively.
    }
    \label{app_fig:MLP_err_surf}
\end{figure}

\begin{figure}
    \centering
    \includegraphics[width=0.98\linewidth]{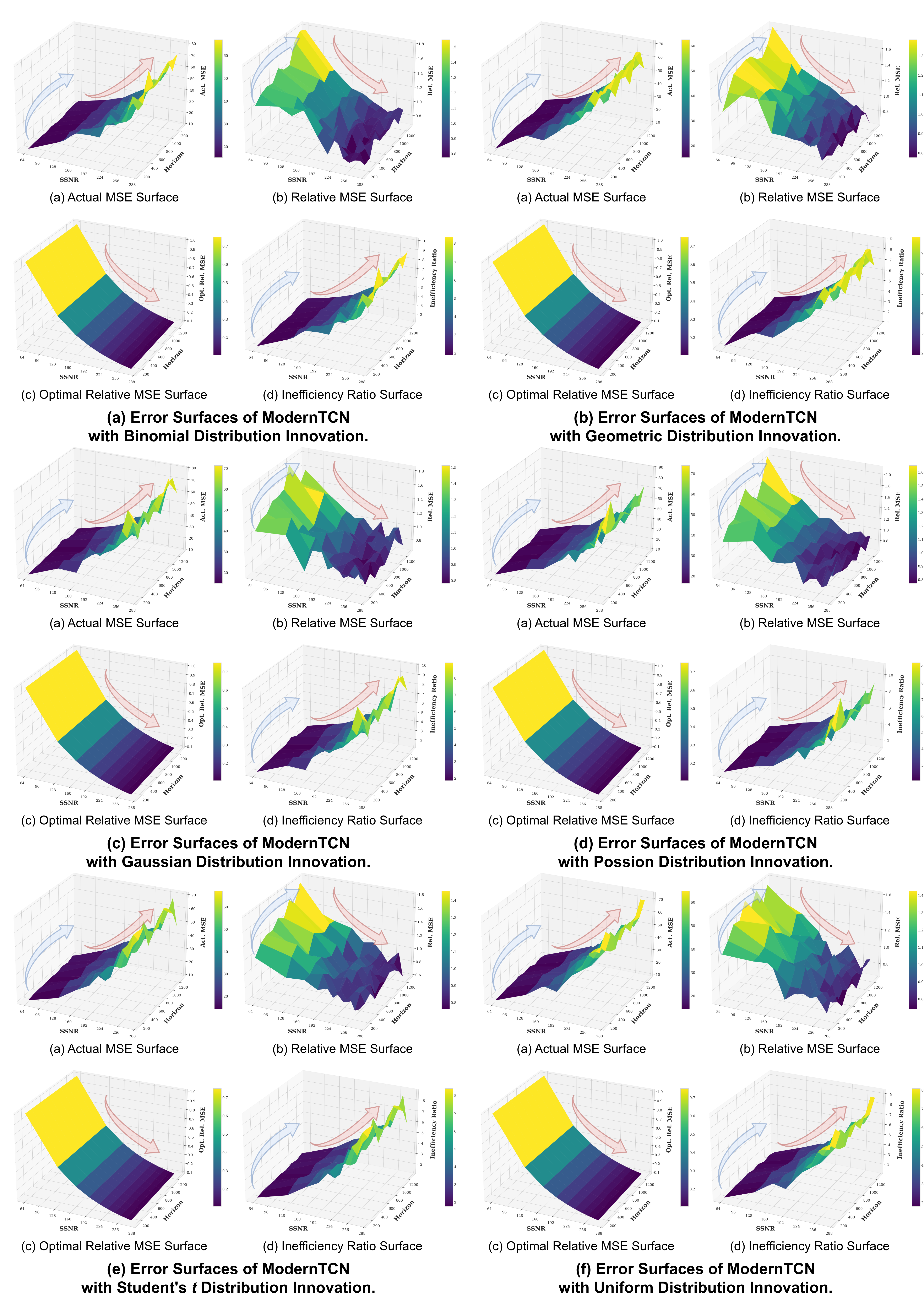}
    \caption{
        Empirical verification of EOB Theory via ModernTCN model. The blue and red arrows indicates the surface variation trend along horizon ($h$) and the total SSNR ($\ssnr_x$), respectively.
    }
    \label{app_fig:ModernTCN_err_surf}
\end{figure}

\begin{figure}
    \centering
    \includegraphics[width=0.98\linewidth]{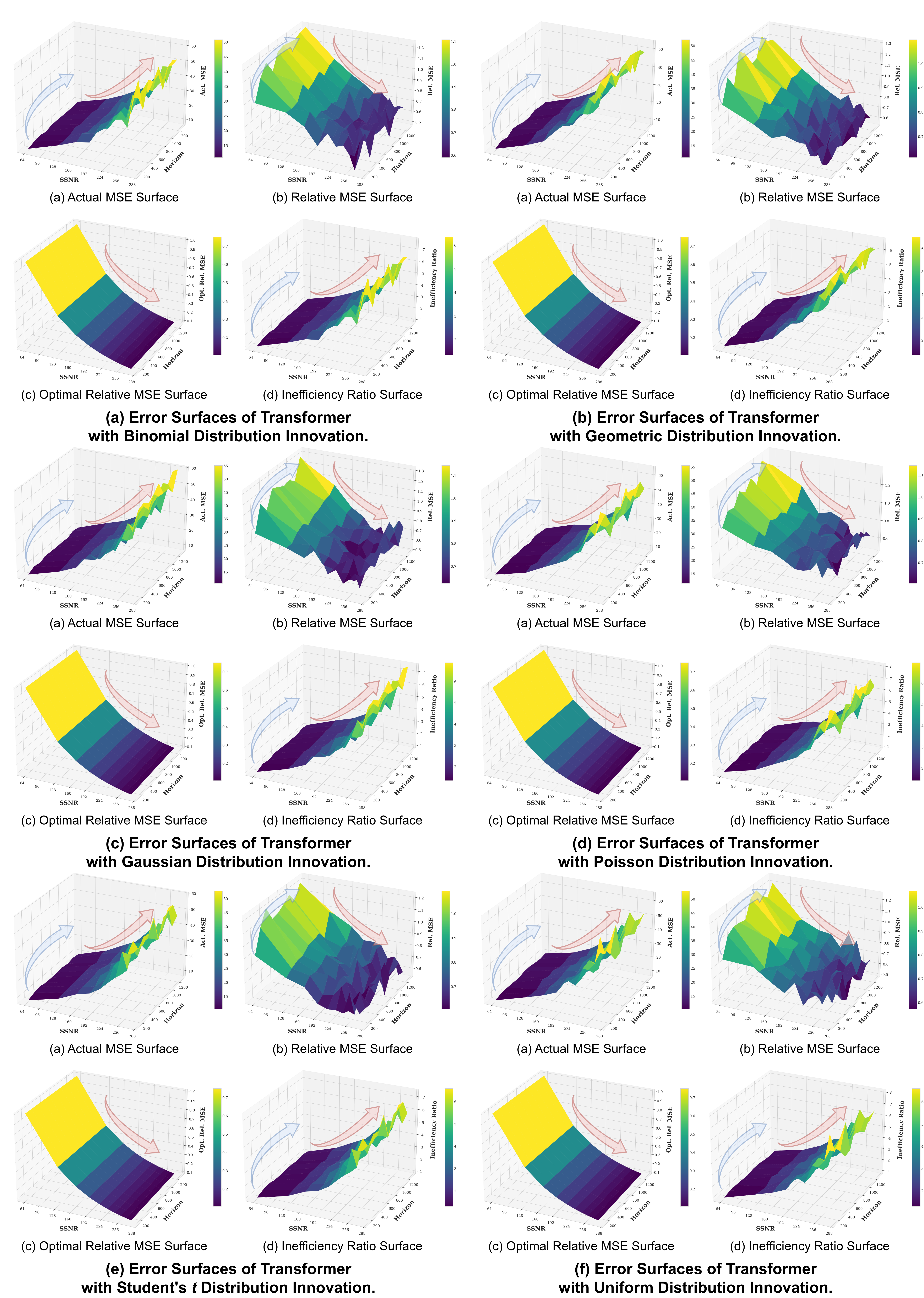}
    \caption{
        Empirical verification of EOB Theory via Transformer model. The blue and red arrows indicates the surface variation trend along horizon ($h$) and the total SSNR ($\ssnr_x$), respectively.
    }
    \label{app_fig:Transformer_err_surf}
\end{figure}

To test whether the EOB prediction depends on a specific noise model or architecture, we experiment with \textit{six distinct innovation noise distributions}: Binomial, Geometric, Gaussian, Poisson, Student's $t$, and Continuous Uniform.
We test these across \textit{five fundamental architectures}: CNN, LSTM, MLP, ModernTCN, and Transformer.
The results for each architecture--innovation combination are presented in Figures~\ref{app_fig:CNN_err_surf}, \ref{app_fig:LSTM_err_surf}, \ref{app_fig:MLP_err_surf}, \ref{app_fig:ModernTCN_err_surf}, and \ref{app_fig:Transformer_err_surf}, displaying (a) actual MSE, (b) relative MSE, (c) optimal relative MSE, and (d) inefficiency ratio surfaces.

Across these settings, the inefficiency ratio consistently increases with SSNR and saturates along the horizon, matching the asymptotic behavior predicted by Eq.~\eqref{eq:EOB_asym_rate}.
This supports the claim that the observed gap is induced by the point-wise objective rather than by a particular model family or innovation distribution.

Detailed error-dynamics analysis is provided in Section~\ref{sec:EOB_verif} of the main text.

\newpage
\subsection{Mechanistic Insight on Trigonometric Series}
\label{app:J2}
\label{app:insight_exp}

\begin{figure}
    \centering
    \begin{minipage}{0.49\linewidth}
        \centering
        \includegraphics[width=\linewidth]{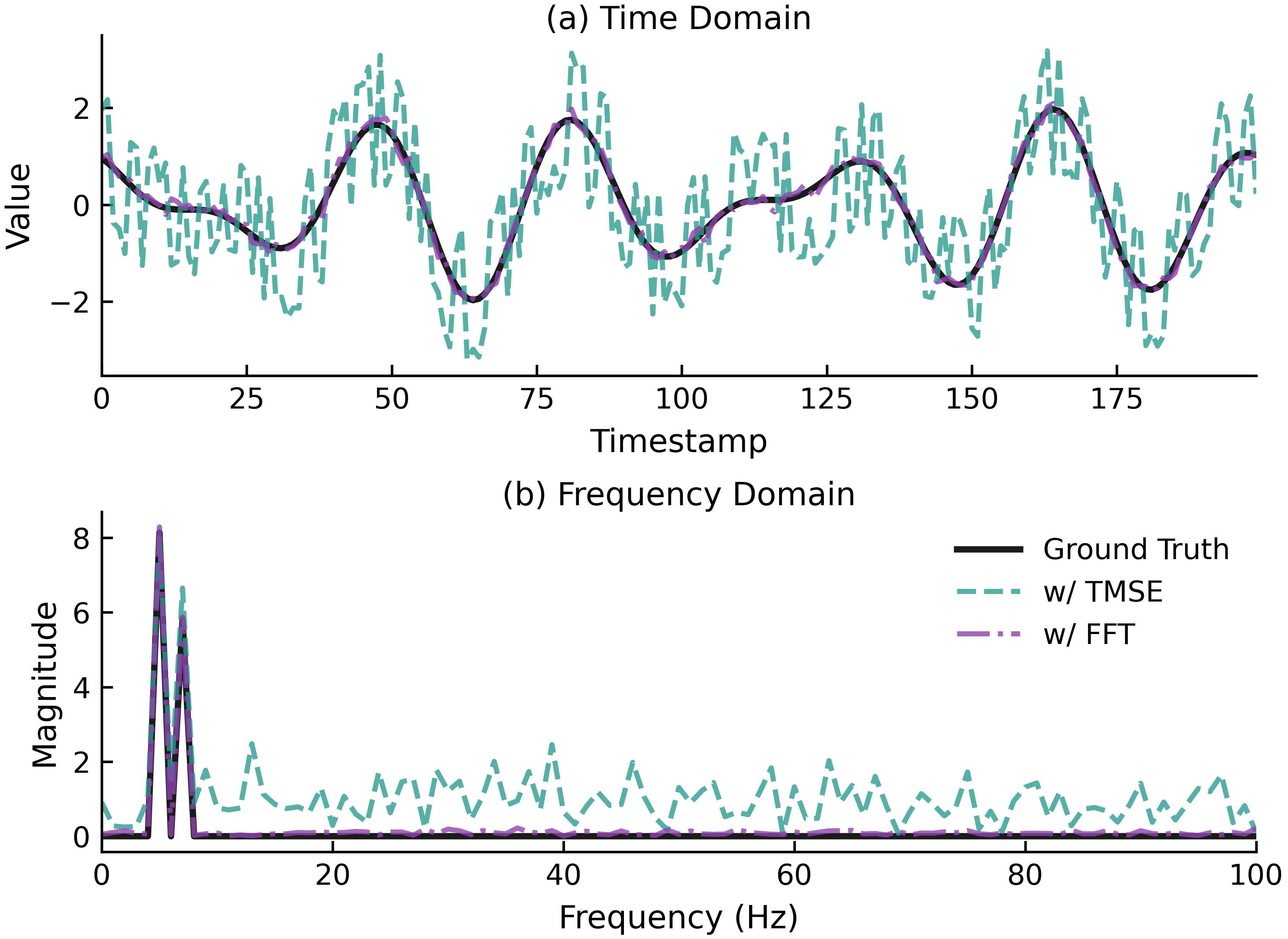}
        \vspace{-0.6em}
        \centerline{\small (a) harmonics $k=2$}
    \end{minipage}
    \hfill
    \begin{minipage}{0.49\linewidth}
        \centering
        \includegraphics[width=\linewidth]{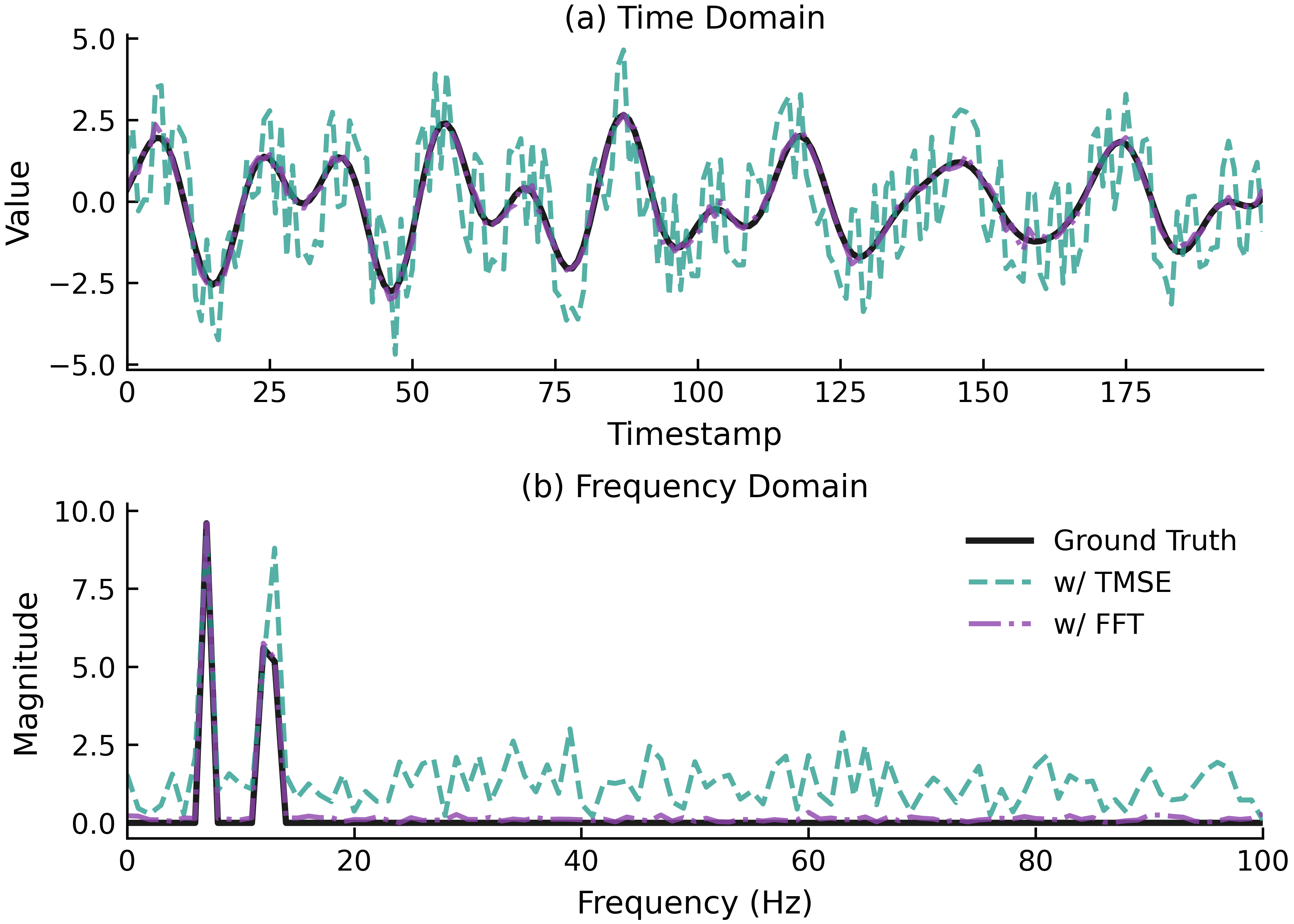}
        \vspace{-0.6em}
        \centerline{\small (b) harmonics $k=3$}
    \end{minipage}
    \vspace{0.4em}

    \begin{minipage}{0.49\linewidth}
        \centering
        \includegraphics[width=\linewidth]{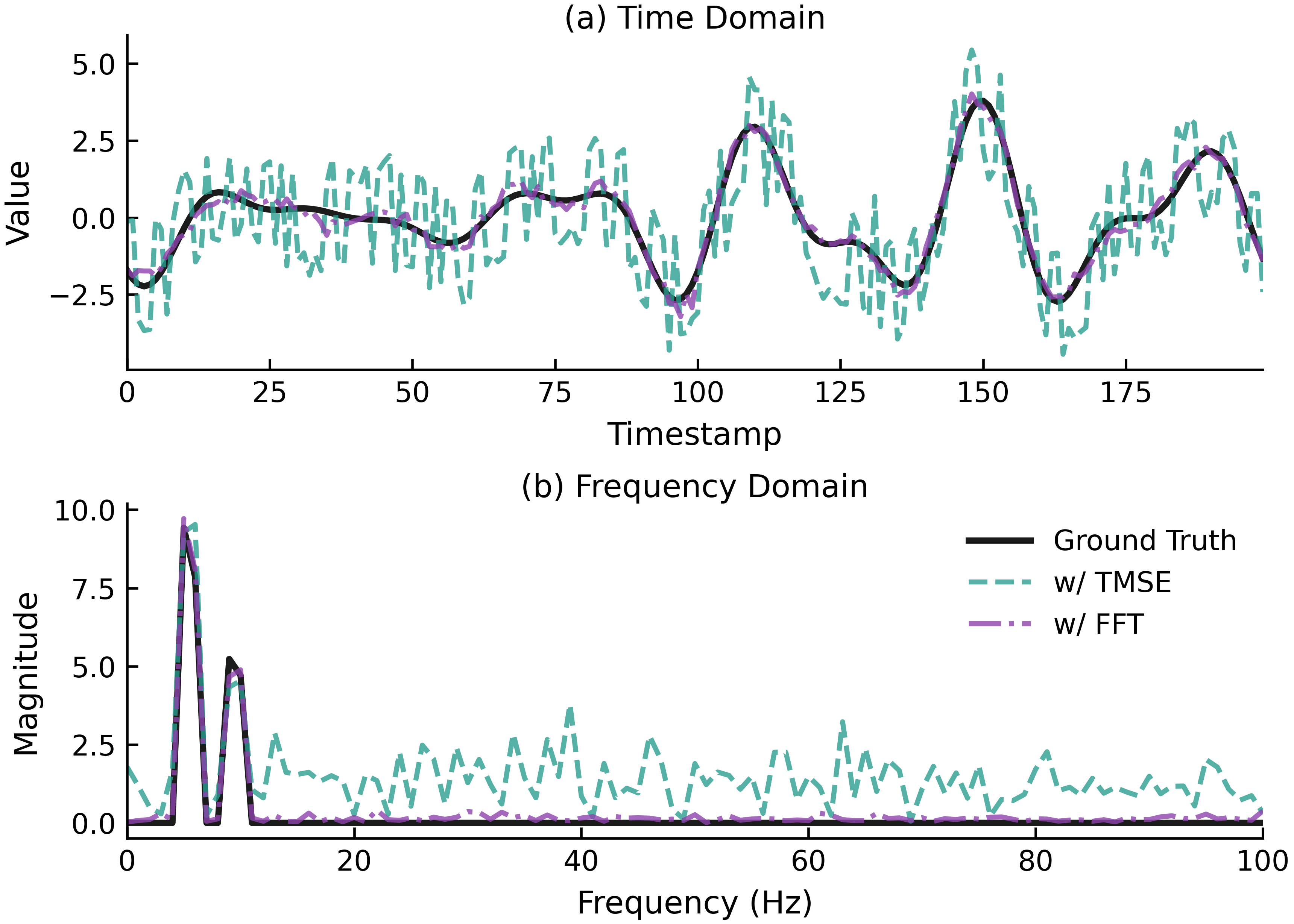}
        \vspace{-0.6em}
        \centerline{\small (c) harmonics $k=4$}
    \end{minipage}
    \hfill
    \begin{minipage}{0.49\linewidth}
        \centering
        \includegraphics[width=\linewidth]{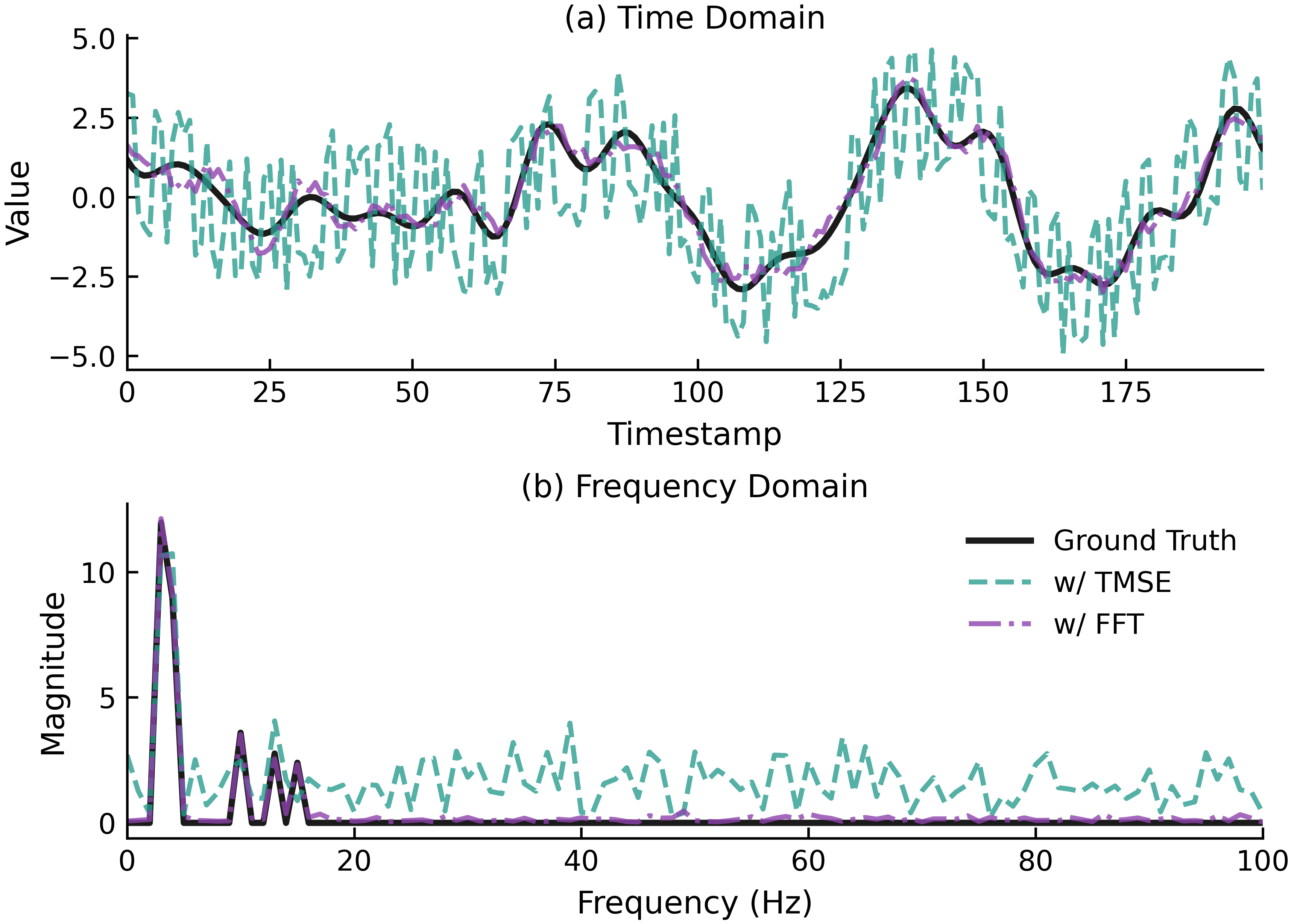}
        \vspace{-0.6em}
        \centerline{\small (d) harmonics $k=5$}
    \end{minipage}
    \caption{Additional insight experiments on pure trigonometric series with harmonics $k=2,\dots,5$. Temporal MSE produces high-frequency artifacts, while FFT amplitude--phase supervision recovers the periodic structure consistently.}
    \label{fig:InsightExp}
\end{figure}

\paragraph{Deconstructing the Empirical Misdiagnosis.}
The failure of neural networks to learn simple periodic functions has often been attributed to architectural limitations, such as the lack of periodic inductive bias or the spectral bias of ReLU-based networks~\citep{NEURIPS2020_11604531}.
Our experiment on pure trigonometric series (Figure~\ref{fig:InsightExp}) provides a complementary explanation.

Temporal MSE captures the coarse trend but produces locally distorted waveforms and broad frequency leakage across bins.
This behavior appears despite the model's ability to represent such smooth periodic functions.
The result indicates that the difficulty is not solely architectural: point-wise MSE treats each timestamp as an independent target and does not explicitly constrain phase-coherent dependence across the sequence.

\paragraph{Empirical Validation of the Paradigm Paradox.}
Trigonometric series represent an extreme high-predictability setting for EOB theory.
As deterministic periodic processes, they have very high temporal SSNR and are therefore strongly penalized by point-wise temporal supervision according to the Paradigm Paradox (Theorem~\ref{thm:paradigm_paradox}).

The temporal MSE predictions in Figure~\ref{fig:InsightExp} show this effect empirically: the time-domain outputs retain the coarse period but introduce high-frequency artifacts.

In contrast, FFT amplitude--phase supervision nearly overlaps with the ground truth and aligns with the true frequency without visible high-frequency noise.
By optimizing approximately orthogonal amplitude and phase coordinates, the supervised targets become close to statistically independent and the frequency-domain SSNR approaches $1$.
This supports the main-text conclusion that the trigonometric failure case is better explained as an objective-induced bias than as insufficient model expressiveness.

\subsection{Plug-and-Play Versatility}
\label{app:J3}
\label{app:plug-and-play}

We evaluate the plug-and-play versatility of $\mcL_{\text{Harm}, \ell_p}$ across five representative architectures: TimeMixer (MLP-based), iTransformer and PatchTST (Transformer-based), TimesNet (CNN-based), and DLinear (Linear-based).
The results in Table~\ref{tab:ablation_plug_and_play} test whether the benefit comes from the supervision objective rather than from changing the modeling module.

\paragraph{Architecture-Level Transfer.}
Across the \textit{Avg} rows, $\mcL_{\text{Harm}, \ell_1}$ improves 55 out of 60 dataset--backbone--metric comparisons and ties in 2 additional cases.
Because the backbone architecture is unchanged, these gains isolate the effect of replacing temporal MSE with the harmonized objective.
This supports the claim that EOB is induced at the point-wise supervision level.

\paragraph{Robustness across All Prediction Horizons.}
Across all tested models, the advantage of $\mcL_{\text{Harm}, \ell_1}$ remains stable as the horizon increases from 96 to 720.
This matches the theory: structural orthogonalization reduces the effective SSNR of the supervised target and therefore reduces the per-step bias accumulated over the output window.

\paragraph{Versatility Across Diverse Data Characteristics.}
The plug-and-play results cover energy, weather, and power-market benchmarks with different temporal resolutions and dependence patterns.
The consistent average gains indicate that the harmonized loss is not tied to a single dataset or architecture.

\begin{table*}[h]
\centering
\caption{
    \textbf{Plug-and-Play Versatility of Harmonized $\ell_1$ Norm.}
    \bst{Bold} typeface denotes the champion performance.    
}
\label{tab:ablation_plug_and_play}
\resizebox{0.95\linewidth}{!}{
\begin{threeparttable}
\begin{tabular}{c|c|cc|cc|cc|cc|cc|cc|cc|cc|cc|cc}
    \toprule
    \multicolumn{2}{c}{\scaleb{Models}} & 
    \multicolumn{4}{c}{\scaleb{TimeMixer}} & \multicolumn{4}{c}{\scaleb{iTransformer}} &
    \multicolumn{4}{c}{\scaleb{PatchTST}} & \multicolumn{4}{c}{\scaleb{TimesNet}} &
    \multicolumn{4}{c}{\scalea{DLinear}} \\

    \cmidrule(lr){3-6} \cmidrule(lr){7-10} \cmidrule(lr){11-14} \cmidrule(lr){15-18} \cmidrule(lr){19-22}

    \multicolumn{2}{c}{\scaleb{Loss}} & 
    \multicolumn{2}{c}{$\mcL_{\text{Harm}, \ell_1}$} & \multicolumn{2}{c}{$\mcL_{\text{TMSE}}$} & 
    \multicolumn{2}{c}{$\mcL_{\text{Harm}, \ell_1}$} & \multicolumn{2}{c}{$\mcL_{\text{TMSE}}$} & 
    \multicolumn{2}{c}{$\mcL_{\text{Harm}, \ell_1}$} & \multicolumn{2}{c}{$\mcL_{\text{TMSE}}$} & 
    \multicolumn{2}{c}{$\mcL_{\text{Harm}, \ell_1}$} & \multicolumn{2}{c}{$\mcL_{\text{TMSE}}$} & 
    \multicolumn{2}{c}{$\mcL_{\text{Harm}, \ell_1}$} & \multicolumn{2}{c}{$\mcL_{\text{TMSE}}$} \\

    \cmidrule(lr){3-4} \cmidrule(lr){5-6} \cmidrule(lr){7-8} \cmidrule(lr){9-10}
    \cmidrule(lr){11-12} \cmidrule(lr){13-14} \cmidrule(lr){15-16} \cmidrule(lr){17-18}
    \cmidrule(lr){19-20} \cmidrule(lr){21-22}
    
    \multicolumn{2}{c}{\scaleb{Metrics}} & 
    \scalea{MSE} & \scalea{MAE} & \scalea{MSE} & \scalea{MAE} &
    \scalea{MSE} & \scalea{MAE} & \scalea{MSE} & \scalea{MAE} &
    \scalea{MSE} & \scalea{MAE} & \scalea{MSE} & \scalea{MAE} &
    \scalea{MSE} & \scalea{MAE} & \scalea{MSE} & \scalea{MAE} &
    \scalea{MSE} & \scalea{MAE} & \scalea{MSE} & \scalea{MAE} \\
    \toprule

    \multirow{5}{*}{{\rotatebox{90}{\scalebox{0.95}{ETTh1}}}}
    & \scalea{96} & 
    \bst{\scalea{0.381}} & \bst{\scalea{0.389}} & \scalea{0.382} & \scalea{0.399} & 
    \bst{\scalea{0.378}} & \bst{\scalea{0.397}} & \scalea{0.393} & \scalea{0.408} & 
    \bst{\scalea{0.379}} & \bst{\scalea{0.389}} & \scalea{0.382} & \scalea{0.400} & 
    \bst{\scalea{0.412}} & \bst{\scalea{0.421}} & \bst{\scalea{0.412}} & \scalea{0.428} & 
    \bst{\scalea{0.390}} & \bst{\scalea{0.403}} & \scalea{0.396} & \scalea{0.411}\\ 
    & \scalea{192} & 
    \bst{\scalea{0.430}} & \bst{\scalea{0.417}} & \scalea{0.437} & \scalea{0.430} & 
    \bst{\scalea{0.429}} & \bst{\scalea{0.426}} & \scalea{0.447} & \scalea{0.440} & 
    \bst{\scalea{0.427}} & \bst{\scalea{0.419}} & \scalea{0.430} & \scalea{0.433} & 
    \bst{\scalea{0.450}} & \bst{\scalea{0.439}} & \scalea{0.457} & \scalea{0.453} & 
    \bst{\scalea{0.444}} & \bst{\scalea{0.438}} & \scalea{0.446} & \scalea{0.442}\\ 
    & \scalea{336} & 
    \bst{\scalea{0.485}} & \bst{\scalea{0.446}} & \bst{\scalea{0.485}} & \scalea{0.452} & 
    \bst{\scalea{0.464}} & \bst{\scalea{0.442}} & \scalea{0.489} & \scalea{0.463} & 
    \bst{\scalea{0.464}} & \bst{\scalea{0.438}} & \scalea{0.474} & \scalea{0.460} & 
    \scalea{0.508} & \bst{\scalea{0.472}} & \bst{\scalea{0.506}} & \scalea{0.479} & 
    \bst{\scalea{0.480}} & \bst{\scalea{0.456}} & \scalea{0.488} & \scalea{0.466}\\ 
    & \scalea{720} & 
    \bst{\scalea{0.495}} & \bst{\scalea{0.466}} & \scalea{0.518} & \scalea{0.488} & 
    \bst{\scalea{0.474}} & \bst{\scalea{0.472}} & \scalea{0.514} & \scalea{0.499} & 
    \bst{\scalea{0.460}} & \bst{\scalea{0.460}} & \scalea{0.529} & \scalea{0.508} & 
    \bst{\scalea{0.510}} & \bst{\scalea{0.489}} & \scalea{0.523} & \scalea{0.498} & 
    \bst{\scalea{0.494}} & \bst{\scalea{0.495}} & \scalea{0.512} & \scalea{0.510}\\ 
    \cmidrule(lr){2-22}
    & \scalea{Avg} & 
    \bst{\scalea{0.448}} & \bst{\scalea{0.430}} & \scalea{0.455} & \scalea{0.443} & 
    \bst{\scalea{0.436}} & \bst{\scalea{0.434}} & \scalea{0.461} & \scalea{0.453} & 
    \bst{\scalea{0.433}} & \bst{\scalea{0.426}} & \scalea{0.454} & \scalea{0.450} & 
    \bst{\scalea{0.470}} & \bst{\scalea{0.455}} & \scalea{0.475} & \scalea{0.464} & 
    \bst{\scalea{0.452}} & \bst{\scalea{0.448}} & \scalea{0.461} & \scalea{0.457}\\ 
    \midrule

    \multirow{5}{*}{{\rotatebox{90}{\scalebox{0.95}{ETTm1}}}}
    & \scalea{96} & 
    \bst{\scalea{0.313}} & \bst{\scalea{0.342}} & \scalea{0.322} & \scalea{0.360} & 
    \bst{\scalea{0.314}} & \bst{\scalea{0.346}} & \scalea{0.349} & \scalea{0.379} & 
    \bst{\scalea{0.326}} & \bst{\scalea{0.353}} & \scalea{0.327} & \scalea{0.367} & 
    \bst{\scalea{0.327}} & \bst{\scalea{0.359}} & \scalea{0.334} & \scalea{0.375} & 
    \bst{\scalea{0.340}} & \bst{\scalea{0.360}} & \scalea{0.355} & \scalea{0.379}\\ 
    & \scalea{192} & 
    \bst{\scalea{0.365}} & \bst{\scalea{0.368}} & \bst{\scalea{0.365}} & \scalea{0.385} & 
    \bst{\scalea{0.369}} & \bst{\scalea{0.373}} & \scalea{0.388} & \scalea{0.397} & 
    \scalea{0.373} & \bst{\scalea{0.377}} & \bst{\scalea{0.371}} & \scalea{0.391} & 
    \bst{\scalea{0.375}} & \bst{\scalea{0.385}} & \scalea{0.389} & \scalea{0.403} & 
    \bst{\scalea{0.381}} & \bst{\scalea{0.380}} & \scalea{0.389} & \scalea{0.396}\\ 
    & \scalea{336} & 
    \bst{\scalea{0.394}} & \bst{\scalea{0.390}} & \scalea{0.395} & \scalea{0.407} & 
    \bst{\scalea{0.411}} & \bst{\scalea{0.400}} & \scalea{0.424} & \scalea{0.420} & 
    \scalea{0.407} & \bst{\scalea{0.398}} & \bst{\scalea{0.399}} & \scalea{0.408} & 
    \bst{\scalea{0.409}} & \bst{\scalea{0.409}} & \scalea{0.417} & \scalea{0.423} & 
    \bst{\scalea{0.412}} & \bst{\scalea{0.402}} & \scalea{0.419} & \scalea{0.418}\\ 
    & \scalea{720} & 
    \bst{\scalea{0.458}} & \bst{\scalea{0.429}} & \bst{\scalea{0.458}} & \scalea{0.444} & 
    \bst{\scalea{0.479}} & \bst{\scalea{0.439}} & \scalea{0.494} & \scalea{0.459} & 
    \scalea{0.465} & \bst{\scalea{0.432}} & \bst{\scalea{0.460}} & \scalea{0.446} & 
    \scalea{0.491} & \bst{\scalea{0.454}} & \bst{\scalea{0.490}} & \scalea{0.461} & 
    \bst{\scalea{0.472}} & \bst{\scalea{0.439}} & \scalea{0.477} & \scalea{0.454}\\ 
    \cmidrule(lr){2-22}
    & \scalea{Avg} & 
    \bst{\scalea{0.383}} & \bst{\scalea{0.382}} & \scalea{0.385} & \scalea{0.399} & 
    \bst{\scalea{0.393}} & \bst{\scalea{0.390}} & \scalea{0.414} & \scalea{0.414} & 
    \scalea{0.393} & \bst{\scalea{0.390}} & \bst{\scalea{0.390}} & \scalea{0.403} & 
    \bst{\scalea{0.401}} & \bst{\scalea{0.402}} & \scalea{0.407} & \scalea{0.416} & 
    \bst{\scalea{0.401}} & \bst{\scalea{0.395}} & \scalea{0.410} & \scalea{0.412}\\ 
    \midrule

    \multirow{5}{*}{{\rotatebox{90}{\scalebox{0.95}{ECL}}}}
    & \scalea{96} & 
    \bst{\scalea{0.158}} & \bst{\scalea{0.245}} & \scalea{0.216} & \scalea{0.318} & 
    \bst{\scalea{0.144}} & \bst{\scalea{0.233}} & \scalea{0.148} & \scalea{0.240} & 
    \bst{\scalea{0.178}} & \bst{\scalea{0.268}} & \scalea{0.180} & \scalea{0.272} & 
    \bst{\scalea{0.168}} & \bst{\scalea{0.266}} & \scalea{0.171} & \scalea{0.276} & 
    \bst{\scalea{0.247}} & \bst{\scalea{0.337}} & \scalea{0.252} & \scalea{0.347} \\ 
    & \scalea{192} & 
    \bst{\scalea{0.170}} & \bst{\scalea{0.256}} & \scalea{0.225} & \scalea{0.333} & 
    \bst{\scalea{0.158}} & \bst{\scalea{0.247}} & \scalea{0.165} & \scalea{0.256} & 
    \bst{\scalea{0.187}} & \bst{\scalea{0.275}} & \scalea{0.192} & \scalea{0.282} & 
    \bst{\scalea{0.183}} & \bst{\scalea{0.279}} & \scalea{0.190} & \scalea{0.293} & 
    \bst{\scalea{0.247}} & \bst{\scalea{0.341}} & \scalea{0.252} & \scalea{0.350} \\ 
    & \scalea{336} & 
    \bst{\scalea{0.188}} & \bst{\scalea{0.245}} & \scalea{0.247} & \scalea{0.355} & 
    \bst{\scalea{0.172}} & \bst{\scalea{0.263}} & \scalea{0.179} & \scalea{0.272} & 
    \bst{\scalea{0.200}} & \bst{\scalea{0.291}} & \scalea{0.208} & \scalea{0.298} & 
    \bst{\scalea{0.198}} & \bst{\scalea{0.294}} & \scalea{0.206} & \scalea{0.309} & 
    \bst{\scalea{0.259}} & \bst{\scalea{0.352}} & \scalea{0.264} & \scalea{0.362} \\ 
    & \scalea{720} & 
    \bst{\scalea{0.229}} & \bst{\scalea{0.308}} & \scalea{0.286} & \scalea{0.383} & 
    \bst{\scalea{0.206}} & \bst{\scalea{0.296}} & \scalea{0.213} & \scalea{0.301} & 
    \bst{\scalea{0.247}} & \bst{\scalea{0.321}} & \scalea{0.250} & \scalea{0.330} & 
    \bst{\scalea{0.283}} & \bst{\scalea{0.347}} & \scalea{0.289} & \scalea{0.363} & 
    \bst{\scalea{0.292}} & \bst{\scalea{0.378}} & \scalea{0.298} & \scalea{0.389} \\ 
    \cmidrule(lr){2-22}
    & \scalea{Avg} & 
    \bst{\scalea{0.186}} & \bst{\scalea{0.264}} & \scalea{0.243} & \scalea{0.347} & 
    \bst{\scalea{0.170}} & \bst{\scalea{0.260}} & \scalea{0.176} & \scalea{0.267} & 
    \bst{\scalea{0.203}} & \bst{\scalea{0.289}} & \scalea{0.208} & \scalea{0.296} & 
    \bst{\scalea{0.208}} & \bst{\scalea{0.297}} & \scalea{0.214} & \scalea{0.310} & 
    \bst{\scalea{0.261}} & \bst{\scalea{0.352}} & \scalea{0.266} & \scalea{0.362} \\ 
    \midrule

    \multirow{5}{*}{{\rotatebox{90}{\scalebox{0.95}{BE}}}}
    & \scalea{96} & 
    \bst{\scalea{0.225}} & \bst{\scalea{0.241}} & \scalea{0.241} & \scalea{0.261} & 
    \bst{\scalea{0.207}} & \bst{\scalea{0.226}} & \scalea{0.251} & \scalea{0.262} & 
    \bst{\scalea{0.235}} & \bst{\scalea{0.258}} & \scalea{0.248} & \scalea{0.273} & 
    \bst{\scalea{0.208}} & \bst{\scalea{0.230}} & \scalea{0.259} & \scalea{0.281} & 
    \bst{\scalea{0.313}} & \bst{\scalea{0.327}} & \bst{\scalea{0.313}} & \scalea{0.336} \\ 
    & \scalea{192} & 
    \bst{\scalea{0.265}} & \bst{\scalea{0.277}} & \scalea{0.276} & \scalea{0.289} & 
    \bst{\scalea{0.255}} & \bst{\scalea{0.264}} & \scalea{0.299} & \scalea{0.298} & 
    \bst{\scalea{0.274}} & \bst{\scalea{0.288}} & \scalea{0.280} & \scalea{0.296} & 
    \bst{\scalea{0.257}} & \bst{\scalea{0.270}} & \scalea{0.293} & \scalea{0.306} & 
    \scalea{0.331} & \bst{\scalea{0.332}} & \bst{\scalea{0.328}} & \scalea{0.339} \\ 
    & \scalea{336} & 
    \bst{\scalea{0.298}} & \bst{\scalea{0.312}} & \scalea{0.307} & \scalea{0.322} & 
    \bst{\scalea{0.297}} & \bst{\scalea{0.300}} & \scalea{0.352} & \scalea{0.342} & 
    \bst{\scalea{0.315}} & \bst{\scalea{0.326}} & \scalea{0.321} & \scalea{0.333} & 
    \bst{\scalea{0.296}} & \bst{\scalea{0.304}} & \scalea{0.326} & \scalea{0.336} & 
    \scalea{0.371} & \bst{\scalea{0.362}} & \bst{\scalea{0.368}} & \scalea{0.370} \\ 
    & \scalea{720} & 
    \bst{\scalea{0.361}} & \bst{\scalea{0.361}} & \scalea{0.378} & \scalea{0.380} & 
    \bst{\scalea{0.421}} & \bst{\scalea{0.383}} & \scalea{0.424} & \scalea{0.393} & 
    \bst{\scalea{0.392}} & \bst{\scalea{0.386}} & \scalea{0.401} & \scalea{0.396} & 
    \bst{\scalea{0.356}} & \bst{\scalea{0.338}} & \scalea{0.385} & \scalea{0.382} & 
    \scalea{0.448} & \bst{\scalea{0.418}} & \bst{\scalea{0.445}} & \scalea{0.428} \\ 
    \cmidrule(lr){2-22}
    & \scalea{Avg} & 
    \bst{\scalea{0.287}} & \bst{\scalea{0.298}} & \scalea{0.300} & \scalea{0.313} & 
    \bst{\scalea{0.295}} & \bst{\scalea{0.293}} & \scalea{0.332} & \scalea{0.324} & 
    \bst{\scalea{0.304}} & \bst{\scalea{0.315}} & \scalea{0.313} & \scalea{0.324} & 
    \bst{\scalea{0.279}} & \bst{\scalea{0.286}} & \scalea{0.316} & \scalea{0.326} & 
    \scalea{0.366} & \bst{\scalea{0.360}} & \bst{\scalea{0.363}} & \scalea{0.368} \\ 
    \midrule

    \multirow{5}{*}{{\rotatebox{90}{\scalebox{0.95}{Weather}}}}
    & \scalea{96} & 
    \bst{\scalea{0.161}} & \bst{\scalea{0.203}} & \scalea{0.163} & \scalea{0.210} & 
    \bst{\scalea{0.174}} & \bst{\scalea{0.209}} & \scalea{0.176} & \scalea{0.217} & 
    \bst{\scalea{0.175}} & \bst{\scalea{0.210}} & \scalea{0.178} & \scalea{0.219} & 
    \bst{\scalea{0.171}} & \bst{\scalea{0.219}} & \scalea{0.172} & \scalea{0.221} & 
    \scalea{0.201} & \bst{\scalea{0.254}} & \bst{\scalea{0.196}} & \scalea{0.257} \\ 
    & \scalea{192} & 
    \bst{\scalea{0.207}} & \bst{\scalea{0.246}} & \scalea{0.208} & \scalea{0.251} & 
    \bst{\scalea{0.220}} & \bst{\scalea{0.250}} & \scalea{0.225} & \scalea{0.258} & 
    \bst{\scalea{0.222}} & \bst{\scalea{0.251}} & \scalea{0.224} & \scalea{0.259} & 
    \bst{\scalea{0.231}} & \bst{\scalea{0.267}} & \bst{\scalea{0.231}} & \scalea{0.270} & 
    \bst{\scalea{0.237}} & \bst{\scalea{0.289}} & \bst{\scalea{0.237}} & \scalea{0.296} \\ 
    & \scalea{336} & 
    \bst{\scalea{0.262}} & \bst{\scalea{0.288}} & \scalea{0.264} & \scalea{0.292} & 
    \bst{\scalea{0.279}} & \bst{\scalea{0.292}} & \scalea{0.282} & \scalea{0.299} & 
    \bst{\scalea{0.279}} & \bst{\scalea{0.293}} & \scalea{0.280} & \scalea{0.299} & 
    \scalea{0.287} & \bst{\scalea{0.304}} & \bst{\scalea{0.284}} & \scalea{0.305} & 
    \bst{\scalea{0.282}} & \bst{\scalea{0.328}} & \scalea{0.283} & \scalea{0.333} \\ 
    & \scalea{720} & 
    \bst{\scalea{0.343}} & \bst{\scalea{0.342}} & \scalea{0.345} & \scalea{0.345} & 
    \bst{\scalea{0.355}} & \bst{\scalea{0.343}} & \scalea{0.359} & \scalea{0.350} & 
    \bst{\scalea{0.355}} & \bst{\scalea{0.342}} & \scalea{0.356} & \scalea{0.348} & 
    \bst{\scalea{0.357}} & \bst{\scalea{0.351}} & \scalea{0.360} & \scalea{0.355} & 
    \bst{\scalea{0.342}} & \bst{\scalea{0.375}} & \scalea{0.347} & \scalea{0.384} \\ 
    \cmidrule(lr){2-22}
    & \scalea{Avg} & 
    \bst{\scalea{0.243}} & \bst{\scalea{0.270}} & \scalea{0.245} & \scalea{0.275} & 
    \bst{\scalea{0.256}} & \bst{\scalea{0.272}} & \scalea{0.260} & \scalea{0.281} & 
    \bst{\scalea{0.258}} & \bst{\scalea{0.274}} & \scalea{0.260} & \scalea{0.281} & 
    \bst{\scalea{0.262}} & \bst{\scalea{0.285}} & \bst{\scalea{0.262}} & \scalea{0.288} & 
    \bst{\scalea{0.266}} & \bst{\scalea{0.312}} & \bst{\scalea{0.266}} & \scalea{0.317} \\ 
    \midrule

    \multirow{5}{*}{{\rotatebox{90}{\scalebox{0.95}{PJM}}}}
    & \scalea{96} & 
    \bst{\scalea{0.217}} & \bst{\scalea{0.296}} & \scalea{0.235} & \scalea{0.321} & 
    \bst{\scalea{0.202}} & \bst{\scalea{0.288}} & \scalea{0.222} & \scalea{0.309} & 
    \bst{\scalea{0.206}} & \bst{\scalea{0.293}} & \scalea{0.234} & \scalea{0.321} & 
    \bst{\scalea{0.199}} & \bst{\scalea{0.290}} & \scalea{0.231} & \scalea{0.309} & 
    \bst{\scalea{0.263}} & \bst{\scalea{0.345}} & \scalea{0.264} & \scalea{0.357} \\ 
    & \scalea{192} & 
    \bst{\scalea{0.276}} & \bst{\scalea{0.343}} & \scalea{0.294} & \scalea{0.362} & 
    \bst{\scalea{0.273}} & \bst{\scalea{0.342}} & \scalea{0.290} & \scalea{0.359} & 
    \bst{\scalea{0.268}} & \bst{\scalea{0.342}} & \scalea{0.289} & \scalea{0.359} & 
    \bst{\scalea{0.270}} & \bst{\scalea{0.343}} & \scalea{0.294} & \scalea{0.357} & 
    \scalea{0.313} & \bst{\scalea{0.378}} & \bst{\scalea{0.310}} & \scalea{0.388} \\ 
    & \scalea{336} & 
    \bst{\scalea{0.303}} & \bst{\scalea{0.370}} & \scalea{0.317} & \scalea{0.386} & 
    \bst{\scalea{0.308}} & \bst{\scalea{0.372}} & \scalea{0.316} & \scalea{0.384} & 
    \bst{\scalea{0.295}} & \bst{\scalea{0.368}} & \scalea{0.323} & \scalea{0.386} & 
    \bst{\scalea{0.312}} & \bst{\scalea{0.374}} & \scalea{0.327} & \scalea{0.391} & 
    \bst{\scalea{0.336}} & \bst{\scalea{0.398}} & \bst{\scalea{0.336}} & \scalea{0.407} \\ 
    & \scalea{720} & 
    \bst{\scalea{0.359}} & \bst{\scalea{0.408}} & \scalea{0.369} & \scalea{0.421} & 
    \scalea{0.356} & \bst{\scalea{0.410}} & \bst{\scalea{0.351}} & \scalea{0.414} & 
    \bst{\scalea{0.352}} & \bst{\scalea{0.407}} & \scalea{0.381} & \scalea{0.429} & 
    \bst{\scalea{0.320}} & \bst{\scalea{0.386}} & \scalea{0.344} & \scalea{0.407} & 
    \scalea{0.395} & \bst{\scalea{0.436}} & \bst{\scalea{0.388}} & \scalea{0.442} \\ 
    \cmidrule(lr){2-22}
    & \scalea{Avg} & 
    \bst{\scalea{0.289}} & \bst{\scalea{0.354}} & \scalea{0.304} & \scalea{0.373} & 
    \bst{\scalea{0.285}} & \bst{\scalea{0.353}} & \scalea{0.294} & \scalea{0.366} & 
    \bst{\scalea{0.280}} & \bst{\scalea{0.353}} & \scalea{0.307} & \scalea{0.374} & 
    \bst{\scalea{0.275}} & \bst{\scalea{0.348}} & \scalea{0.299} & \scalea{0.366} & 
    \scalea{0.327} & \bst{\scalea{0.389}} & \bst{\scalea{0.324}} & \scalea{0.398} \\ 
    \bottomrule
\end{tabular}
\end{threeparttable}
}
\end{table*}

\newpage
\subsection{Long-Term Forecasting Results and Analysis}
\label{app:J4}
\label{app:long_term_forecast}

For the long-term forecasting task, we adopt $\mcL_{\text{Harm}, \ell_p}$ with the $\ell_1$ norm, setting the hyperparameters to $\gamma=0.5$ and $\beta=0.3$ and \textbf{excluding the temporal MSE term}.
For structural orthogonality, dataset-specific strategies are implemented as variants of DFT and DWT.

Tables~\ref{tab:longterm_app} and~\ref{tab:longterm_app_part2} evaluate whether the supervision-level correction remains competitive against 13 forecasting baselines.

\paragraph{1. Statistical Ranking Against Full Baselines}
As summarized by the Count statistics, our method achieves the best performance in 17 out of 44 MSE cases and 31 out of 44 MAE cases.
It also reaches the Top-2 in 27 out of 44 MSE cases and 41 out of 44 MAE cases.
These rankings show that the harmonized objective remains competitive across 11 benchmarks covering energy, traffic, exchange-rate, and weather data.

\paragraph{2. Backbone Synergy: $\mcL_{\text{Harm}, \ell_p}$ vs. iTransformer}
Since our framework is deployed on the iTransformer backbone, the comparison between Ours and vanilla iTransformer measures the net effect of changing the supervision objective while keeping the backbone fixed.
Across 11 datasets, $\mcL_{\text{Harm}, \ell_p}$ yields mean per-dataset MSE/MAE gains of 5.18\%/5.10\% over iTransformer.
For example, on BE, the average MSE is reduced from 0.332 to 0.295, corresponding to an 11.1\% improvement.
This supports the objective-level interpretation: the performance ceiling of a strong architecture can still be constrained by point-wise temporal supervision.

\paragraph{3. Empirical Echo of EOB Theory and SSNR}
Table~\ref{tab:eob_mse_relation_app} gives a direct empirical echo of the EOB mechanism.
EOB theory identifies $\textit{SSNR}$ as the structural source of stochastic EOB under point-wise factorized supervision.
Accordingly, we compare the original temporal-domain $\textit{SSNR}_z$ with the $\textit{SSNR}_z$ after DFT/DWT structural orthogonalization, and align this reduction with the average forecasting-error change over four horizons.
The transform lowers $\textit{SSNR}_z$ on all 11 datasets, with an average reduction of 27.82\%.
This reduction is accompanied by lower forecasting errors in almost all cases: averaged across datasets, $\mathcal{L}_{\text{Harm},\ell_1}$ improves over temporal MSE by 5.18\% in MSE and 5.10\% in MAE.
The largest improvements appear on BE and FR, where $\textit{SSNR}_z$ drops by 58.03\% and 62.85\%, while MSE decreases by 11.14\% and 16.57\%, respectively.
Traffic is the main exception in MSE, despite its reduced $\textit{SSNR}_z$, although its MAE still improves by 4.96\%.
Thus, Table~\ref{tab:eob_mse_relation_app} does not imply a one-to-one deterministic mapping from $\textit{SSNR}_z$ to error, but supports the predicted direction: reducing structural dependence in the supervised target tends to reduce the forecasting error induced by point-wise temporal supervision.

\begin{table}[h]
\renewcommand{\arraystretch}{0.82} \setlength{\tabcolsep}{2.6pt} \scriptsize
\centering
\caption{
    \textbf{Empirical Relation Between Structural EOB and Forecasting Error.}
    $\Delta$SSNR denotes the relative reduction from the original temporal-domain $\textit{SSNR}_z$ to the DFT/DWT-domain $\textit{SSNR}_z$.
    MSE metric columns report average temporal-MSE and $\mathcal{L}_{\text{Harm},\ell_1}$ results over forecasting horizons $T \in \{96, 192, 336, 720\}$.
}
\label{tab:eob_mse_relation_app}
\begin{threeparttable}
\resizebox{0.86\linewidth}{!}{%
\begin{tabular}{c|cc|c|cc|cc}
    \toprule
    \multirow{2}{*}{\scaleb{Dataset}} &
    \multicolumn{2}{c|}{\scaleb{$\textit{SSNR}_z$}} &
    \multirow{1}{*}{\scaleb{$\Delta \textit{SSNR}$}} &
    \multicolumn{2}{c|}{\scaleb{MSE Metric}} &
    \multicolumn{2}{c}{\scaleb{Gain}} \\
    & \scalea{Orig.} & \scalea{DFT/DWT} & \scalea{(\%)} &
    \scalea{Temporal} & \scalea{$\mathcal{L}_{\text{Harm},\ell_1}$} & \scalea{MSE (\%)} & \scalea{MAE (\%)} \\
    \toprule
    \scalea{ETTh1} & \scalea{3.90} & \scalea{2.53} & \scalea{35.13} & \scalea{0.461} & \scalea{0.436} & \scalea{5.42} & \scalea{4.19} \\
    \scalea{ETTh2} & \scalea{6.38} & \scalea{5.03} & \scalea{21.16} & \scalea{0.390} & \scalea{0.373} & \scalea{4.36} & \scalea{3.88} \\
    \scalea{ETTm1} & \scalea{6.56} & \scalea{4.18} & \scalea{36.28} & \scalea{0.414} & \scalea{0.393} & \scalea{5.07} & \scalea{5.80} \\
    \scalea{ETTm2} & \scalea{8.58} & \scalea{6.69} & \scalea{22.03} & \scalea{0.291} & \scalea{0.282} & \scalea{3.09} & \scalea{4.18} \\
    \scalea{ECL} & \scalea{4.67} & \scalea{4.18} & \scalea{10.49} & \scalea{0.176} & \scalea{0.170} & \scalea{3.41} & \scalea{2.62} \\
    \scalea{Traffic} & \scalea{2.68} & \scalea{2.26} & \scalea{15.67} & \scalea{0.422} & \scalea{0.429} & \scalea{-1.66} & \scalea{4.96} \\
    \scalea{Weather} & \scalea{64.03} & \scalea{57.14} & \scalea{10.76} & \scalea{0.260} & \scalea{0.256} & \scalea{1.54} & \scalea{3.20} \\
    \scalea{Exchange} & \scalea{3.95} & \scalea{3.23} & \scalea{18.23} & \scalea{0.385} & \scalea{0.366} & \scalea{4.94} & \scalea{2.62} \\
    \scalea{BE} & \scalea{10.84} & \scalea{4.55} & \scalea{58.03} & \scalea{0.332} & \scalea{0.295} & \scalea{11.14} & \scalea{9.57} \\
    \scalea{FR} & \scalea{11.44} & \scalea{4.25} & \scalea{62.85} & \scalea{0.344} & \scalea{0.287} & \scalea{16.57} & \scalea{11.51} \\
    \scalea{PJM} & \scalea{19.08} & \scalea{16.14} & \scalea{15.41} & \scalea{0.294} & \scalea{0.285} & \scalea{3.06} & \scalea{3.55} \\
    \midrule
    \scalea{Avg} & \scalea{12.55} & \scalea{10.05} & \scalea{27.82} & \scalea{0.343} & \scalea{0.334} & \scalea{5.18} & \scalea{5.10} \\
    \bottomrule
\end{tabular}%
}
\end{threeparttable}
\end{table}

\paragraph{4. Competitiveness Across Forecasting Horizons}
$\mcL_{\text{Harm}, \ell_p}$ remains broadly competitive across forecasting horizons from 96 to 720, especially in MAE, where it reaches Top-2 performance in 41 out of 44 cases.
Its MSE advantage is less uniform at longer horizons, indicating that structural orthogonalization reduces a supervision-level bias but does not eliminate all horizon-specific optimization factors.
We further evaluate training randomness in Table~\ref{tab:forecasting_variance_app}: over 20 random seeds, the forecasting losses mostly fluctuate at the $10^{-4}$--$10^{-3}$ scale, indicating stable optimization behavior.
Together, the horizon-level ranking and seed-level variance results complement the SSNR--horizon analysis in Sec.~\ref{sec:EOB_verif}.

Table~\ref{tab:forecasting_variance_app} reports the per-dataset forecasting loss variance of $\mcL_{\text{Harm}, \ell_p}$ over 20 random seeds.

\begin{table}[ht]
\renewcommand{\arraystretch}{0.82} \setlength{\tabcolsep}{2.0pt} \scriptsize
\centering
\caption{
    \textbf{Forecasting Loss Variance Across Random Seeds.}
    The reported MSE/MAE variance values are computed over 20 random seeds for $\mcL_{\text{Harm}, \ell_p}$.
}
\label{tab:forecasting_variance_app}
\begin{threeparttable}
\resizebox{\linewidth}{!}{
\begin{tabular}{c|c|ccccccccccc}
    \toprule
    \scaleb{Horizon} & \scaleb{Metric} &
    \scaleb{ETTh1} & \scaleb{ETTh2} & \scaleb{ETTm1} & \scaleb{ETTm2} & \scaleb{ECL} & \scaleb{Traffic} &
    \scaleb{Weather} & \scaleb{Exchange} & \scaleb{BE} & \scaleb{FR} & \scaleb{PJM} \\
    \toprule
    \multirow{2}{*}{\scalea{96}} 
    & \scalea{MSE} & \scalea{0.0011} & \scalea{0.0017} & \scalea{0.0004} & \scalea{0.0005} & \scalea{0.0003} & \scalea{0.0007} & \scalea{0.0011} & \scalea{0.0022} & \scalea{0.0040} & \scalea{0.0005} & \scalea{0.0016} \\
    & \scalea{MAE} & \scalea{0.0007} & \scalea{0.0014} & \scalea{0.0005} & \scalea{0.0006} & \scalea{0.0004} & \scalea{0.0005} & \scalea{0.0008} & \scalea{0.0029} & \scalea{0.0025} & \scalea{0.0002} & \scalea{0.0011} \\
    \midrule
    \multirow{2}{*}{\scalea{192}} 
    & \scalea{MSE} & \scalea{0.0026} & \scalea{0.0043} & \scalea{0.0008} & \scalea{0.0003} & \scalea{0.0012} & \scalea{0.0007} & \scalea{0.0005} & \scalea{0.0030} & \scalea{0.0033} & \scalea{0.0002} & \scalea{0.0020} \\
    & \scalea{MAE} & \scalea{0.0016} & \scalea{0.0028} & \scalea{0.0004} & \scalea{0.0005} & \scalea{0.0013} & \scalea{0.0004} & \scalea{0.0004} & \scalea{0.0029} & \scalea{0.0021} & \scalea{0.0002} & \scalea{0.0013} \\
    \midrule
    \multirow{2}{*}{\scalea{336}} 
    & \scalea{MSE} & \scalea{0.0011} & \scalea{0.0022} & \scalea{0.0010} & \scalea{0.0006} & \scalea{0.0015} & \scalea{0.0009} & \scalea{0.0003} & \scalea{0.0044} & \scalea{0.0008} & \scalea{0.0010} & \scalea{0.0008} \\
    & \scalea{MAE} & \scalea{0.0007} & \scalea{0.0015} & \scalea{0.0004} & \scalea{0.0004} & \scalea{0.0015} & \scalea{0.0003} & \scalea{0.0003} & \scalea{0.0028} & \scalea{0.0006} & \scalea{0.0011} & \scalea{0.0008} \\
    \midrule
    \multirow{2}{*}{\scalea{720}} 
    & \scalea{MSE} & \scalea{0.0101} & \scalea{0.0037} & \scalea{0.0005} & \scalea{0.0007} & \scalea{0.0039} & \scalea{0.0021} & \scalea{0.0005} & \scalea{0.0126} & \scalea{0.0050} & \scalea{0.0030} & \scalea{0.0030} \\
    & \scalea{MAE} & \scalea{0.0052} & \scalea{0.0025} & \scalea{0.0003} & \scalea{0.0004} & \scalea{0.0028} & \scalea{0.0009} & \scalea{0.0004} & \scalea{0.0059} & \scalea{0.0034} & \scalea{0.0021} & \scalea{0.0016} \\
    \bottomrule
\end{tabular}
}
\end{threeparttable}
\end{table}

\begin{table}[ht]
\renewcommand{\arraystretch}{0.80} \setlength{\tabcolsep}{2.4pt} \scriptsize
\centering
\caption{
    \textbf{Long-Term Forecasting Results (Part I).}
    The input length is fixed to 96. \bst{Bold} typeface and \subbst{underlined} text represent the champion and runner-up performance, respectively. \emph{Avg} indicates results averaged over forecasting lengths $T \in \{96, 192, 336, 720\}$.
}
\label{tab:longterm_app}
\renewcommand{\multirowsetup}{\centering}
\begin{threeparttable}
\resizebox{0.89\linewidth}{!}{%
\begin{tabular}{c|c|cc|cc|cc|cc|cc|cc|cc}
    \toprule
    \multicolumn{2}{l}{\multirow{2}{*}{\scaleb{Models}}} & 
    \multicolumn{2}{c}{\scaleb{$\mcL_{\text{Harm}, \ell_p}$}} & \multicolumn{2}{c}{\scaleb{FreDF}} & \multicolumn{2}{c}{\scaleb{TimeMixer}} & \multicolumn{2}{c}{\scaleb{iTransformer}} & \multicolumn{2}{c}{\scaleb{PatchTST}} & \multicolumn{2}{c}{\scaleb{TimesNet}} & \multicolumn{2}{c}{\scaleb{DLinear}} \\
    \multicolumn{2}{c}{} &
    \multicolumn{2}{c}{\scaleb{Ours}} & \multicolumn{2}{c}{\scaleb{(2025)}} & \multicolumn{2}{c}{\scaleb{(2024)}} & \multicolumn{2}{c}{\scaleb{(2024)}} & \multicolumn{2}{c}{\scaleb{(2023)}} & \multicolumn{2}{c}{\scaleb{(2023)}} & \multicolumn{2}{c}{\scaleb{(2023)}} \\
    \cmidrule(lr){3-4} \cmidrule(lr){5-6} \cmidrule(lr){7-8} \cmidrule(lr){9-10}
    \cmidrule(lr){11-12} \cmidrule(lr){13-14} \cmidrule(lr){15-16}
    \multicolumn{2}{l}{\scaleb{Metrics}} & 
    \scalea{MSE} & \scalea{MAE} & \scalea{MSE} & \scalea{MAE} & \scalea{MSE} & \scalea{MAE} & \scalea{MSE} & \scalea{MAE} & \scalea{MSE} & \scalea{MAE} & \scalea{MSE} & \scalea{MAE} & \scalea{MSE} & \scalea{MAE} \\
    \toprule

    \multirow{5}{*}{{\rotatebox{90}{\scalebox{0.95}{ETTh1}}}} & \scalea{96} & 
    \bst{\scalea{0.378}} & \subbst{\scalea{0.397}} & \bst{\scalea{0.378}} & \bst{\scalea{0.396}} & 
    \scalea{0.382} & \scalea{0.399} & \scalea{0.393} & \scalea{0.408} & 
    \scalea{0.382} & \scalea{0.400} & \scalea{0.412} & \scalea{0.428} & 
    \scalea{0.396} & \scalea{0.411} \\ 

     & \scalea{192} & 
    \subbst{\scalea{0.429}} & \subbst{\scalea{0.426}} & \bst{\scalea{0.428}} & \bst{\scalea{0.423}} & 
    \scalea{0.437} & \scalea{0.430} & \scalea{0.447} & \scalea{0.440} & 
    \scalea{0.430} & \scalea{0.433} & \scalea{0.457} & \scalea{0.453} & 
    \scalea{0.446} & \scalea{0.442} \\
\bottomrule

\end{tabular}%
}
\end{threeparttable}
\end{table}

\begin{table}[ht]
\renewcommand{\arraystretch}{0.85} \setlength{\tabcolsep}{2.4pt} \scriptsize
\centering
\caption{
    \textbf{Long-Term Forecasting Results (Part II).}
    The input length is fixed to 96. \bst{Bold} typeface and \subbst{underlined} text represent the champion and runner-up performance, respectively. \emph{Avg} indicates results averaged over forecasting lengths $T \in \{96, 192, 336, 720\}$.
}
\label{tab:longterm_app_part2}
\renewcommand{\multirowsetup}{\centering}
\begin{threeparttable}
\resizebox{0.9\linewidth}{!}{%
\begin{tabular}{c|c|cc|cc|cc|cc|cc|cc|cc|cc}
    \toprule
    \multicolumn{2}{l}{\multirow{2}{*}{\scaleb{Models}}} & 
    \multicolumn{2}{c}{\scaleb{$\mcL_{\text{Harm}, \ell_p}$}} & \multicolumn{2}{c}{\scaleb{TSMixer}} & \multicolumn{2}{c}{\scaleb{FreTS}} & \multicolumn{2}{c}{\scaleb{MICN}} & \multicolumn{2}{c}{\scaleb{Pyraformer}} & \multicolumn{2}{c}{\scaleb{FEDformer}} & \multicolumn{2}{c}{\scaleb{Autoformer}} & \multicolumn{2}{c}{\scaleb{Transformer}} \\
    \multicolumn{2}{c}{} &
    \multicolumn{2}{c}{\scaleb{Ours}} & \multicolumn{2}{c}{\scaleb{(2023)}} & \multicolumn{2}{c}{\scaleb{(2023)}} & \multicolumn{2}{c}{\scaleb{(2023)}} & \multicolumn{2}{c}{\scaleb{(2022)}} & \multicolumn{2}{c}{\scaleb{(2022)}} & \multicolumn{2}{c}{\scaleb{(2021)}} & \multicolumn{2}{c}{\scaleb{(2017)}} \\
    \cmidrule(lr){3-4} \cmidrule(lr){5-6} \cmidrule(lr){7-8} \cmidrule(lr){9-10}
    \cmidrule(lr){11-12} \cmidrule(lr){13-14} \cmidrule(lr){15-16} \cmidrule(lr){17-18}
    \multicolumn{2}{l}{\scaleb{Metrics}} & 
    \scalea{MSE} & \scalea{MAE} & \scalea{MSE} & \scalea{MAE} & \scalea{MSE} & \scalea{MAE} & \scalea{MSE} & \scalea{MAE} & \scalea{MSE} & \scalea{MAE} & \scalea{MSE} & \scalea{MAE} & \scalea{MSE} & \scalea{MAE} & \scalea{MSE} & \scalea{MAE} \\
    \toprule

    \multirow{5}{*}{{\rotatebox{90}{\scalebox{0.95}{ETTh1}}}} & \scalea{96} & 
    \bst{\scalea{0.378}} & \subbst{\scalea{0.397}} & \scalea{0.495} & \scalea{0.504} & 
    \scalea{0.398} & \scalea{0.410} & \scalea{0.421} & \scalea{0.435} & 
    \scalea{0.721} & \scalea{0.643} & \subbst{\scalea{0.380}} & \scalea{0.420} & 
    \scalea{0.466} & \scalea{0.464} & \scalea{0.921} & \scalea{0.762} \\
\bottomrule

\end{tabular}%
}
\end{threeparttable}
\end{table}

\clearpage

\subsection{Missing Data Imputation Results}
\label{app:J5}
\label{app:imputation}

For the missing data imputation task, we adopt $\mcL_{\text{Harm}, \ell_p}$ with the $\ell_1$ norm, setting the hyperparameters to $\gamma=0.3$ and $\beta=0.3$.
For structural orthogonality, dataset-specific strategies are implemented as variants of DFT and DWT.
We use autoregressive reconstruction for the unmasked part of the series.
The imputation results are summarized in Table~\ref{tab:imp_app}.
We evaluate the models across four missing rates ($p_{\text{miss}} \in \{ 0.125,0.250,0.375,0.500 \}$).

\paragraph{1. Direct Gain Over iTransformer}
Since the proposed objective is deployed on iTransformer, the direct comparison with vanilla iTransformer measures the net effect of changing the supervision objective while keeping the backbone fixed.
Using the \emph{Avg} rows aggregated over 9 datasets, $\mcL_{\text{Harm}, \ell_p}$ reduces MSE/MAE by 27.4\%/19.4\% over iTransformer.
This gain is larger than in long-term forecasting, suggesting that structural orthogonalization is particularly helpful when the task requires recovering missing values from correlated context.

\paragraph{2. Ranking Against Imputation Baselines}
Out of 36 experimental scenarios, our method achieves 13 first places and 20 second places in MSE, resulting in a 91.7\% Top-2 coverage.
It also achieves 13 first places and 8 second places in MAE, corresponding to a 58.3\% Top-2 coverage.
These rankings show that the supervision-level correction is especially strong in MSE, while remaining competitive but less uniformly dominant in MAE.

\paragraph{3. Robustness Across Missing Rates}
As the missing rate $p_{\text{miss}}$ increases from 0.125 to 0.5, the task transitions from simple interpolation to more difficult signal recovery.

The MSE Top-2 statistics remain strong across missing rates, reaching Top-2 performance on 8, 8, 9, and 8 out of 9 datasets as $p_{\text{miss}}$ increases.
The MAE ranking is more variable, with Top-2 counts of 6, 5, 6, and 4 out of 9 datasets.
This indicates that the benefit is not limited to low-mask interpolation, while the MAE response becomes less uniform at higher missing rates.
Masking introduces local discontinuities in the time domain, while frequency/wavelet supervision emphasizes structural components that are less tied to individual missing timestamps.
This is consistent with the EOB view that point-wise temporal reconstruction can be biased when the target values remain strongly correlated.

\paragraph{4. Statistical Alignment and Debiasing}
From the EOB perspective, imputation is a bidirectional reconstruction task.
Although bidirectional context reduces local uncertainty, the masked targets can still be highly correlated.
Temporal MSE treats the reconstruction of each masked point as an independent event, while the harmonized objective constrains structural components in an approximately orthogonal domain.
This helps explain why the same supervision-level correction improves imputation without changing the backbone architecture.

\begin{table}[ht]
\renewcommand{\arraystretch}{1} \setlength{\tabcolsep}{2.4pt} \scriptsize
\centering
\caption{
    \textbf{Missing Data Imputation Results.}
    The input length is fixed to 96. \bst{Bold} typeface and \subbst{underlined} text represent the champion and runner-up performance, respectively. \emph{Avg} indicates results averaged over masking ratios $p_{\text{miss}} \in \{0.125, 0.25, 0.375, 0.5\}$.
}
\label{tab:imp_app}
\renewcommand{\multirowsetup}{\centering}
\begin{threeparttable}
\begin{tabular}{c|c|cc|cc|cc|cc|cc|cc|cc|cc|cc|cc}
    \toprule
    \multicolumn{2}{l}{\multirow{2}{*}{\scaleb{Models}}} & 
    \multicolumn{2}{c}{\scaleb{$\mcL_{\text{Harm}, \ell_p}$}} & \multicolumn{2}{c}{\scaleb{FreDF}} &
    \multicolumn{2}{c}{\scaleb{iTransformer}}   & \multicolumn{2}{c}{\scaleb{FreTS}} &
    \multicolumn{2}{c}{\scaleb{TimesNet}}       & \multicolumn{2}{c}{\scaleb{MICN}}  &
    \multicolumn{2}{c}{\scaleb{TiDE}}           & \multicolumn{2}{c}{\scaleb{DLinear}} &
    \multicolumn{2}{c}{\scaleb{FEDformer}}      & \multicolumn{2}{c}{\scaleb{Autoformer}} \\
    \multicolumn{2}{c}{} &
    \multicolumn{2}{c}{\scaleb{\textbf{(ours)}}} & \multicolumn{2}{c}{\scaleb{(2025)}} & 
    \multicolumn{2}{c}{\scaleb{(2024)}} & \multicolumn{2}{c}{\scaleb{(2023)}} & 
    \multicolumn{2}{c}{\scaleb{(2023)}} & \multicolumn{2}{c}{\scaleb{(2023)}} & 
    \multicolumn{2}{c}{\scaleb{(2023)}} & \multicolumn{2}{c}{\scaleb{(2023)}} & 
    \multicolumn{2}{c}{\scaleb{(2022)}} & \multicolumn{2}{c}{\scaleb{(2021)}} \\
    
    \cmidrule(lr){3-4} \cmidrule(lr){5-6} \cmidrule(lr){7-8} \cmidrule(lr){9-10}
    \cmidrule(lr){11-12} \cmidrule(lr){13-14} \cmidrule(lr){15-16} \cmidrule(lr){17-18}
    \cmidrule(lr){19-20} \cmidrule(lr){21-22}
    
    & \scaleb{$p_{\text{miss}}$}  & 
    \scalea{MSE} & \scalea{MAE} & \scalea{MSE} & \scalea{MAE} &
    \scalea{MSE} & \scalea{MAE} & \scalea{MSE} & \scalea{MAE} &
    \scalea{MSE} & \scalea{MAE} & \scalea{MSE} & \scalea{MAE} &
    \scalea{MSE} & \scalea{MAE} & \scalea{MSE} & \scalea{MAE} &
    \scalea{MSE} & \scalea{MAE} & \scalea{MSE} & \scalea{MAE} \\
\bottomrule
\end{tabular}
\end{threeparttable}
\end{table}

\clearpage

\subsection{Ablation Experiments}
\label{app:J6}

Table~\ref{tab:ablation_harmnorm_app} disentangles the two ingredients of the proposed objective.
Compared with iTransformer, w/o Harm still applies the DWT/DFT-based structural orthogonalization but removes the harmonized norm weighting.
The large and consistent gap between iTransformer and w/o Harm therefore shows that the main enhancement comes from transforming the supervised target into a more orthogonal structural domain.
The further gap between w/o Harm and $\mcL_{\text{Harm}, \ell_p}$ is smaller, indicating that harmonized norm optimization mainly contributes an additional robust optimization gain by reducing scale-dominated training effects rather than serving as the primary source of improvement.

\begin{table}[ht]
\renewcommand{\arraystretch}{0.82} \setlength{\tabcolsep}{1.7pt} \scriptsize
\centering
\caption{
    \textbf{Ablation Study on Harmonized Norm.}
     w/o Harm removes the harmonized norm weighting from the proposed objective.
    The $\mcL_{\text{Harm}, \ell_p}$ and iTransformer columns are copied from Table~\ref{tab:longterm_app} for direct comparison.
}
\label{tab:ablation_harmnorm_app}
\renewcommand{\multirowsetup}{\centering}
\begin{threeparttable}
\resizebox{\linewidth}{!}{
\begin{tabular}{c|c|cc|cc|cc||c|c|cc|cc|cc}
    \toprule
    \multicolumn{2}{c|}{\multirow{2}{*}{\scaleb{Dataset}}} & 
    \multicolumn{2}{c|}{\scaleb{$\mcL_{\text{Harm}, \ell_p}$}} & \multicolumn{2}{c|}{\scaleb{w/o Harm}} & \multicolumn{2}{c||}{\scaleb{iTransformer}} & 
    \multicolumn{2}{c|}{\multirow{2}{*}{\scaleb{Dataset}}} & 
    \multicolumn{2}{c|}{\scaleb{$\mcL_{\text{Harm}, \ell_p}$}} & \multicolumn{2}{c|}{\scaleb{w/o Harm}} & \multicolumn{2}{c}{\scaleb{iTransformer}} \\
    \multicolumn{2}{c|}{} & \scalea{MSE} & \scalea{MAE} & \scalea{MSE} & \scalea{MAE} & \scalea{MSE} & \scalea{MAE} & 
    \multicolumn{2}{c|}{} & \scalea{MSE} & \scalea{MAE} & \scalea{MSE} & \scalea{MAE} & \scalea{MSE} & \scalea{MAE} \\
    \toprule
    \multirow{5}{*}{{\rotatebox{90}{\scalebox{0.95}{ETTh1}}}} 
    & \scalea{96} & \bst{\scalea{0.378}} & \bst{\scalea{0.397}} & \subbst{\scalea{0.381}} & \subbst{\scalea{0.399}} & \scalea{0.393} & \scalea{0.408} & 
    \multirow{5}{*}{{\rotatebox{90}{\scalebox{0.95}{Weather}}}} 
    & \scalea{96} & \subbst{\scalea{0.174}} & \bst{\scalea{0.209}} & \bst{\scalea{0.170}} & \subbst{\scalea{0.210}} & \scalea{0.176} & \scalea{0.217} \\
    & \scalea{192} & \bst{\scalea{0.429}} & \bst{\scalea{0.426}} & \bst{\scalea{0.429}} & \subbst{\scalea{0.430}} & \subbst{\scalea{0.447}} & \scalea{0.440} &  
    & \scalea{192} & \bst{\scalea{0.220}} & \bst{\scalea{0.250}} & \subbst{\scalea{0.225}} & \subbst{\scalea{0.251}} & \subbst{\scalea{0.225}} & \scalea{0.258} \\
    & \scalea{336} & \bst{\scalea{0.464}} & \bst{\scalea{0.442}} & \subbst{\scalea{0.473}} & \subbst{\scalea{0.449}} & \scalea{0.489} & \scalea{0.463} &  
    & \scalea{336} & \bst{\scalea{0.279}} & \bst{\scalea{0.292}} & \bst{\scalea{0.279}} & \subbst{\scalea{0.293}} & \subbst{\scalea{0.282}} & \scalea{0.299} \\
    & \scalea{720} & \bst{\scalea{0.474}} & \bst{\scalea{0.472}} & \subbst{\scalea{0.485}} & \subbst{\scalea{0.475}} & \scalea{0.514} & \scalea{0.499} &  
    & \scalea{720} & \bst{\scalea{0.355}} & \bst{\scalea{0.343}} & \subbst{\scalea{0.357}} & \bst{\scalea{0.343}} & \scalea{0.359} & \subbst{\scalea{0.350}} \\
    \cmidrule(lr){2-8} \cmidrule(lr){10-16}
    & \scalea{Avg} & \bst{\scalea{0.436}} & \bst{\scalea{0.434}} & \subbst{\scalea{0.442}} & \subbst{\scalea{0.438}} & \scalea{0.461} & \scalea{0.453} &  
    & \scalea{Avg} & \bst{\scalea{0.256}} & \bst{\scalea{0.272}} & \subbst{\scalea{0.258}} & \subbst{\scalea{0.274}} & \scalea{0.260} & \scalea{0.281} \\
    \midrule
    \multirow{5}{*}{{\rotatebox{90}{\scalebox{0.95}{ETTh2}}}} 
    & \scalea{96} & \bst{\scalea{0.288}} & \bst{\scalea{0.336}} & \subbst{\scalea{0.291}} & \subbst{\scalea{0.338}} & \scalea{0.303} & \scalea{0.353} & 
    \multirow{5}{*}{{\rotatebox{90}{\scalebox{0.95}{Exchange}}}} 
    & \scalea{96} & \bst{\scalea{0.087}} & \bst{\scalea{0.207}} & \subbst{\scalea{0.091}} & \subbst{\scalea{0.213}} & \subbst{\scalea{0.091}} & \subbst{\scalea{0.213}} \\
    & \scalea{192} & \bst{\scalea{0.365}} & \bst{\scalea{0.385}} & \subbst{\scalea{0.370}} & \subbst{\scalea{0.391}} & \scalea{0.389} & \scalea{0.404} &  
    & \scalea{192} & \bst{\scalea{0.185}} & \bst{\scalea{0.306}} & \subbst{\scalea{0.186}} & \subbst{\scalea{0.307}} & \scalea{0.190} & \scalea{0.312} \\
    & \scalea{336} & \bst{\scalea{0.414}} & \bst{\scalea{0.424}} & \subbst{\scalea{0.419}} & \bst{\scalea{0.424}} & \scalea{0.431} & \subbst{\scalea{0.439}} &  
    & \scalea{336} & \bst{\scalea{0.354}} & \subbst{\scalea{0.430}} & \subbst{\scalea{0.358}} & \bst{\scalea{0.429}} & \scalea{0.365} & \scalea{0.439} \\
    & \scalea{720} & \bst{\scalea{0.423}} & \bst{\scalea{0.440}} & \bst{\scalea{0.423}} & \subbst{\scalea{0.443}} & \subbst{\scalea{0.436}} & \scalea{0.452} &  
    & \scalea{720} & \bst{\scalea{0.840}} & \bst{\scalea{0.693}} & \subbst{\scalea{0.857}} & \subbst{\scalea{0.704}} & \scalea{0.895} & \scalea{0.717} \\
    \cmidrule(lr){2-8} \cmidrule(lr){10-16}
    & \scalea{Avg} & \bst{\scalea{0.373}} & \bst{\scalea{0.396}} & \subbst{\scalea{0.376}} & \subbst{\scalea{0.399}} & \scalea{0.390} & \scalea{0.412} &  
    & \scalea{Avg} & \bst{\scalea{0.366}} & \bst{\scalea{0.409}} & \subbst{\scalea{0.373}} & \subbst{\scalea{0.413}} & \scalea{0.385} & \scalea{0.420} \\
    \midrule
    \multirow{5}{*}{{\rotatebox{90}{\scalebox{0.95}{ETTm1}}}} 
    & \scalea{96} & \bst{\scalea{0.314}} & \bst{\scalea{0.346}} & \subbst{\scalea{0.325}} & \subbst{\scalea{0.355}} & \scalea{0.349} & \scalea{0.379} & 
    \multirow{5}{*}{{\rotatebox{90}{\scalebox{0.95}{BE}}}} 
    & \scalea{96} & \bst{\scalea{0.207}} & \bst{\scalea{0.226}} & \subbst{\scalea{0.219}} & \subbst{\scalea{0.235}} & \scalea{0.251} & \scalea{0.262} \\
    & \scalea{192} & \bst{\scalea{0.369}} & \bst{\scalea{0.373}} & \subbst{\scalea{0.374}} & \subbst{\scalea{0.381}} & \scalea{0.388} & \scalea{0.397} &  
    & \scalea{192} & \bst{\scalea{0.255}} & \bst{\scalea{0.264}} & \subbst{\scalea{0.272}} & \subbst{\scalea{0.278}} & \scalea{0.299} & \scalea{0.298} \\
    & \scalea{336} & \subbst{\scalea{0.411}} & \subbst{\scalea{0.400}} & \bst{\scalea{0.409}} & \bst{\scalea{0.399}} & \scalea{0.424} & \scalea{0.420} &  
    & \scalea{336} & \bst{\scalea{0.297}} & \bst{\scalea{0.300}} & \subbst{\scalea{0.307}} & \subbst{\scalea{0.309}} & \scalea{0.352} & \scalea{0.342} \\
    & \scalea{720} & \subbst{\scalea{0.479}} & \bst{\scalea{0.439}} & \bst{\scalea{0.478}} & \subbst{\scalea{0.444}} & \scalea{0.494} & \scalea{0.459} &  
    & \scalea{720} & \subbst{\scalea{0.421}} & \bst{\scalea{0.383}} & \bst{\scalea{0.416}} & \subbst{\scalea{0.389}} & \scalea{0.424} & \scalea{0.393} \\
    \cmidrule(lr){2-8} \cmidrule(lr){10-16}
    & \scalea{Avg} & \bst{\scalea{0.393}} & \bst{\scalea{0.390}} & \subbst{\scalea{0.397}} & \subbst{\scalea{0.395}} & \scalea{0.414} & \scalea{0.414} &  
    & \scalea{Avg} & \bst{\scalea{0.295}} & \bst{\scalea{0.293}} & \subbst{\scalea{0.303}} & \subbst{\scalea{0.303}} & \scalea{0.332} & \scalea{0.324} \\
    \midrule
    \multirow{5}{*}{{\rotatebox{90}{\scalebox{0.95}{ETTm2}}}} 
    & \scalea{96} & \bst{\scalea{0.176}} & \bst{\scalea{0.253}} & \bst{\scalea{0.176}} & \subbst{\scalea{0.255}} & \subbst{\scalea{0.184}} & \scalea{0.270} & 
    \multirow{5}{*}{{\rotatebox{90}{\scalebox{0.95}{FR}}}} 
    & \scalea{96} & \bst{\scalea{0.228}} & \bst{\scalea{0.213}} & \subbst{\scalea{0.236}} & \subbst{\scalea{0.223}} & \scalea{0.275} & \scalea{0.245} \\
    & \scalea{192} & \bst{\scalea{0.241}} & \bst{\scalea{0.296}} & \subbst{\scalea{0.243}} & \subbst{\scalea{0.298}} & \scalea{0.252} & \scalea{0.313} &  
    & \scalea{192} & \bst{\scalea{0.254}} & \bst{\scalea{0.247}} & \subbst{\scalea{0.268}} & \subbst{\scalea{0.257}} & \scalea{0.303} & \scalea{0.277} \\
    & \scalea{336} & \bst{\scalea{0.305}} & \bst{\scalea{0.337}} & \subbst{\scalea{0.309}} & \subbst{\scalea{0.343}} & \scalea{0.316} & \scalea{0.353} &  
    & \scalea{336} & \bst{\scalea{0.302}} & \bst{\scalea{0.285}} & \subbst{\scalea{0.314}} & \subbst{\scalea{0.294}} & \scalea{0.360} & \scalea{0.317} \\
    & \scalea{720} & \bst{\scalea{0.406}} & \bst{\scalea{0.396}} & \subbst{\scalea{0.410}} & \subbst{\scalea{0.397}} & \scalea{0.412} & \scalea{0.406} &  
    & \scalea{720} & \bst{\scalea{0.363}} & \bst{\scalea{0.332}} & \subbst{\scalea{0.375}} & \subbst{\scalea{0.341}} & \scalea{0.439} & \scalea{0.377} \\
    \cmidrule(lr){2-8} \cmidrule(lr){10-16}
    & \scalea{Avg} & \bst{\scalea{0.282}} & \bst{\scalea{0.321}} & \subbst{\scalea{0.284}} & \subbst{\scalea{0.323}} & \scalea{0.291} & \scalea{0.335} &  
    & \scalea{Avg} & \bst{\scalea{0.287}} & \bst{\scalea{0.269}} & \subbst{\scalea{0.298}} & \subbst{\scalea{0.279}} & \scalea{0.344} & \scalea{0.304} \\
    \midrule
    \multirow{5}{*}{{\rotatebox{90}{\scalebox{0.95}{ECL}}}} 
    & \scalea{96} & \bst{\scalea{0.144}} & \bst{\scalea{0.233}} & \subbst{\scalea{0.148}} & \subbst{\scalea{0.234}} & \subbst{\scalea{0.148}} & \scalea{0.240} & 
    \multirow{5}{*}{{\rotatebox{90}{\scalebox{0.95}{PJM}}}} 
    & \scalea{96} & \bst{\scalea{0.202}} & \bst{\scalea{0.288}} & \subbst{\scalea{0.211}} & \subbst{\scalea{0.296}} & \scalea{0.222} & \scalea{0.309} \\
    & \scalea{192} & \subbst{\scalea{0.158}} & \bst{\scalea{0.247}} & \bst{\scalea{0.156}} & \subbst{\scalea{0.248}} & \scalea{0.165} & \scalea{0.256} &  
    & \scalea{192} & \bst{\scalea{0.273}} & \bst{\scalea{0.342}} & \subbst{\scalea{0.280}} & \subbst{\scalea{0.348}} & \scalea{0.290} & \scalea{0.359} \\
    & \scalea{336} & \bst{\scalea{0.172}} & \bst{\scalea{0.263}} & \bst{\scalea{0.172}} & \subbst{\scalea{0.265}} & \subbst{\scalea{0.179}} & \scalea{0.272} &  
    & \scalea{336} & \bst{\scalea{0.308}} & \bst{\scalea{0.372}} & \subbst{\scalea{0.311}} & \subbst{\scalea{0.374}} & \scalea{0.316} & \scalea{0.384} \\
    & \scalea{720} & \bst{\scalea{0.206}} & \subbst{\scalea{0.296}} & \subbst{\scalea{0.209}} & \bst{\scalea{0.295}} & \scalea{0.213} & \scalea{0.301} &  
    & \scalea{720} & \scalea{0.356} & \bst{\scalea{0.410}} & \subbst{\scalea{0.355}} & \subbst{\scalea{0.411}} & \bst{\scalea{0.351}} & \scalea{0.414} \\
    \cmidrule(lr){2-8} \cmidrule(lr){10-16}
    & \scalea{Avg} & \bst{\scalea{0.170}} & \bst{\scalea{0.260}} & \subbst{\scalea{0.171}} & \subbst{\scalea{0.261}} & \scalea{0.176} & \scalea{0.267} &  
    & \scalea{Avg} & \bst{\scalea{0.285}} & \bst{\scalea{0.353}} & \subbst{\scalea{0.289}} & \subbst{\scalea{0.357}} & \scalea{0.294} & \scalea{0.366} \\
    \midrule
    \multirow{5}{*}{{\rotatebox{90}{\scalebox{0.95}{Traffic}}}} 
    & \scalea{96} & \scalea{0.396} & \bst{\scalea{0.253}} & \bst{\scalea{0.391}} & \subbst{\scalea{0.260}} & \subbst{\scalea{0.393}} & \scalea{0.269} &  
    &  &  &  &  &  &  &  \\
    & \scalea{192} & \scalea{0.418} & \bst{\scalea{0.262}} & \subbst{\scalea{0.417}} & \subbst{\scalea{0.269}} & \bst{\scalea{0.413}} & \scalea{0.277} &  
    &  &  &  &  &  &  &  \\
    & \scalea{336} & \subbst{\scalea{0.434}} & \bst{\scalea{0.268}} & \bst{\scalea{0.424}} & \subbst{\scalea{0.274}} & \bst{\scalea{0.424}} & \scalea{0.283} &  
    &  &  &  &  &  &  &  \\
    & \scalea{720} & \scalea{0.468} & \bst{\scalea{0.287}} & \subbst{\scalea{0.460}} & \subbst{\scalea{0.294}} & \bst{\scalea{0.458}} & \scalea{0.300} &  
    &  &  &  &  &  &  &  \\
    \cmidrule(lr){2-8}
    & \scalea{Avg} & \scalea{0.429} & \bst{\scalea{0.268}} & \subbst{\scalea{0.423}} & \subbst{\scalea{0.274}} & \bst{\scalea{0.422}} & \scalea{0.282} &  
    &  &  &  &  &  &  &  \\
    \bottomrule
\end{tabular}
}
\end{threeparttable}
\end{table}


\end{document}